# Towards CONUS-Wide ML-Augmented Conceptually-Interpretable Modeling of Catchment-Scale Precipitation-Storage-Runoff Dynamics

Yuan-Heng Wang[1,2], Yang Yang[3,4], Fabio Ciulla[1], Hoshin V. Gupta[2], Charuleka Varadharajan[1]

[1] Earth and Environmental Science Area, Lawrence Berkeley National Lab, Berkeley, CA
[2] Department of Hydrology and Atmospheric Science, University of Arizona, Tucson, AZ
[3] School for the Environment, University of Massachusetts Boston, MA
[4] Department of Civil Engineering, The University of Hong Kong, Hong Kong SAR, China

Corresponding Author: Yuan-Heng Wang, Ph.D.

Email: yhwang0730@gmail.com

## Abstract

While many recent studies are dedicated to ML-based large-sample hydrologic modeling, these efforts have not necessarily translated into predictive improvements that are grounded in enhanced physical-conceptual understanding. Here, we report on a CONUS-wide large-sample study (spanning diverse hydro-geo-climatic conditions) using ML-augmented physically-interpretable catchment-scale models of varying complexity, based in the Mass-Conserving Perceptron (MCP). Results were evaluated using attribute masks such as snow regime, forest cover, and climate zone. Our results indicate the importance of selecting model architectures of appropriate model complexity, based on how process dominance varies with hydrological regime. Benchmark comparisons show that physically-interpretable mass-conserving MCP-based models can achieve performance comparable to data-based models based in the Long Short-Term Memory network (LSTM) architecture. Overall, this study highlights the potential of a theory-informed, physically grounded approach to large-sample hydrology, with emphasis on mechanistic understanding and the development of parsimonious and interpretable model architectures, thereby laying the foundation for future "*models of everywhere*" that architecturally encode information about spatially- and temporally-varying process dominance.

## Plain Language Summary

We conducted CONUS-wide ML-based large-sample catchment-scale modeling of daily precipitation–storage–runoff (PSR) dynamics using models of varying complexity and process dominance constructed using mass-conserving-perceptron (MCP) units. In contrast to state-of-the-art ML approaches that apply a single "*universal*" ML architecture across all catchments, our approach identifies parsimonious basin-specific architectures for each catchment based on complexity and process-dominance. In particular, we leverage ML to propose physically-interpretable MCP-based computational units that reveal how hydrological process dominance varies geographically across the continental US. This study lays the foundation for future "*models of everywhere*" that explicitly acknowledge geo-hydro-climatic variability, so that predictive power can be firmly rooted in improved theoretical understanding, as is essential for the development of domain-specific ML-based models in the Geosciences.

## 1. Introduction

### 1.1 Large-Sample Hydrology as a Foundation for Discovering Generalizable Hydrological Dynamics

[1] Early efforts in large-sample hydrology (LSH) date back several decades (*Gupta et al., 2014*). In the context of surface hydrology, LSH emphasizes the analysis of data from a large number of catchments representing diverse geo-hydro-climatological regions—rather than focusing solely on a few intensively studied sites—so as to gain deeper insights into hydrologic processes across diverse conditions (*Duan et al., 2006; Addor et al., 2017; Kratzert et al., 2023*).

[2] Although large-sample studies may sacrifice some site-specific detail and face challenges such as poor comparability, lack of uncertainty estimates, and incomplete representation of human impacts (*Addor et al., 2020*), they remain essential for advancing a broader understanding of hydrologic behavior and model performance under changing environmental conditions (*Wagener et al., 2010; Montanari et al., 2013; Troch et al., 2015; Kreibich et al., 2025*). More importantly, LSH enables the systematic testing of model hypotheses across diverse catchments and conditions (*Gupta et al., 2008; Gong et al., 2013; Kumar & Gupta, 2020; Nearing et al., 2020a*).

[3] In modeling catchment-scale precipitation–storage–runoff (PSR) dynamics, conceptual rainfall–runoff (CRR) models remain widely used due to their computational efficiency and ability to represent dominant processes at the catchment scale (*Burnash & Larry Ferral, 1973; Perrin et al., 2003; Wang & Gupta, 2024ab*). Their use of parsimonious directed-graph structures makes them physically-interpretable (*Weijs & Ruddell, 2020; Gharari et al., 2021*; *Gupta & Nearing, 2014*), enabling hypothesis testing and systematic comparison of process representations across locations (*Clark et al., 2008; Fenicia et al., 2008; Knoben et al., 2019a;*).

## 1.2 Current LSH-Based Deep Learning Boosts Predictive Skill but Offers Limited Physical Insight

[4] The predictive accuracy of conceptual rainfall–runoff (CRR) models remains generally inferior to that of data-driven approaches (*Hsu et al., 1995; Kratzert et al., 2019b; Wang & Gupta, 2024b*). In recent years, deep learning (DL) methods have achieved substantial improvements in hydrological prediction by learning complex, high-dimensional relationships directly from data (*Shen, 2018; Kratzert et al., 2018*).

[5] This shift toward large-sample modeling has encouraged the development of "universal" models that can be applied across diverse catchments (*Blair et al, 2019; Beven, 2025*). In particular, Long Short-Term Memory (LSTM) networks trained on large-sample datasets have become the dominant paradigm for regional hydrological modeling (*Kratzert et al., 2024*). These models consistently outperform both site-specific models and traditional CRR approaches in terms of predictive accuracy (*Kratzert et al., 2019a,b; Lees et al., 2021; Mai et al., 2022; Wang et al., 2022; Arsenault et al., 2023*).

[6] Despite these advances, extracting physically interpretable insights from DL models remains challenging (*Jiang et al., 2024; Maier et al., 2024; Messeri & Crockett, 2024*). Existing approaches attempt to address this limitation through post-hoc analyses or by incorporating physical constraints into model architectures (*Hoedt et al., 2021; Frame et al., 2022; Wi & Steinschneider, 2024; Yu et al., 2025*), but these efforts have not yet resolved the trade-off between predictive accuracy and physical interpretability.

## 1.3 Pre-ML Investigations into Dominant Processes and Catchment Hydrological Similarity

[7] Despite recent advances in predictive accuracy, translating these gains into physically interpretable understanding remains challenging. Prior to deep learning, substantial effort focused on understanding catchment PSR dynamics through (i) hypothesis testing with CRR models, (ii) regionalization and parameter transfer, (iii) catchment classification, and (iv) linking catchment attributes to hydrological behavior.

[8] A long-standing approach has been to select or tailor model structures to represent dominant processes based on expert knowledge, whereby models formalize process hypotheses. The literature has long advocated for process-based modeling, emphasizing models that "*provide the right answers for the right reasons*" (*Kirchner, 2006; Gupta et al., 2008*). Supported by diagnostic evaluation and process adequacy (*Bahremand, 2016*), this approach offers strong physical interpretability but remains limited by subjective heuristics and challenges in transferring model structures across diverse regions.

[9] To reduce subjectivity, multiple working hypotheses and modular frameworks have been developed to compare alternative conceptual structures. Frameworks such as FUSE, SuperflexPy, and Raven enable systematic evaluation of process representations (*Clark et al., 2008; Fenicia et al., 2011; Molin et al., 2021; Craig et al., 2020*). Recent work has also explored latent-space approaches to characterize catchment behavior (*Yang and Chui, 2023*).

[10] However, these approaches face several challenges. Multiple distinct model structures can produce similar predictive results (the equifinality problem; *Beven, 2006*), while structural and data errors can affect model evaluation. Inference may also depend on the choice of evaluation data and objective functions (*Knoben et al., 2019a; Knoben et al., 2020; Vrugt, 2024*), making it difficult to identify robust and transferable representations of dominant hydrological processes.

[11] Meanwhile, regionalization and parameter-transfer approaches attempt to link catchment descriptors to model parameters and emergent hydrologic behavior. Driven by the PUB challenge, transfer functions relating climate, topography, soils, and land cover to conceptual model parameters have been widely investigated (*Hrachowitz et al., 2013; Swain & Patra, 2017; Feigl et al, 2026*). While these approaches provide insight into process controls, they face challenges such as attribute collinearity, non-stationarity, human impacts, and limited transferability across regions (*Wagener et al., 2010; Addor et al., 2020*).

[12] Finally, systematic exploration of model-structure space has shown that "*what works where*" depends on assembling appropriate combinations of storages, flux pathways, and constitutive functions (*Fenicia et al., 2011; Clark et al., 2011; Molin et al., 2021*). Different model structures can achieve similar predictive performance while implying different internal processes when evaluated using diagnostic measures such as water balance, recession analysis, or seasonality (*Knoben et al., 2020; Kiraz et al., 2023; Hartmann et al., 2013*). These findings highlight the difficulty of identifying physically meaningful and unique process representations across diverse catchments.

### 1.4 Advancing Hydrological Process Understanding with Conceptually Interpretable DL

[13] In the age of DL, the following question arises:

> *Should we strive for a single universal model that adapts to context internally, or should we develop families of models tailored to catchment-specific process dominance with transparent pathways for interpretation?*

[14] In this regard, modern machine learning (ML) technologies continue to revolutionize the development of ***interpretable*** geoscientific models (*Fleming et al., 2021; Shen et al., 2023*). Arguably, *Jiang et al. (2020)* and *Tsai et al (2021)* conducted the first large-sample studies that explored this approach. By embedding a CRR model into a neural network architecture, they showed that predictive accuracy approaching that of the LSTM network can be achieved (*Feng et al., 2022*), while offering improved behavioral expressivity and physical interpretability. However, directly embedding a single fixed CRR structure into a neural network may not improve predictions, particularly when the selected CRR structure is not well aligned with the dominant processes of a given region.

[15] More recently, studies have shown that relatively simple, conceptually interpretable models can be learned directly from data while achieving predictive performance comparable to state-of-the-art approaches (*De La Fuente et al., 2024, 2025; Wang and Gupta, 2024ab*). For example, the Regional HydroLSTM framework enables regionally adaptive model behavior, while mass-conserving perceptron (MCP) architectures demonstrate that physically interpretable neural networks can achieve near-optimal performance with simplified structures consistent with traditional CRR models (*Wang & Gupta, 2025*).

### 1.5 Goals and Contributions of this Study

[16] Further progress in hydrological understanding will likely be driven by advances in two complementary directions:

(1) Development of new kinds of data sources from which novel information can be gleaned.

(2) Use of creative ways to bridge the gaps between current physical understanding (hydrological theory) and the power of ML to extract relevant information from data.

[17] This study follows the second direction, leveraging machine learning to develop representations of how hydrological process dominance varies geographically across CONUS. Building on prior work introducing the mass-conserving perceptron (MCP; *Wang ,2023; Wang & Gupta 2024ab; Wang & Gupta 2025*) as a physically interpretable computational unit, we focus on its use for constructing parsimonious, interpretable models that support both predictive performance and hypothesis testing. Consistent with this objective, we employ relatively simple MCP-based model architectures trained individually at each catchment. This approach allows

us to examine how model structure and complexity vary geographically in response to differences in climate, topography, and vegetation.

[18] This approach does not take direct and immediate advantage of the information (contained in the data) that may be shared across multiple locations (*Kratzert et al., 2019, 2024; Feigl et al., 2020, 2026; De La Fuente et al., 2025*). However, previous efforts that do so have mainly focused on the use of "*universal*" model architectures. Instead, our interest lies in ML-based strategies that account for regional diversity in hydrological process dominance and associated model-structural complexity, and we consider the results reported here to be a necessary step towards achieving that more ambitious goal, one that we will address in future work.

[19] The core contribution of this study is a CONUS-wide large-sample investigation using MCP-based models to examine the relationship between predictive performance and model complexity. Rather than pursuing a single universal architecture, we highlight the importance of aligning model complexity with regional process characteristics. In doing so, this work advances the development of parsimonious, interpretable ML-augmented hydrological models that support both local mechanistic understanding and strong predictive skill.

### 1.6 Organization

[20] Section 2 briefly presents the methods used in this study, including data sets, regional masks, MCP-based component modules used for model construction, and procedures used for model training and assessment. Section 3 presents the results of our experiments, along with a benchmark comparison against some relevant ML benchmarks. Our conclusions and a discussion of implications appear in Section 4, along with suggestions for future work. To maintain brevity, many of the details are provided in the Appendices and supplementary materials.

## 2. Methods

### 2.1 Study Region and Data Used

#### 2.1.1 Meteorological Input Forcing and Streamflow Output

[21] To develop and test MCP-based catchment-scale models for analyzing PSR dynamics across diverse geo-hydro-climatic regions, we selected the continental United States (CONUS). These catchments span a wide range of hydro-climatic conditions, including regions where snow processes are either dominant or negligible (see **Section 2.3**). Accordingly, the model architectures were designed to represent water storage in both snow and soil moisture.

[22] For model development and evaluation, we used the CAMELS (Catchment Attributes and Meteorology for Large-sample Studies) dataset (*Addor et al., 2017*), which provides data for 671 minimally disturbed catchments across the United States and is curated by NCAR. These catchments range in size from 4 to 25,000 km$^2$ and represent diverse geological and eco-climatic settings (*Newman et al., 2015*).

[23] Meteorological inputs include precipitation and potential evapotranspiration (PET). Precipitation was obtained from the Maurer forcing dataset, and PET was prepared following *Gnann (2022)*, where PET is based on the Priestley–Taylor formulation with the default coefficient of 1.26. Similar CAMELS-based PET treatments have also been used in large-sample hydrologic modeling studies (e.g., *Knoben et al., 2020*). We selected 513 catchments with continuous streamflow records from WY 1982 to WY 2008 for model training and evaluation.

#### 2.1.2 Snow Water Equivalent (SWE)

[24] As a target variable for representing snow accumulation and melt dynamics, we used the University of Arizona (UA) ground-based daily 4-km SWE product (*Broxton et al., 2019*). Catchment-averaged SWE was computed at daily timescale from WY 1982 to WY 2008.

[25] The UA SWE product was selected due to its strong agreement with the CONUS 1-km SNODAS product (Barrett, 2003) and its improved performance relative to other gridded SWE datasets (e.g., SSMI/S and GlobSnow-2) across various land cover and snow conditions (*Cho et al., 2020*). It has also been used to guide development of a higher-resolution (800 m) SWE dataset (*Broxton et al., 2024*). Further details on data development and evaluation can be found in *Zeng et al. (2018)*.

## 2.2 Regional Classification and Spatial Masks

[26] To support analysis, we define several regional classifications and spatial masks, including geographic regions, snow regimes, forest cover, and climate zones. Key information is summarized in **Table 1**, with details in Sections 2.2.1–2.2.4.

### 2.2.1 Geographical Masking Based on Selected U.S. Topographic Regions

[27] Overall, of the 513 selected catchments, 326 lie in the eastern CONUS and 187 in the western CONUS, as delineated by the 100$^{th}$ meridian. To facilitate analysis and discussion, we pay special attention to five regions (***Figure S1***) chosen for their geographic diversity and hydrologic significance, particularly in relation to snow accumulation and melt dynamics. Specifically these are the *Appalachian Mountains* (AM; 103 catchments; purple), *Rocky Mountains* (RM; 54 catchments; red), *Colorado River Catchment* (CRB; 34 catchments; blue), *Sierra Nevada* (SN; 10 catchments; cyan), and *Cascades* (CC; 33 catchments; dark green). A catchment was considered to fall within a given region (***Figure 1a***) if it intersected any portion of that region's boundary as defined by the shapefile. Because these regional boundaries are not delimited by catchment boundaries, the regional labels are not necessarily mutually exclusive; in particular, 20 catchments are classified as belonging to both the CRB and RM regions.

[28] These five regions exhibit diverse elevation patterns (***Figure 1b***), with only the catchments in the AM and CC regions located at low elevations (average elevations of 538.54 m and 933.30 m, respectively, which are both below 1500 m). The average elevations of catchments in the CRB (2338.40 m) and SN (1988.79 m) are in a middle range ($\geq$ 1500 and $<$ 2500 m), while the average elevations of catchments in the RM (2530.35 m) are in a high range ($\geq$ 2500 m). These regions exhibit complex interactions between snow cover and forested landscapes, influencing the timing and magnitude of snowmelt and runoff.

### 2.2.2 Snowy Mask

[29] Snow processes are represented in many conceptual hydrologic models and occur in a substantial portion of CONUS catchments (*Bergström, 1975; Wi et al., 2017; Nearing et al., 2020b; Feigl et al., 2020*). To define snowy and non-snowy catchments, we evaluated several snow signature metrics proposed by *Zeng et al. (2018),* including median annual maximum SWE, April 1 SWE, and the number of snowy days over WY1982–2008.

[30] Median annual maximum SWE effectively delineates non-snowy catchments across much of the CONUS (**Figure S2a,b**), but fails to identify non-snowy catchments along the northwestern coast. For these catchments, we instead used snow fraction from the CAMELS dataset (**Figure S2c)**, classifying catchments with less than 5% snow cover as non-snowy.

[31] Combining both criteria, snowy catchments were defined as those with median annual maximum SWE greater than 3 mm (from UA SWE) and snow fraction greater than 5% (from CAMELS). Of the 513 catchments used in this study, 360 were classified as snowy and 153 as non-snowy (***Figure 1c***). Additional details are provided in **Figure S2**.

### 2.2.3 Forest Mask

[31] Forest cover influences snow accumulation and ablation at the catchment scale (*Broxton et al., 2015*). To examine whether broad forest cover is associated with model behavior, catchments were classified into two categories: *"forest-covered"* and *"open."*

[32] A forest mask was constructed for the 513 selected CAMELS catchments (**Figure 1d**). Following *Broxton et al. (2014)*, a catchment was classified as forest-covered if more than 50% of the 1-km MODIS land cover pixels within the catchment were categorized as evergreen needleleaf, evergreen broadleaf, deciduous broadleaf, mixed forest, or woody savanna (***Figure S3a,b***). Consistency was verified by comparing forest fractions derived from MODIS data with CAMELS attributes, which showed similar spatial patterns (***Figure S3c,d***). Using this approach, 282 catchments were classified as forest-covered and 231 as predominantly open.

#### 2.2.4 Climate Region Masks

[32] Climate zones for each of the 513 catchments were classified using the 1-km Köppen–Geiger maps (*Beck et al., 2023*). For each map period, namely 1961–1990 (***Figure S4a***) and 1991–2000 (***Figure S4b***), the catchment-level climate class was first assigned by majority rule, i.e., as the climate class occupying the largest fraction of the catchment area. The two period-specific catchment classifications were then combined to derive the final classification (***Figure S4c***). When the two period-specific classifications differed, the final class was determined by weighting each classification by the number of simulation days falling within the corresponding period.

[33] The resulting main and total climate classifications for the 513 CAMELS catchments are shown in **Figure 2a** and **Figure 2b**, respectively. Four of the five main climate classes (excluding Tropical) and 15 of the 30 subclasses are represented. The main classification exhibits clear regional patterns, while the total classification shows similarity to the spatial distribution of the aridity index (**Figure 2c**).

### 2.3 Hydrological Models Based on The Mass-Conserving-Perceptron (MCP)

#### 2.3.1 The Mass-Conserving-Perceptron (MCP)

[34] The mass-conserving perceptron (MCP; **Figure 3a**) is a machine-learning-based, physically interpretable computational unit introduced by *Wang & Gupta (2024a)*. It is structurally analogous to a node in a gated recurrent neural network but differs by explicitly enforcing mass conservation at the nodal level.

[35] Let $X_t$ denote the nodal water-storage state at time t (e.g., conceptual soil-moisture storage). We use $X_t$, rather than $SM_t$, to emphasize that this state represents conceptual soil-water storage rather than directly observed soil moisture. The $O_t$ and $L_t$ denote observed and unobserved outgoing fluxes (e.g., streamflow and evapotranspiration), and $U_t$ denote the incoming flux (e.g., precipitation). Mass conservation is enforced as: $X_{t+1} = X_t - O_t - L_t + U_t$.

[36] The MCP incorporates recurrence to represent system memory and allows gating functions governing process dynamics to be learned from data. It also supports representation of unobserved mass exchanges within the system.

[37] In this study, we use MCP units to construct catchment-scale models (hereafter referred to as SOIL-MCP) that simulate conceptual soil-water moisture storage and flux dynamics through directed-graph architectures of varying complexity (*Wang and Gupta, 2024b; Wang and Gupta 2025*). Additional architectural and mathematical details are provided in **Appendix A1**.

#### 2.3.2 The SNOW Mass-Conserving-Perceptron (SNOWMCP)

[38] The SNOW mass-conserving perceptron (SNOWMCP; **Figure 3b**) is introduced to represent snowpack accumulation and melt processes. The discrete-time update equation for snow water equivalent (SWE) is defined as $SWE_{t+1} = SWE_t - O_t^s - L_t^s + U_t^s$ where $SWE_t$ denotes the nodal state at time $t$, $O_t^s$ and $L_t^s$ represent observed and unobserved outgoing fluxes (e.g., snowmelt and sublimation), and $U_t^s$ denotes the incoming flux (e.g., snowfall). Additional architectural and mathematical details are provided in **Appendix A2**.

[39] The internal state of SNOWMCP, which is physically interpretable as SWE, is first pre-trained against the University of Arizona (UA) SWE product to constrain snow storage dynamics. Subsequently, when SNOWMCP

is embedded within the precipitation–storage–runoff (PSR) framework, its parameters are initialized from the pre-trained model. The system is then fine-tuned by training the combined fluxes—snowmelt generated by SNOWMCP and precipitation bypass—against observed streamflow, ensuring consistency between internal snow dynamics and catchment-scale discharge.

#### 2.3.3 MCP-based Hydrological Models Considered for this CONUS-Wide Modeling Study

[40] Approximately 70% of the catchments considered in this study are classified as snow-dominated, where both (i) snow accumulation and melt and (ii) soil moisture storage dynamics play essential roles. Accordingly, a minimal viable hydrological representation for such systems requires at least two interacting state variables.

[41] These two modules are coupled to construct HYDRO-MCP models. We consider two coupling configurations (***Figure 3***), which differ in how snowmelt generated by SNOW-MCP is routed through the system.

- In the first configuration (***Figure 3c***), the snowmelt flux $O_t^{snow}$ generated by SNOW-MCP at each timestep is treated as an additional input to the SOIL-MCP module by adding it to the rainfall flux $U_t^R$. In other words, the total input ($I_t$) to the SOIL-MCP module is $I_t = O_t^{snow} + U_t^R$. The catchment outlet streamflow ($O_t$) is then equivalent to the streamflow $O_t^{soil}$ generated by the SOIL-MCP module.
- In the second configuration (***Figure 3d***), the snowmelt flux $O_t^{snow}$ bypasses the SOIL-MCP module entirely and is routed directly to the catchment outlet. In this case, the SOIL-MCP module receives only the rainfall flux $U_t^R$ as its input $I_t$, and the downstream streamflow $O_t$ is computed as the sum $O_t^{soil} + O_t^{snow}$ of the outputs from the SOIL-MCP and SNOW-MCP modules.

### 2.4 Procedures Used for Model Training and Assessment

[42] Following (*Wang & Gupta, 2024a,b; Wang & Gupta 2025*), we adopt the robust data allocation method developed by *Zheng et al. (2022)* that partitions the data ($\mathcal{D}$) into *training* ($\mathcal{D}_{train}$), *selection* ($\mathcal{D}_{select}$), and *testing* ($\mathcal{D}_{test}$), in the ratio of 2:1:1, while maintaining distributional consistency of the streamflow records (*Chen et al., 2022; Shen et al., 2022; Maier et al., 2023*).

[43] Model training is based on the Kling–Gupta Efficiency ($KGE$; *Gupta et al., 2009*). Performance is evaluated using its skill-score version $KGE_{ss}$ (*Knoben et al., 2019b; Khatami et al., 2020*). Model selection is guided by the Akaike Information Criterion (AIC; *Akaike, 1974*), which helps to balance model fit and complexity, thereby favoring parsimonious models that generalize better to unseen data.

[44] Details on the data-splitting algorithm, evaluation metrics, model selection, and training procedure are provided in **Appendices B–E** including **Table S1**.

## 3. Experimental Results

[45] We report results from a progression of model configurations of increasing complexity.

[46] Section 3.1 evaluates an over-simplified single-state hydrological model, $HMCP(1, B)$, (where "$1$" denotes a single state variable and "$B$" indicates the basic formulation), using only the SOIL-MCP module to represent PSR dynamics across 513 CONUS catchments, without distinguishing between snow and soil-water processes. The objective is to assess the best achievable performance under a single-state representation, across different hypotheses regarding catchment gating processes.

[47] In Section 3.2, we evaluate two *"process-augmented"* variants of the single-state model: $HMCP(1, MR$ (where "$MR$" denotes mass relaxation, allowing subsurface water exchange), and $HMCP(1, AL)$ (where "$AL$" denotes augmented loss, linking evapotranspiration to both atmospheric demand and soil moisture availability).

[48] In Section 3.3, we evaluate single-state SNOW-MCP models for representing PSR dynamics across the CONUS. We consider a basic formulation, $SNOWMCP(1, B)$ (where "B" denotes a temperature-based rain–snow partitioning), and two more flexible variants, $SNOWMCP(1, G)$ and $SNOWMCP(1, C)$ (where "G" for

"*Generic*" and "C" for "*Complex*" denote increasing levels of complexity), in which the gating functions additionally exploit other meteorological variables such as vapor pressure deficit, solar radiation, and daily temperature extremes.

[49] In Section 3.4, we evaluate more complex two-state models that explicitly distinguish between snow and soil-water dynamics. We consider both series and parallel configurations of the states, $HYDROMCP(2,S)$ and $HYDROMCP(2,P)$ (where "S" and "P" denote *"series"* and *"parallel"* structures), and examine the effects of different training strategies and augmented formulations that incorporate additional meteorological inputs.

[50] In Section 3.5, we present streamflow and SWE time series for two representative catchments, providing a focused diagnostic analysis of the MCP-based models.

[51] In Section 3.6, we examine the geographical distribution of *"optimal-parsimonious"* model architectures across the CONUS, based on the tradeoff between predictive performance and model complexity. This analysis addresses the central objective of this study.

[52] Finally, in Section 3.7, we compare the performance of the aforementioned model architectures against several ML benchmarks.

[53] In **Table 2**, we summarize the $KGE_{ss}$ performance values across different percentiles for the models described in Sections 3.1–3.4, while **Tables S2–S4** present the three individual components of the KGE. **Table 3** reports $KGE_{ss}$, root mean square error ($RMSE$) and mean absolute error ($MAE$) values for the two catchments selected in Section 3.5 to compare the performance of various MCP-based models across diverse hydrologic water years. Further, **Table S5** compares streamflow and SWE skill metrics for the single-state SNOWMCP model under three training configurations—streamflow-only, SWE-only, and joint streamflow–SWE training—to evaluate its performance for both streamflow and SWE relative to the two-state HYDROMCP model.

[54] We also summarize the basin counts and regional geographic distribution of $KGE_{ss}$ metrics and compare model selection outcomes based on the AIC criterion and the maximum KGE metric for model architectures in **Section 3.5** (**Table 3**) and benchmarks in Section 3.6 (**Table 4**). Further, the number of basins, $KGE_{ss}$ skills selected by the AIC criterion in individual experiments with SOILMCP, SNOWMCP, and HYDRO-MCP are reported in **Table S6** and **Table S7**, and the results selected based on KGE are in **Table S8** and **Table S9**. We refer readers to this detailed information if further examination is necessary.

## 3.1 Preliminary Baseline: Testing a Simple MCP-Based Model with a Single State Variable (HMCP)

### 3.1.1 Overall Assessment

[55] **Figure 4a** summarizes the geographical distribution of testing period $KGE_{ss}$ performance for the single-cell-state "Basic" $HMCP(1,B)$ model across the CONUS. Median performance across the training, selection, and testing periods is $0.74$ / $0.74$ / $0.72$, indicating consistent performance. Surprisingly, this single-state model shows strong continental-scale results, with 88% of catchments achieving $KGE_{ss} > 0.50$.

[56] Since this model does not include a dedicated component for tracking snowpack dynamics, its performance is lower in snowy catchments ($0.71$) than in non-snowy ones ($0.79$). Performance is also higher in the east ($0.75$) than in the west ($0.64$), where snow processes play a more important role. Consistently, performance decreases from temperate to cold and arid regions (**Figure 2**).

[57] Forested catchments show a higher average performance (0.76) than non-forested/open catchments (0.69). This suggests that broad forest-cover classification alone is unlikely to explain the dominant spatial pattern of single-state model performance at the continental scale, although it does not rule out more specific vegetation-related controls.

[58] **Figures 4b-4d** show the spatial distribution of the three KGE components ($r^{KGE}$, $\alpha^{KGE}$, and $\beta^{KGE}$). While variability and mass balance are generally well reproduced ($\alpha^{KGE}$, and $\beta^{KGE}$ close to 1), correlation is lower

(median $r^{KGE}$ ≈ 0.64), indicating difficulty in capturing hydrograph timing and shape. Thus, although the $HMCP(1,B)$ model reproduces long-term variability and mass balance, it struggles with short-term streamflow dynamics.

### 3.1.2 Performance across the Eastern CONUS

[59] Looking more closely at $KGE_{ss}$ performance in the eastern CONUS, higher skill is generally observed in the south, with performance decreasing northward (**Figure 4a**). Catchments in the *Appalachian Mountains* (AM) show lower correlation and larger variability errors than the eastern regional average (**Figures 4b–c**). These spatial patterns are broadly consistent with climate-related controls on model performance, particularly the transition from temperate to colder regions (**Figure 2a**). They also reflect a key limitation of the basic $HMCP(1,B)$ architecture: the absence of an explicit snow module, which makes it more difficult to represent streamflow dynamics in basins where snow accumulation and melt processes become increasingly important.

### 3.1.3 Performance across the Western CONUS

[60] In contrast, performance across the western US is relatively poor, particularly in mountainous regions, with median $KGE_{ss}$ around 0.55 (compared to 0.72 nationwide). This likely reflects the dominance of high-elevation catchments where snow processes are more important. In terms of flow variability and mass balance ($\alpha^{KGE}$, and $\beta^{KGE}$), a general negative bias is observed (**Figures 4c–4d**). Performance is particularly low in the *Rocky Mountains* (RM), with median $KGE_{ss}$ around 0.53, and even lower in the *Colorado River Basin* (CRB) (0.51). This pattern highlights the need for improved representation of snow dynamics in high-elevation catchments.

[61] Along the Pacific West Coast, performance improves northward (**Figure 4a**). In forest regions, performance is higher in the *Cascades* (CC; 0.80) than in the *Sierra Nevada* (SN; 0.56). The lower performance in the SN likely reflects the combined effects of arid conditions and snow processes that are not explicitly represented in the single-state $HMCP(1,B)$ model. In the northwest region, performance is relatively high (median $KGE_{ss}$ ≈ 0.83) and further improves for non-snowy catchments (≈0.92). This reinforces the importance of explicitly representing snowpack dynamics. At the same time, strong performance can still be achieved in snowy regions when conditions are sufficiently dry, suggesting that soil-moisture dynamics may be less critical in such cases.

## 3.2 Augmented Baseline Results: Testing Alternative Process Hypotheses

### 3.2.1 "Relaxing" the Requirement for Mass Conservation

[62] In the $HMCP(1,B)$ model, mass conservation is strictly enforced at the catchment scale (Section 2.3.1). However, this assumption may be incomplete, as water can enter or leave the system through unobserved subsurface exchanges (e.g., groundwater interactions). To account for this, we follow Wang & Gupta (2024a) and introduce a cell-state–dependent mass-relaxation term within the $HMCP(1)$ framework, referred to here as $HMCP(1,MR)$. Details of the formulation are provided in **Appendix A1**.

[63] Results are shown in **Figures 5** and **6**. Overall, $HMCP(1,MR)$ outperforms $HMCP(1,B)$ in 71% of catchments, with median $KGE_{ss}$ increasing from 0.72 to 0.76. Improvements are spatially widespread and are most pronounced in the mountainous western US and the northwest (**Figures 5e–h**). The largest gains occur in linear correlation ($r^{KGE}$), indicating improved representation of hydrograph timing and shape, while changes in variability ($\alpha^{KGE}$) and mass balance ($\beta^{KGE}$) are more moderate.

[64] **Figure 6a** shows the spatial distribution of threshold values learned for the mass-relaxation gate, indicating regions of net water gain or loss. The gaining ratio (GR; **Figures 6b–c**), defined as the fraction of timesteps when the cell state is below the threshold, reveals that many catchments, particularly snowy ones, predominantly gain water over time. This suggests that the mass-relaxation gate compensates for the absence of explicit snow processes by introducing additional water into the system. This behavior is especially pronounced in the mountainous western U.S.

[65] However, the influence of forest cover is less clear. In forest-dominated regions, the gaining ratio varies substantially across basins, with higher values in the Cascades (CC) likely reflecting strong snow influence. In contrast, lower values in the Appalachian Mountains (AM) and Sierra Nevada (SN) may indicate missing processes in the $HMCP(1, MR)$ architecture, such as vegetation–snow interactions (see **Figure S6**).

### 3.2.2 Augmenting the Loss Gate

[66] In the $HMCP(1)$ models reported above, evapotranspiration loss is parameterized as a function of potential evapotranspiration ( $PET_t$ ), representing atmospheric demand. Here, we test an augmented formulation in which loss depends on both atmospheric demand and available soil moisture ($X$), referred to as $HMCP(1, AL)$ (see **Appendix A1**).

[67] Results are shown in **Figures S5a–d**, with differences relative to $HMCP(1, B)$) in **Figures S5e–h**. Overall, performance improvements are modest, with median $KGE$ increasing from $0.72$ to $0.74$ and $72\%$ of catchments showing improvement. These gains are relatively uniform across catchment types, with no clear association with hydroclimatic or land-cover factors.

[68] Similar limited improvements are observed when imposing $PET_t$ as an upper bound on evapotranspiration loss (**Figure E2**).

## 3.3 MCP-Based Modeling of Snow Accumulation and Melt (SNOWMCP)

[69] In the previous sections, no distinction was made between soil moisture and snowpack dynamics, with a single MCP state variable used to represent total water storage. Here, we explicitly model snow accumulation and melt using a class of single-state models, denoted $SNOWMCP(1)$, in which the state variable represents snow water equivalent (SWE). The architecture is analogous to $HMCP(1, B)$, but adapted to represent snow processes (see **Appendix A2**).

[70] It is important to note that the SNOWMCP-based approaches described below were applied only to the 502 catchments for which the UA SWE product contained non-zero values during the simulation period. The remaining 11 catchments had zero SWE for all time steps from WY1982 to WY2008 and were therefore excluded from SWE-based snow-module pretraining and evaluation. For catchments classified as non-snowy but with non-zero UA SWE values, the same SWE-based procedure was applied, although the snow-related signal was generally weak because SWE was near zero for most days.

### 3.3.1 Basic Snow Model Formulation

[71] The first model, $SNOWMCP(1, B)$ (where *'B'* denotes the basic formulation), assumes that the output (snowmelt) and loss gates depend on the current snow water equivalent ($SWE_t$) and mean temperature ($T_t$), while the input (rain–snow partition) gate depends solely on temperature. We evaluate this formulation by training the model against "*observed*" SWE from the UA SWE dataset, such that the simulated state closely matches $SWE_t$ at each catchment. Note that the model is trained to reproduce SWE, not streamflow.

[72] **Figure 7a** shows that the median testing period $KGE_{ss}$ reaches $0.95$ for snowy catchments but is substantially lower for non-snowy ones ($\sim 0.52$), with an overall median of $0.93$. These results indicate that the simple temperature-based formulation is highly effective for representing SWE dynamics, particularly in snow-dominated regions.

### 3.3.2 Augmented Snow Model Formulation

[73] We next examine whether augmenting the model with additional meteorological variables beyond temperature improves $SNOWMCP(1)$ performance.

[74] The first augmented formulation, $SNOWMCP(1, G)$ (where "*G*" denotes a more generic representation), incorporates additional meteorological inputs into the gating functions, including vapor pressure deficit (for the rain–snow partition and loss gates) and solar radiation (for the melt gate), to account for humidity and energy

controls on snow processes. The second formulation, $SNOWMCP(1, C)$ where "*C*" denotes increased complexity), further extends $SNOWMCP(1, G)$ by incorporating daily minimum and maximum temperatures to capture diurnal variability in snow processes.

[75] Both augmented models were trained and evaluated against the UA SWE dataset (**Figures 7b-c**). Overall, increasing model complexity leads to only marginal changes in performance, with slight improvement in snowy catchments and slight degradation in non-snowy ones, resulting in nearly identical median performance across all basins. These results indicate that the basic $SNOWMCP(1, B)$ architecture already provides excellent performance in snowy catchments using only temperature and SWE. Additional inputs offer limited benefit in this standalone setting, but may become more important when the $SNOWMCP$ module is coupled with a $SOILMCP$ module, as explored in the next section.

### 3.3.3 Basic Snow Model Formulation Used for Streamflow Simulation

[76] We next evaluate the basic $SNOWMCP(1, B)$ architecture for streamflow simulation by training it directly against observed discharge, *without* using SWE as the target. This configuration, denoted $SNOWMCP(1, BQ)$, retains a state variable that primarily represents snow storage (SWE). The underlying assumption is that, in snow-dominated catchments, streamflow dynamics can be largely explained by snow processes alone, without explicitly modeling soil moisture.

[77] From **Figures 8a-d** we see that the geographical distribution of testing period performance is similar to that obtained using the baseline $HMCP(1)$ model (see Section 3.1). However, the median testing period $KGE_{ss}$ over all 513 catchments has improved significantly (from $0.72 \rightarrow 0.78$), consistent with the fact that 70% of the catchments are classified as "*snowy*", for which the process of rain-snow partitioning is arguably important to the representation of catchment PSR dynamics. Clearly (and as should be expected), it is important to explicitly include a representation of snow accumulation and depletion dynamics in any model used to simulate catchment-scale PSR dynamics across the CONUS.

[78] **Figures 8a–d** show that the spatial pattern of performance is similar to that of the baseline $HMCP(1)$ model (Section 3.1), but with a substantial increase in median testing period $KGE_{ss}$ ($0.72 \rightarrow 0.78$). This improvement highlights the importance of explicitly representing snow accumulation and melt processes for streamflow simulation across the CONUS.

### 3.3.4 Augmented Snow Model Formulation Used for Streamflow Simulation

[79] We finally evaluate the augmented $SNOWMCP(1, G)$ and $SNOWMCP(1, C)$ architectures for streamflow simulation by training them directly against observed discharge, without using SWE as the target. These configurations are denoted $SNOWMCP(1, GQ)$ and $SNOWMCP(1, CQ)$.

[80] **Figure 9** summarizes the performance of the augmented snow models. Overall, augmenting the gating functions leads to a modest improvement in streamflow prediction (median $KGE_{ss}$ increases from $0.78$ to $0.80$), with more pronounced gains at lower performance percentiles. Notably, these improvements are primarily observed in non-snowy catchments, while only minor gains are seen in snowy regions (median $KGE_{ss}$ increases from $0.80$ to $0.81$). This suggests that additional meteorological inputs help capture processes that are more relevant outside snow-dominated regimes.

[81] **Figure 9b-e** show the spatial distribution of model performance across the CONUS for the $SNOWMCP(1, GQ)$ architecture. High performance is observed in high-elevation mountainous catchments in the western U.S. and humid coastal regions in the northwest, including the Cascades (CC). Lower performance occurs near the 100th meridian, in the southern Appalachian Mountains (AM), and across the southwestern U.S. These regions are generally characterized by lower correlation, underestimation of flow variability, and bias in mass balance.

[82] Differences relative to the baseline $SNOWMCP(1, BQ)$ model are shown in **Figure S7**. Improvements in snowy catchments are primarily observed in the north-central U.S. (e.g., North Dakota) and the eastern U.S.,

with only marginal gains in the mountainous western regions and along the northwest coast. These regions are characterized by relatively large snow storage and late-season melt (e.g., high maximum SWE, substantial April 1 SWE, and delayed melt timing; see **Figure S6**). Notably, little to no improvement is found in parts of the Rocky Mountains (RM), where persistent snowpack conditions suggest that temperature-based gating is already sufficient. Further augmentation with daily minimum and maximum temperatures ($SNOWMCP(1, CQ)$) yields additional gains in some snowy catchments (**Figure S8**), indicating potential benefits from refining gating formulations.

## 3.4 Coupled Two-State-Variable Hydrologic Models (HYDROMCP)

### 3.4.1 Basic Coupled Model Formulation

[83] Building on the single-state MCP models, we now introduce a two-state hydrological model, $HYDROMCP$, which explicitly represents snow and soil moisture dynamics through coupled $SNOWMCP$ and $SOILMCP$ modules. This separation allows the model to capture distinct snow-driven and liquid water–driven processes that jointly control streamflow dynamics.

[84] We evaluate two coupling strategies, denoted $HYDROMCP(2, S)$ and $HYDROMCP(2, P)$ (**Figures 3c–d**). In the series configuration (S), snowmelt contributes to soil storage before generating streamflow, whereas in the parallel configuration (P), snowmelt bypasses the soil module and directly contributes to streamflow. Model training is performed by *"fine-tuning"* from the best-performing configurations identified in previous stages.

[85] To evaluate the coupled models, we consider three training strategies that differ in how the snow and soil modules are optimized. In the first, only the soil module is trained while the snow module remains fixed following pre-training on SWE. In the second, both modules are jointly trained using streamflow as the sole target. In the third, both modules are jointly trained using a combined objective that equally weights streamflow and SWE performance ($KGE_Q$ and $KGE_S$).

[86] $HYDROMCP(2, B)$ denotes the basic two-state hydrologic architecture obtained by coupling $HMCP(1, B)$ with $SNOWMCP(1, B)$. Within this architecture, we tested two routing assumptions: serial routing, denoted as $HYDROMCP(2, S)$, and parallel bypass, denoted as $HYDROMCP(2, P)$. Thus, the $HYDROMCP(2, B)$ results reported in **Figure 10** represent, for each basin, the better-performing configuration selected from these two routing assumptions (see also **Figures S9–S11**).

[87] Not all training runs converged successfully; results are therefore reported for the 474 catchments with successful training. In general, the parallel configuration ($HYDROMCP(2, P)$) exhibits greater training stability than the series configuration, consistent with prior findings that parallel structures are often more effective for representing streamflow dynamics.

[88] Overall, fixing the snow module during training (Strategy 1) resulted in lower performance. The best results were obtained when both modules were jointly trained using streamflow as the sole target (Strategy 2), whereas adding SWE as a second target (Strategy 3) did not provide further benefit. This suggests that, after SWE-based pretraining, streamflow contains sufficient information to tune the coupled system for discharge simulation, while the additional SWE target may introduce a trade-off between internal snow-state matching and downstream flow optimization. Strategy 2 performed best across the majority of catchments, with the parallel configuration generally outperforming the series formulation.

[89] **Figure 10** shows the spatial distribution of the best $KGE_{ss}$ performance across the 474 successfully trained catchments. Overall, the HYDROMCP model achieves consistently strong performance (median $KGE_{ss} \approx 0.79$), with comparable skill across snowy and non-snowy, as well as forested and non-forested catchments.

[90] Examining poorly performing catchments ($KGE_{ss} \approx 0.0 - 0.1$; **Figure 10a**), we find that errors are primarily associated with low correlation ($0.2 - 0.6$; **Figure 10b**), overestimation of flow variability (**Figure 10c**), and positive bias in mass balance (**Figure 10d**). These patterns point to deficiencies in the representation of snowpack dynamics.

[91] Training failures are primarily observed in the eastern CONUS, including Florida and parts of the Appalachian Mountains (AM), and are more common in forested regions (**Figure S9**). This pattern suggests that the current two-state $HYDROMCP$ formulation may be insufficient to represent dominant processes in vegetation-influenced catchments. This limitation indicates the need to further expand the conceptual MCP architecture toward a more process-aware representation that incorporates soil, canopy, and vegetation dynamics in future model development (*Wang et al., 2024; Feng & Konings, 2025; Jiang et al., 2025; Zhao et al., 2025; Farmani et al., 2026*). Additional details are provided in **Text S1** of the Supplementary Materials.

### 3.4.2 Augmented Coupled Model Formulation

[92] In this section, we examine the effect of augmenting gating functions within the $HYDROMCP$ model. Providing the $SOILMCP$ loss gate with both atmospheric demand (PET) and water availability results in mixed performance outcomes, with improvements at some locations and degradation at others (see **Text S2**). However, this modification increases the number of successfully trained catchments (from 474 to 484), with many of the improvements occurring in the eastern United States (**Figure S12**).

[93] We next evaluate $HYDROMCP$ models using augmented snow modules ( $SNOWMCP(1, G)$ and $SNOWMCP(1, C)$) in place of the baseline formulation. In contrast to the mixed results observed for soil-gate augmentation, these configurations lead to performance improvements as summarized in **Figure S13**.

[94] **Figure S13a** shows that augmenting the snow gating functions in $HYDROMCP$ leads to a consistent, though modest, improvement in overall performance, with median $KGE_{ss}$ increasing from $0.79$ to $0.81$. These gains are broadly consistent across different catchment types, including snowy and non-snowy, as well as forested and non-forested regions. The largest improvements are observed in parts of the eastern CONUS, particularly within the Appalachian region. Overall, performance gains are primarily associated with improved linear correlation, indicating better representation of streamflow timing.

[95] The spatial patterns in **Figure S13 (b–e),** together with the differences relative to the baseline $HYDROMCP(2, B)$ model **(Figure S13f–i)**, show that the poorest performance is concentrated in the eastern CONUS. These catchments ($KGE_{ss} < 0.1$) exhibit moderate correlation ($\approx 0.2 - 0.6$) but substantial positive biases (>30%) in both flow variability and mass balance. They are located in temperate to cold climate zones and are characterized by limited snow accumulation, early spring snowmelt, and relatively short snow seasons ($\sim 125$ days; **Figure S6**), indicating that transitional snow regimes remain challenging to represent. This limits the information available from streamflow-based training to uniquely identify snow-storage and melt-release dynamics, allowing snowmelt-related errors to be partially compensated by rainfall–runoff or loss components. As a result, the model can retain moderate timing skill while producing biased flow variability and mass balance.

[96] Further inclusion of daily minimum and maximum temperatures in the gating functions leads to additional performance gains (**Figures S13j–q**). These improvements are particularly pronounced in snowy catchments, with a substantial number showing notable increases in $KGE_{ss}$ (e.g., $\geq 0.05$ and $\geq 0.3$; **Figure S14b**).

## 3.5 Illustrative Time Series Diagnostics in Selected Basins

[97] We present streamflow and SWE time series for two representative catchments to support a focused diagnostic analysis. Basin 02296500 (Charlie Creek near Gardner, Florida) represents a non-snowy basin with SWE equal to zero throughout the simulation period, whereas basin 13023000 (Greys River above Reservoir near Alpine, Wyoming) represents a snowy basin with seasonal SWE dynamics The evaluated models include $HMCP(1, B)$, $SNOWMCP(1, B)$, $SNOWMCP(1, BQ)$, and $HYDROMCP(2, B)$, with $NMCP_{wsh}(DS, 5)$ and $LSTM(5)$ used as benchmark models.

[98] For the two selected catchments, representative time series were chosen based on local flow conditions. The wettest, median, and driest water years were generally identified using annual mean streamflow. The only exception was basin 02296500, for which the driest year was selected as the year with the second-smallest annual peak flow. This choice was made because the year with the lowest annual mean streamflow had daily flow below 1 mm throughout the year, making it less representative for evaluating model behavior.

[99] **Figure 11a–c** shows the streamflow time series for basin 02296500 during the three selected water years. This non-snowy basin achieved the highest overall testing $KGE_{ss}$ (0.92) among all evaluated catchments and is therefore used as a representative case to illustrate strong model performance under relatively pristine hydrologic conditions. Because this basin has no persistent snowpack according to the UA SWE product, with SWE equal to zero at all time points, the evaluation focuses only on the streamflow hydrograph. Overall, as summarized in **Table 3**, $HMCP(1, B)$ outperforms $NMCP_{wsh}(DS, 5)$ and $LSTM(5)$ in terms of $KGE_{ss}$ during the dry year, while maintaining comparable performance during the median and wet years. Similar conclusions are obtained from root mean square error (RMSE) and mean absolute error (MAE), suggesting broadly comparable predictive accuracy across models.

[100] **Figure 11d–i** shows the streamflow time series for basin 13023000 during the three selected water years. For clarity, the MCP model results are presented separately from the benchmark results. $HMCP(1, B)$ shows poorer streamflow performance than $SNOWMCP(1, BQ)$ and $HYDROMCP(2, B)$ as it consistently fails to simulate streamflow during the dry year (**Figure 11d**) and exhibits underestimation during the median and wet years (**Figures 11fh**). Both $SNOWMCP(1, BQ)$ and $HYDROMCP(2, B)$ still show room for improvement relative to the two benchmark models across all flow-regime years (**Figure 11egi**). Although $HYDROMCP(2, B)$ does not show a clear advantage over $SNOWMCP(1, BQ)$ in terms of KGEss, RMSE, and MAE for streamflow prediction, it maintains higher skill in reproducing SWE dynamics approaching the performance of $SNOWMCP(1, B)$, whose internal state is directly trained against SWE (**Figures 11j–l**). This case demonstrates that stagewise model development can improve both predictive accuracy and conceptual interpretability.

[101] It is important to note that the SWE state in the SNOWMCP tank becomes an *"informational"* state when the SNOWMCP architecture is used to predict streamflow and is therefore trained only against streamflow. Our analysis shows that, under this setting, $SNOWMCP(1, BQ)$ yields negative median KGEss values for SWE in snowy basins. The additional experiment with $SNOWMCP(1, BQ)^*$, in which the model was trained simultaneously against SWE and streamflow using an equally weighted KGE objective, confirms an unavoidable tradeoff between maintaining SWE accuracy and purely optimizing streamflow prediction accuracy (see **Figure S15 & Table S5**). In contrast, $HYDROMCP(2, B)$ can still maintain some skill in reproducing SWE dynamics, although this appears to be more likely in snow-dominant catchments as the number of snowy days and annual maximum SWE increase (**Figures S16ab**).

[102] Overall, the MCP-based models used in this study remain conceptual hydrologic model representations, but differ from traditional conceptual models by replacing time-invariant parameters with time-varying gating functions. The MCP computational unit can therefore be viewed as an extension of conceptual storage models, in which storage, output, and loss pathways are explicitly organized under a mass-conserving structure. Accordingly, the coupled HMCP–SNOWMCP framework used in this study should be interpreted as a process-inspired conceptual hydrologic model, rather than a fully process-resolving physical model.

[103] This distinction is important because structural interpretability does not necessarily guarantee that all internal states are physically realistic. For instance, when the SNOWMCP module is trained only against streamflow, its SWE-related state may function as an informational state that supports streamflow prediction, but may not accurately reproduce observed SWE dynamics. More direct hydrologic interpretation of this internal state requires additional constraints, such as training against SWE or jointly against SWE and streamflow. We therefore recognize that further efforts are needed to strengthen the physical consistency of MCP-based models and to move them further away from being characterized as purely data-driven models (*Mendoza et al., 2015*). Potential directions include training against multiple output fluxes, incorporating additional process constraints, or adding physical processes explicitly into the model structure (*Farmani et al., 2026*). Given that the current study focuses primarily on streamflow modeling, we leave these developments for future work.

### 3.6 Selection of Optimal Model Architectures

[104] The previous sections evaluated three MCP-based model classes across the CONUS: single-state $HMCP$, snow-focused $SNOWMCP$, and two-state $HYDROMCP$. The results show that no single architecture

consistently provides the best performance across all catchments. This section identifies the hydroclimatic conditions under which different model architectures provide relatively better performance, while balancing predictive accuracy and structural complexity. To this end, we adopt a simplicity bias, favoring more parsimonious models when performance differences are small (*Weijs & Ruddell, 2020; Wilson 2025*).

### 3.6.1 Candidate Architectures and AIC-Based Model

[105] We consider nine candidate architectures spanning the HMCP, SNOWMCP, and HYDROMCP model classes.

1) For the single-state $HMCP$ models, we include a basic 7-parameter formulation ($HMCP(1,B)$), an 8-parameter loss-augmented variant ($HMCP(1,AL)$), and a 10-parameter mass-relaxation version ($HMCP(1,MR)$).
2) For the single-state $SNOWMCP$ models, we consider an 11-parameter basic model ($SNOWMCP(1,BQ)$), a 14-parameter augmented version ($SNOWMCP(1,GQ)$) and a 20-parameter formulation ($SNOWMCP(1,CQ)$) with additional temperature information.
3) For the two-state $HYDROMCP$ models, we include an 18-parameter baseline ($HYDROMCP(2,B)$), a 21-parameter augmented variant ($HYDROMCP(2,G)$), and a 27-parameter formulation ($HYDROMCP(2,C)$) with further augmentation of the snow component. In each case, the selected version corresponds to the best-performing configuration across coupling strategies and training methods.

[106] To determine the optimal model architecture at each location, we use the Akaike Information Criterion (AIC), which balances model fit and complexity. The likelihood component is evaluated using a set of candidate probability distributions (six in total; Strupczewski et al., 2001), allowing flexibility in representing different residual behaviors and reducing the risk of distributional misspecification (see **Appendix D** for details). For each case, the distribution yielding the highest likelihood is used in the AIC computation.

### 3.6.2 Selection Among All the Model Architectures

[107] Considering all nine model architectures, the simplest single-state $HMCP$ models are most frequently selected across the CONUS, reflecting their robustness across diverse hydroclimatic conditions (**Figure 12**). In contrast, $SNOWMCP$ models are selected less often but achieve higher performance, indicating their effectiveness in snow-dominated environments. The more complex $HYDROMCP$ models are selected only in a subset of catchments, suggesting that increased structural complexity provides limited additional benefit in most cases. Overall, these results demonstrate that no single model architecture is universally optimal, and that model performance depends strongly on hydroclimatic regime.

[108] **Figure 12a** shows the geographical distribution of selected model architectures. $HMCP$ models are selected across most of the CONUS, while $SNOWMCP$ models are largely confined to snow-influenced regions and are notably absent in parts of the northeastern U.S. and near the Great Lakes. The more complex $HYDROMCP$ models occur most frequently in the western mountainous regions and in parts of the non-snowy southeastern U.S., suggesting that additional process complexity may be required in these areas, particularly to account for soil-moisture dynamics.

[109] A more detailed comparison of model selection using KGE versus AIC, together with architecture-selection results for the HMCP, SNOWMCP, and HYDROMCP model classes, is provided in **Text S3** and **Figures S17–S18**. These results suggest that adopting a multi-criteria selection framework may better constrain necessary model complexity and help identify missing process representations across different hydroclimatic regions.

## 3.7 Comparison with Machine Learning Benchmarks

### 3.7.1 Context for the Comparison

[110] Recent studies have demonstrated that data-driven recurrent neural networks, such as LSTMs, can achieve state-of-the-art catchment-scale streamflow prediction when trained across multiple catchments (e.g., *Kratzert et al., 2018; Feng et al., 2020; Nearing et al., 2021; Kratzert et al., 2024; Nearing et al., 2024*). This enables the development of "universal" models applicable to ungauged basins. However, despite their strong predictive performance, such models provide limited insight into the geographical distribution of hydrological processes and the associated model complexity, largely due to their high dimensionality (often exceeding 64–256 states) and limited interpretability.

[111] In contrast, recent studies have shown that much simpler and interpretable models, with only a few cell states (on the order of 1–5), can achieve comparable performance (*De La Fuente et al., 2024; Wang & Gupta, 2024ab; Wang & Gupta, 2025*). This level of complexity is consistent with traditional physical-conceptual hydrological models (e.g., SACSMA, GR4J, HBV; *Burnash & Ferral, 1973; Bergström, 1992; Perrin et al., 2003*). However, such models are typically trained independently at each catchment, limiting their ability to leverage shared information across locations.

[112] A recent exception is the Regional HydroLSTM framework (*De La Fuente et al., 2025*), which enables parameter sharing through data-driven regionalization, but does not enforce mass conservation. Meanwhile, recent work on MCP-based neural networks has demonstrated that interpretable, mass-conserving architectures can achieve near-optimal performance comparable to locally trained LSTM models, while using only a small number of cell states (typically 3–5) (*Wang & Gupta, 2025*). These models are structurally consistent with conceptual rainfall–runoff models and can further benefit from sharing cell-state information across gates, analogous to information flow in LSTM architectures.

[113] In this section, we compare the nine MCP-based model architectures (Section 3.5) against selected machine learning benchmarks (MLBs), which serve as estimates of the best achievable performance from data-driven models. All models are trained independently at each catchment, so this comparison does not exploit information sharing across locations. Here, the focus is on predictive performance, in contrast to Section 3.5 where model complexity guided model selection.

### 3.7.2 Machine Learning Benchmark Models

As MLBs we use the following:

1) An eighty-five parameter physically-interpretable "*distributed-state*" MCP-based network developed and tested by (*Wang & Gupta, 2025*), that has five-cell-states *with* information sharing across gates, denoted as $NMCP_{wsh}(5, DS)$.
2) A 166-parameter single layer LSTM network having five-cell-states *with* information sharing across gates, denoted as $LSTM(5)$.

[114] Both benchmark models are provided access to the same input data used by the MCP-based architectures. However, a key distinction is that the $NMCP$ benchmark explicitly enforces mass conservation at both the nodal and network levels, whereas the $LSTM$ benchmark does not.

### 3.7.3 Benchmarking Results

[115] We compare the CONUS-wide *"optimal"* MCP-based models ($OPTMCP$) with the MLBs. The cumulative distribution functions of the performance metrics are shown in **Figure 13**, while detailed box-plot comparisons grouped by geographic region and by associated geographic, climatologic, and hydrologic classifications are provided in **Figure S19-22**.

[116] The $OPTMCP$ models (typically using one or two cell states) achieve a median testing period $KGE_{ss} = 0.85$ across all 513 catchments, comparable to the five-state mass-conserving $NMCP_{wsh}(5, DS)$ benchmark,

but slightly lower than the non-mass-conserving $LSTM(5)$ benchmark (median $KGE_{ss} = 0.89$). These results indicate that physically constrained and parsimonious MCP-based models can achieve near state-of-the-art performance, with only modest loss relative to fully data-driven approaches.

[117] A geographical comparison (**Figure 13a**) shows that the OPTMCP model outperforms the five-state mass-conserving $NMCP_{wsh}(5, DS)$ benchmark at nearly half of the catchments ($240$ out of $513$). Decomposing performance into KGE components reveals that $OPTMCP$ achieves improved flow variability (**Figure 14c**), comparable mass balance (**Figure 13d**), but slightly lower linear correlation (**Figure 13b**). These results suggest that the parsimonious MCP-based models capture hydrological variability effectively, but may sacrifice some temporal accuracy relative to more complex architectures. This trade-off reflects the impact of structural simplicity and physical constraints on model behavior.

[118] Next, we apply the AIC criterion to select among the $OPTMCP$, $NMCP_{wsh}(5, DS)$ and $LSTM(5)$ models at each location, to examine how necessary model complexity varies geographically. The resulting spatial distribution of selected models is shown in **Figure 14a**. The distribution of selected architectures is relatively balanced, with $OPTMCP$, $NMCP_{wsh}$ and $LSTM$ models selected at $27\%$, $40\%$, and $33\%$ % of catchments, respectively. These results indicate that no single model consistently outperforms the others, and that the required level of model complexity varies across hydroclimatic regimes. Model selection with complexity penalization increases the number of one- to two-state models (from 67 to $140$) while substantially reducing the number of non–mass-conserving models (from 375 to 168). Despite this shift toward simpler and physically consistent architectures, predictive performance remains high, with median $KGE_{ss} = 0.87$ compared to $0.89$ when models are selected solely based on KGE (**Figure 14b,d**). These results demonstrate that incorporating complexity penalties enables more parsimonious and physically meaningful model selection with only minimal loss in predictive accuracy, and that the required level of model complexity varies across hydroclimatic regimes.

[119] Mass-conserving architectures tend to be favored in forest-dominated regions. For example, the non–mass-conserving $LSTM(5)$ model is selected less frequently in the Appalachian Mountains (32 out of 103 basins), whereas simpler mass-conserving $OPTMCP$ models dominate in the Sierra Nevada (9 out of 10 basins). In contrast, $LSTM(5)$ is selected more often in the Cascades ($\sim 40\%$; 13 out of 33 basins), likely reflecting stronger snow influences. These patterns suggest that the importance of mass conservation and model structure varies with regional hydroclimatic conditions.

[120] The relatively low selection frequency of $OPTMCP$ models ($27\%$ of catchments) suggests that important processes are not fully captured by the simple one- to two-state representations explored in this study. In particular, the superior performance of non–mass-conserving $LSTM(5)$ models in snow-dominated basins highlights the need to further improve SNOW-MCP architectures, especially in representing coupled mass–energy processes (*Anderson 1976; Tarboton and Luce, 1996*).

[121] More broadly, these results emphasize the need for future benchmarking efforts that better balance predictive performance and structural complexity, while accounting for variations in model architecture (e.g., number of cell states and information sharing) across diverse hydroclimatic regimes.

## 4. Summary, Conclusions and Outlook

### 4.1 Summary

[122] Large-sample hydrology (LSH) emphasizes the analysis of data across diverse geo-hydro-climatological regions to gain insight into hydrological processes under varying conditions, supporting the development of more general and robust hydrological theory. In this context, physical-conceptual rainfall–runoff (CRR) models, though often viewed as *"lossy over-compressions,"* remain valuable for testing alternative structural hypotheses and advancing process understanding across locations. In contrast, *"universal-architecture"* machine learning

approaches (e.g., LSTM-based models) trained on multi-catchment datasets have become the standard for large-scale hydrological prediction, addressing the PUB problem while achieving state-of-the-art performance.

[123] While the predictive advantages of ML-based approaches are clear, extracting physically interpretable insights from such models remains challenging. Recent work (including our own) has shown that relatively simple "ML-augmented" representations of PSR dynamics can be conceptually interpretable, can learn directly from data, and can achieve strong predictive performance, while relying on representational architectures that are substantially less complex than those of generic ML-based approaches. Building on this, this work develops MCP-based model architectures to investigate how hydrological process dominance and required model complexity vary geographically across the CONUS. To facilitate this, we adopt a catchment-by-catchment training strategy, enabling systematic exploration of regional variations in model structure and complexity in response to differences in climate, topography, and vegetation.

[124] Although this catchment-by-catchment approach does not directly exploit information shared across locations, it avoids reliance on generic ML representations that do not explicitly distinguish between geo-hydro-climatological settings at the architectural level. Instead, our focus is on developing ML-based strategies that account for regional variability in process dominance and model-structural complexity. To investigate these patterns, we focus on the CONUS and design model architectures that account for water storage in both snow and soil moisture. Analysis is conducted across five regions using masks to examine variations in relation to topography, snow processes, vegetation, and climate, representing a necessary step toward more geographically adaptive modeling frameworks.

[125] We constructed and tested several parsimonious one- and two-state MCP-based, ML-augmented, mass-conserving conceptual representations of daily PSR dynamics, including:

(i) A *"generic"* single-state model that does not distinguish between snowpack and soil-moisture dynamics
(ii) A *"snow-dominant"* single-state model focused on snow processes
(iii) A two-state *"snow–soil"* models that explicitly distinguish between snowpack and soil-moisture dynamics.

[126] The experiments clearly demonstrate that process dominance and necessary model complexity vary with geo-hydro-climatology across the CONUS. Notable findings include:

1) Strong and stable predictive performance can be achieved using parsimonious MCP-based representations that are physically interpretable, with geographical patterns of performance and complexity largely controlled by climate-related factors such as aridity and topography.

2) In snow-dominated regions, representation of snow dynamics is often more important than soil-moisture dynamics, and simple temperature-based formulations can perform remarkably well. However, augmenting model structure with additional hydrometeorological inputs or groundwater interactions yields localized improvements, suggesting the presence of missing processes in simplified representations.

3) When AIC is used for model selection, the *"generic"* $HMCP$ model (8–10 parameters) is selected at 60% of catchments, the *"snow-dominant"* $SNOWMCP$ model (11–20 parameters) at 18%, and the two-state $HYDROMCP$ model (18–27 parameters) at 22%, reflecting spatial variability in required model complexity.

4) Based on these findings, an OPTMCP framework with spatially adaptive complexity (1–2 cell states; 8–27 parameters) achieves a median $KGE_{ss} = 0.85$ across the CONUS, comparable to the more complex $NMCP_{wsh}(5, DS)$ benchmark (85 parameters), while outperforming it at nearly half of the catchments.

5) Although OPTMCP does not outperform the non–mass-conserving $LSTM(5)$ benchmark ($KGE_{ss} = 0.89$), it achieves competitive performance with substantially lower complexity and greater physical interpretability.

6) Geographical analysis shows that $OPTMCP$ is selected over ML benchmarks at 27% of catchments, indicating both the effectiveness of parsimonious, physically consistent models and the presence of missing processes in simplified representations.

### 4.2 Conclusions and Outlook

[127] Our previous MCP-based work (*Wang and Gupta, 2024ab; Wang & Gupta 2025*) focused primarily on the development and testing of the MCP unit and its associated model architectures, largely using data from a limited number of catchments. While these studies demonstrated the feasibility of the MCP framework, they did not explicitly address the role of geographically varying process dominance and associated model complexity, which is the central focus of this paper.

[128] Our view is that the future of hydrologic science lies in a successful synthesis of (i) theory-driven, physical-conceptual, hypothesis-based modeling focused on understanding, and (ii) data-driven ML-based modeling focused on maximizing predictive performance. While it remains to be seen how best to achieve this integration, we suggest that, at least for now, greater emphasis should be placed on the development of parsimonious and interpretable model architectures that acknowledge the reality of geographically (and temporally) varying process dominance and associated model complexity, rather than continuing to focus primarily on *"models of everywhere"* (MOEs). Such architectures can then be embedded within ML-based frameworks to enable next-generation MOEs that explicitly account for spatial variability in process dominance, while still benefiting from training across all available catchments.

[129] This pathway preserves decades of progress in hydrological science while leveraging the rapidly advancing capabilities of modern ML. We expect such an approach to yield improved predictive performance grounded in stronger theoretical understanding, thereby increasing confidence in out-of-sample generalization, counterfactual (e.g., scenario) analysis, and projections under changing climatic and anthropogenic conditions. We welcome and encourage constructive discussion and debate on these and related aspects of geoscientific model development in support of advancing scientific knowledge.

## Acknowledgements

The authors would like to thank WRR editorial team including Yi Zheng (Editor), an anonymous Associate Editor and three anonymous reviewers for taking the time to provide constructive comments. The first author (YHW) would like to thank Timothy O. Hodson and Patrick Broxton for valuable discussions, and the Department of Hydrology and Atmospheric Sciences and the Data Science Institute at the University of Arizona for providing HPC computational resources.

## Conflict of Interest

The authors declare no conflicts of interest relevant to this study

## Data Availability Statement

The data associated with this study have been deposited in the ESS-DIVE repository and are currently private while under repository review. They will be made publicly available upon publication of the peer-reviewed article.

## Appendix A: Mathematic Description of an MCP-Based unit

### A1 Architecture of the SOIL-MCP Unit

The architecture of a very simple MCP-based model can be illustrated by describing the single-node SOIL-MCP component (**Figure 3a**), which can be used to model mass-conservative PSR dynamics via the discrete time update equation:

$$X_{t+1} = X_t - O_t - L_t + U_t \tag{A1}$$

Here, at time step $t+1$, the nodal mass state $X_{t+1}$ of the system (which can be considered to be analogous to soil moisture storage) is computed by adding the mass of input (precipitation) flux $U_t$ that enters the node, and subtracting the masses of output fluxes $O_t$ (streamflow) and $L_t$ (evapotranspirative loss) that leave the node, during the time interval from $t$ to $t+1$.

Consistent with hydrological theory, we assume that $O_t$ and $L_t$ depend causally on the value of the state $X_t$ through the process parameterization equations $O_t = G_t^O \cdot X_t$ and $L_t = G_t^L \cdot X_t$, where $G_t^O$ and $G_t^L$ are context-dependent time-varying "*output*" and "*loss*" conductivity gating functions respectively, so that Eqn (A1) can be rewritten as:

$$X_{t+1} = X_t - G_t^O \cdot X_t - G_t^L \cdot X_t + U_t \tag{A2a}$$

$$X_{t+1} = G_t^R \cdot X_t + U_t \tag{A2b}$$

where the "*remember*" gate $G_t^R$ represents the fraction of the state $X_t$ that is retained by the system from one time step to the next. To ensure physical realism, we require that the time-evolving values of each of these gates ($G_t^O$, $G_t^L$ and $G_t^R$) remain both non-negative and less than or equal to $1.0$ at all times. Further, we require $G_t^R + G_t^O + G_t^L = 1$ to ensure conservation of mass, which means that the remember gate is computed from knowledge of the output and loss gates as $G_t^R = 1 - G_t^O - G_t^L$ ; this of course places a strict constraint on the relative values that $G_t^O$ and $G_t^L$ can take on. Note that Eqn (A1) simplifies to $X_{t+1} = X_t - O_t + U_t$ when system losses are not considered, which enables application to the modeling of processes such as surface channel routing and groundwater flow routing, as presented in *Wang & Gupta (2024b)*. In such cases, the mass conservation constraint reduces to $G_t^R + G_t^O = 1$.

We follow our previous work (*Wang & Gupta, 2024ab; Wang & Gupta, 2025*) and parameterize the output and loss gates as $G_t^O = f_{ML}^O(X_t)$ and $G_t^L = f_{ML}^L(PE_t)$ respectively, where the functional forms of $f_{ML}^O(\cdot)$ and $f_{ML}^L(\cdot)$ are assumed to be simple monotonic non-decreasing sigmoid functions; the application/learning of more complex functional forms is discussed by *Wang & Gupta (2024a)*. Specifically, the output and loss gates are initially represented as $G_t^O = \kappa_O \cdot \sigma(S_t^O)$ and $G_t^L = \kappa_L \cdot \sigma(S_t^L)$ respectively, where $S_t^O = a_O + b_O X_t$ and $S_t^L = a_L + b_L PE_t$, and where $\sigma$ can be any appropriate ML activation function (here chosen to be the sigmoid activation function $\sigma(S) = 1/(1+\exp(-S))$), and $\kappa_O$, $a_O$, $b_O$, $\kappa_L$, $a_L$, and $b_L$ are trainable parameters. Further, to ensure that $G_t^O$, $G_t^L$ and $G_t^R$ each remain on $[0,1]$ and also that $G_t^O + G_t^L + G_t^R = 1$, we actually set $\kappa_O = \exp(c_O)/\Psi$ and $\kappa_L = \exp(c_L)/\Psi$ where $\Psi = \exp(c_O) + \exp(c_L) + \exp(c_R)$ (which is equivalent to implementing the *SoftMax* function on the gates) and instead train on the set of seven parameters $\{a_O, b_O, c_O, a_L, b_L, c_L, c_R\}$ all of which can vary on $(-\infty, +\infty)$.

Note that one can implement a mass-relaxation gate ($f_t^{MR}$) to account for unobserved mass exchanges within the system (*Wang & Gupta, 2024a*). In this paper, we adopt the cell-state–dependent mass-relaxation gate proposed in Wang & Gupta (2024a) into the single-node architecture $HMCP(1,B)$ discussed in **Section 3.1**, denoted as $HMCP(1,MR)$ (see **Section 3.2.1** in the main text). The mass relaxation gate $f_t^{MR} = \kappa_{MR} \cdot Tanh(a_{MR} \cdot (\widetilde{X_t} - \tilde{c}_{MR}))$ where $\widetilde{X_t}$ is the scale cell state time series, $0 \leq \kappa_{MR} \leq 1$, $a_{MR}$ is set to be larger than 0. When the $X_t > c_{MR}$ at the unscaled domain, we have $0 < f_t^{MR} \leq \kappa_{MR}$ meaning the water is flowing out of the node. Conversely, when $X_t < c_{MR}$, we have $-\kappa_{MR} \leq f_t^{MR} < 0$ meaning water is flowing into the node.

Another variant of the SOIL-MCP unit involves augmenting the loss gate function to incorporate the cell state, rather than using only PET as the contextual variable (*Wang & Gupta, 2025*). As a result, $S_t^L = a_L + b_L\, PE_t + b_L^* X_t$, while the other formulations remain the same as summarized in 5), except that one additional parameter, $b_L^*$, is introduced into the loss gate. We denoted this model as $HMCP(1, AL)$ in the current paper with the discussion of performance in **Section 3.2.2**.

### A2 Architecture of the SNOW-MCP Unit

Similarly, the simple single-node SNOW-MCP component (**Figure 3b**) represents the mass-conserving process of snowpack accumulation and melt via the discrete time update equation:

$$SWE_{t+1} = SWE_t - O_t^s - L_t^s + U_t^s \tag{A3}$$

Here, at time step $t+1$, the mass state $SWE_{t+1}$ of the system (which can be considered to be analogous to storage of snow water equivalent) is computed by adding the mass of input flux $U_t^s$ (water in the form of snow) that enters the node, and subtracting the masses of output fluxes $O_t^s$ (snowmelt) and $L_t^s$ (evaporation or sublimation loss) that leave the node, during the time interval from $t$ to $t+1$. Here, $U_t^s$ is the snowfall input which is assumed to depend on the value of the total water mass input $U_t$ through the process parameterization equation $U_t^s = G_t^{RS} \cdot U_t$, where $G_t^{RS}$ is a context-dependent time-varying "*rain-snow partition"* conductivity gating function. Since $G_t^{RS}$ must vary between 0 and 1, the rainfall partitioning amount $U_t^R$ can be represented as $U_t - U_t^s = (1 - G_t^{RS}) \cdot U_t$

Further, $O_t^s$ and $L_t^s$ are assumed to depend on the value of the state $SWE_t$ through the process parameterization equations $O_t^s = G_t^{SO} \cdot SWE_t = G_t^M \cdot SWE_t$ and $L_t^s = G_t^{SL} \cdot SWE_t$, where $G_t^M$ and $G_t^{SL}$ are context-dependent time-varying *"snow output"(melt)* and *"snow loss"* conductivity gating functions respectively, so that Eqn (A3) can be rewritten as:

$$SWE_{t+1} = SWE_t - G_t^M \cdot SWE_t - G_t^{SL} \cdot SWE_t + U_t^s \tag{A4a}$$

$$SWE_{t+1} = G_t^{SR} \cdot SWE_t + U_t^s \tag{A4b}$$

where the "*remember*" gate $G_t^{SR}$ for snowpack represents the fraction of the state $SWE_t$ that is retained by the system from one time step to the next. Similar to the SOIL-MCP architecture, the values of the $G_t^M$, $G_t^{SL}$ and $G_t^{SR}$ gates all vary with time, while remaining both non-negative and less than or equal to $1.0$ at all times. For mass conservation to hold, we require that $G_t^{SR} + G_t^M + G_t^{SL} = 1$. Accordingly, the remember gate $G_t^{SR}$ is computed as $G_t^{SR} = 1 - G_t^M - G_t^{SL}$ from knowledge of the melt (output) and loss gates.

We parameterize the corresponding rain-snow partition, melt and snow-loss gates as $G_t^{RS} = f_{ML}^{RS}(T_t)$, $G_t^M = f_{ML}^M(T_t, SWE_t)$ and $G_t^{SL} = f_{ML}^{SL}(T_t, SWE_t)$ respectively, where the functional forms of $f_{ML}^{RS}(\cdot)$, $f_{ML}^M(\cdot)$ and $f_{ML}^{SL}(\cdot)$ are (similarly to the SOIL-MCP) assumed to be simple monotonic non-decreasing sigmoid functions. Similarly, the rain-snow partition, melt and snow-loss gates are initially represented as $G_t^{RS} = \sigma(S_t^{RS})$, $G_t^M = \kappa_M \cdot \sigma(S_t^M)$ and $G_t^{SL} = \kappa_{SL} \cdot \sigma(S_t^{SL})$ respectively, where $S_t^{RS} = a_{RS} + b_{RS}\, T_t$, $S_t^M = a_M + b_M\, T_t + b_M^*\, SWE_t$ and $S_t^{SL} = a_{SL} + b_{SL}\, T_t + b_{SL}^*\, SWE_t$, where $\sigma$ is chosen to be the sigmoid activation function, and $\kappa_M$, $a_M$, $b_M$, $b_{SL}^*$, $\kappa_{SL}$, $a_{SL}$, $b_{SL}$ and $b_{SL}^*$ are trainable parameters. To ensure conservation of mass (when $G_t^{SR} + G_t^M + G_t^{SL} = 1$), we set $\kappa_M = \exp(c_M)/\Psi$ and $\kappa_{SL} = \exp(c_{SL})/\Psi$ where $\Psi = \exp(c_M) + \exp(c_{SL}) + \exp(c_{SR})$ and instead train on the set of nine parameters $\{a_{RS}, b_{RS}, a_M, b_M, b_M^*, c_M, a_{SL}, b_{SL}, b_{SL}^*, c_{SL}, c_{SR}\}$ all of which can vary on $(-\infty, +\infty)$.

Of course, the behaviors of the gates $G_t^{RS}$, $G_t^M$ and $G_t^{SL}$ may actually depend on a broad set of contextual variables beyond mean temperature. For instance, the rain–snow partitioning process can be influenced by factors related to atmospheric wetness conditions (*Harpold et al., 2017; Jennings et al., 2018; Jennings et al.,*

*2025*). These factors can also be extended to the processes of melt and sublimation loss, where topography, vegetation, solar radiation, and snowpack properties play important roles (see summary in *Wang et al., 2019*). Preliminary tests showed no performance gain from adding vapor pressure deficit, dew point temperature, or shortwave radiation as gating-context variables. We therefore use simple gate representations based mainly on mean temperature (*Patil & Stieglitz, 2014*), with melt and loss gates additionally conditioned on internal SWE. More complex mass–energy coupling architectures are left for future work (*Anderson, 1976*).

## Appendix B: Details of the Data-Splitting Algorithm

Over each of the 513 catchments, we set the $\mathcal{D}_{train}$: $\mathcal{D}_{select}$ : $\mathcal{D}_{test}$ partitioning ratio to be 2:1:1 respectively for both streamflow and the SWE. Taking streamflow as an example, the data splitting procedure first sorts the streamflow data based on magnitude. Next, the timestep associated with the largest streamflow magnitude is paired with the timestep associated with the smallest streamflow value, continuing with the next largest and smallest values and so on, until all time steps have been paired. These pairs are then sequentially allocated, in the abovementioned ratio, to the three independent sets (following the sequence of $\mathcal{D}_{train} \rightarrow \mathcal{D}_{test} \rightarrow \mathcal{D}_{select} \rightarrow \mathcal{D}_{train}$ etc.) until all pairs have been assigned. Since we adopt data from WY 1982 to WY 2008, resulting in 9,862 data points, the training subset consists of 4,930 time steps, while the selection and testing subsets each contain 2,466 time steps. Also, WY 1982 is repeated three times at the beginning for the purpose of internal state initialization, resulting in a total of 10,957 time steps.

## Appendix C: Details of Metrics used for Training and Evaluation

The metric used for model training was the *Kling-Gupta Efficiency* ($KGE$; Eqn. C1) (*Gupta et al., 2009)*. Notice that there is essentially no difference between training against $KGE$ (Eqn. C1) and $KGE_{ss}$ (Eqn. C2), given that the transformation from one to the other involves simple scaling (*Knoben et al., 2019b; Khatami et al., 2020*). Performance assessment was conducted using $KGE_{ss}$ and the components of $KGE$ (Eqns C3-C5):

$$KGE = 1 - \sqrt{((\rho^{KGE} - 1)^2 + (\beta^{KGE} - 1)^2 + (\alpha^{KGE} - 1)^2)} \quad \text{(C1)}$$

$$KGE_{ss} = 1 - \frac{(1 - KGE)}{\sqrt{2}} \quad \text{(C2)}$$

$$\alpha^{KGE} = \frac{\sigma_s}{\sigma_o} \quad \text{(C3)}$$

$$\beta^{KGE} = \frac{\mu_s}{\mu_o} \quad \text{(C4)}$$

$$\rho^{KGE} = \frac{Cov_{so}}{\sigma_s \sigma_o} \quad \text{(C5)}$$

where $\sigma_s$ and $\sigma_o$ are the standard deviations, and $\mu_s$ and $\mu_o$ are the means, of the corresponding data-period simulated and observed streamflow hydrographs respectively and, similarly, $Cov_{so}$ is the covariance between the simulated and observed values. Note that $KGE$ (and therefore $KGE_{ss}$) is maximized when $\alpha^{KGE}$, $\beta^{KGE}$ and $\rho^{KGE}$ are all $1.0$.

Following *Wang & Gupta (2025),* we define $\alpha_*^{KGE}$ and $\beta_*^{KGE}$ as shown in Eqns (C6-C7), to circumvent the issue that $\alpha^{KGE}$ and $\beta^{KGE}$ are optimal at $1.0$, which can cause the values to be either larger or smaller. This modification ensures that the $\alpha_*^{KGE}$ and $\beta_*^{KGE}$ range from negative infinity ($-\infty$) to the optimal value at $1.0$.

$$\alpha_*^{KGE} = 1 - |1 - \alpha^{KGE}| \quad \text{(C6)}$$

$$\beta_*^{KGE} = 1 - |1 - \beta^{KGE}| \quad \text{(C7)}$$

## Appendix D: Model Selection Metric

We use the Akaike Information Criterion (AIC; *Akaike, 1974*) to guide model selection. AIC is a widely used criterion grounded in information theory, derived from minimizing the Kullback–Leibler (KL) divergence between the true data-generating process and a candidate model (*Cover & Thomas, 2006*). It balances model fit and complexity by penalizing the number of parameters, thereby favoring parsimonious models that generalize better to unseen data. Formally, the AIC is expressed as Eqn. (D1).

$$AIC = -2\ln(l) + 2k \tag{D1}$$

where the first term represents the negative log-likelihood, and the second term accounts for model complexity and helps to reduce overfitting. Here, $l$ denotes the likelihood, which is typically estimated from an *"out-of-sample"* dataset (testing set) to preserve unbiasedness and mitigate overfitting (*Burnham & Anderson, 2002*). The parameter $k$ in the second term refers to the number of independently estimated parameters.

In this study, the likelihood term was computed following Strupczewski et al. (2001). For each candidate MCP-based architecture and basin, residuals at the testing time points were fitted to six candidate probability density functions: Normal, Fisher–Tippett Type I / Gumbel, gamma / Pearson Type III special case, lognormal, lower-bounded lognormal, and Pearson Type III. These candidate PDFs span symmetric and skewed residual shapes with both two- and three-parameter forms, providing a flexible residual-error representation across heterogeneous hydroclimatic conditions, including both dry and wet basins. No single residual distribution was assumed a priori; instead, the residual PDF was selected for each model–basin combination based on the maximum likelihood among the six candidates. To ensure comparability across model architectures and basins, $k$ included both the independently estimated MCP model parameters and the fitted PDF parameters. The optimal architecture at each basin was then selected as the one with the minimum AIC. For a comprehensive theoretical overview of the $AIC$, we refer the reader to the established summaries provided by *Hodson et al. (2024)*.

## Appendix E: Note on Training Procedure

### E1 Details of the Training Procedure and Hyperparameter Selection

We implemented the same model training protocol sued in our previous work *(Wang & Gupta, 2024ab; Wang & Gupta, 2025)*, which included a "three-year" spin-up phase wherein the first water year's data (WY 1982) was repeated three times at the beginning of the 27-year simulation to reduce potential state initialization bias. We employed the ADAM optimizer (*Kingma & Ba, 2014*) to iteratively update model parameters. The loss function and its gradients were computed using streamflow and SWE values/timesteps designated for the training subset.

Each model architecture with newly added components was trained 10 times using random parameter initialization. The run with the highest $KGE$ score (and thus $KGE_{ss}$), evaluated on the selection set, was retained. If the model used only components inherited from existing models (i.e., not initialized from scratch), the optimal version was selected solely based on its best performance on the *selection* set. We refer the reader to the Supplementary Materials (***Table S1***) for additional details on the parameter initialization, learning rates and number of epochs used for each network configuration.

In several cases, performance metrics could not be derived across any of the random seed initializations. This issue typically arises when snowpack- or temperature-related inputs are incorporated into the model architecture for catchments that lack sufficient snow presence. For such cases, we initially train the model using the default settings, and subsequently re-train the networks at these locations with gradient clipping applied (maximum gradient magnitude set to 1.0). Catchments that still fail to produce valid results under the snowpack-informed architecture are excluded from further analysis (see ***Section 3.4***).

## E2 Notes on the Catchments Selected for Analysis

Snowpack accumulation and melt are not the dominant hydrologic processes across all 513 study catchments. Nevertheless, from a modeling perspective, it remains valuable to assess model performance in regions characterized by ephemeral rather than persistent snowpack, despite the potentially low predictive skill in such cases. Here, the SNOW-MCP and Hydro-MCP configurations were trained using the procedure described in **Appendix E1**. Catchments that did not yield valid performance metrics, due to issues such as exploding gradients or vanishing (zero) gradients, were excluded from the analysis (11 for SNOW-MCP and 37 for HYDRO-MCP).

## E3 Notes on the Role of the SNOW-MCP Architecture

For the SNOW-MCP unit, It is important to distinguish between the procedures of model development and application. During the model development stage, the internal cell state of the SNOW-MCP was fitted to the UA SWE data to emulate snowpack dynamics (**Section 3.3.1** and **Section 3.3.2**). During the model application stage (**Section 3.3.3** and **Section 3.3.4**), when only the SNOW component was used to model rainfall–runoff dynamics, the combination of precipitation bypass and snowmelt generated through SNOW-MCP was trained against streamflow.

## E4 Notes on the Relaxation of Constraint on Potential Evapotranspiration

As proposed in *Wang & Gupta (2024a)*, we constrained the computed (*"actual"*) evapotranspirative loss ($L_t = G_t^L \cdot X_t$) to be less than or equal to the *"potential"* evapotranspirative loss $D_t$ by replacing the *"unconstrained"* loss gate $G_t^L$ described in **Section 2.3.1**, with the "*physically-constrained*" loss gate defined as $G_t^{L^{con}} = G_t^L - ReLU(G_t^L - \frac{D_t}{X_t})$ and denoted as $MC\{O_\sigma L_\sigma^{con}\}$ in *Wang & Gupta (2025)*, and denoted here as $HMCP(1, B^{(con)})$.

We ran this physically-constrained version over all 513 selected catchment and show the spatial distribution of $KGE_{ss}$ skill in **Figure E1a**. Note that the derived PET time series and the associated constraint fail to represent the actual physical process at certain locations; specifically, 14 and 45 of the catchments exhibited $KGE_{ss}$ values lower than $-1$ and $0$, respectively. While low skill was observed particularly in the mountainous western U.S. and the Midwest, there is no clear indication of the underlying causes for the severe performance degradation (i.e., $KGE_{ss} < -1$, as shown in the box plot of **Figure E1b**), especially when compared to the improved performance under relaxed PET constraints as shown in $MC\{O_\sigma L_\sigma\}$, , denoted here as $HMCP(1, B)$. A similar observation is found when constraining the loss-gate–augmented $MC\{O_\sigma L_{\sigma+}\}$ model, denoted here as $HMCP(1, AL)$, with the PET constraint, resulting in $MC\{O_\sigma L_{\sigma+}^{con}\}$ (denoted here as $HMCP(1, AL^{(con)})$), as summarized in **Figure E2a-b**.

A future more detailed examination of these poorly performing sites is warranted to diagnose the sources of error and to evaluate potential shortcomings in the PET assumptions or model architecture. Such an investigation aligns with the objectives of depth-focused studies targeting a single or a few specific locations and is therefore beyond the scope of the present manuscript. Accordingly, the "non-*physically-constrained*" version of the loss function was employed as the baseline representation of the rainfall–runoff component throughout this study.

## Table Captions

**Table 1.** Summary of geographic and environmental characteristics (wetness, elevation, snow cover, forest cover, and climatologic classification) for the regions analyzed in this study.

**Table 2.** Summary of $KGE_{ss}$ metric for the MCP-based architectures across geographic regions in the CONUS

**Table 3.** Summary of streamflow and SWE performance metrics for the two selected basins.

**Table 4.** Summary of basin counts and $KGE_{ss}$ metrics for the model architectures of HMCP, SNOWMCP, and HYDROMCP across geographic regions, comparing model selection outcomes based on the AIC criterion versus the maximum KGE metric.

**Table 5.** Summary of basin counts and $KGE_{ss}$ metrics for the model benchmarks of $OPTMCP$, $NMCP_{wsh}(5, DS)$ , and $LSTM(5)$ across geographic regions, comparing model selection outcomes based on the AIC criterion versus the maximum KGE metric.

## Figure Captions

**Figure 1:** Geographical information for the 513 CAMELS-US catchments used in this study, including (a) selected geographic regions, (b) elevation (m), (c) snowy versus non-snowy basin delineation, and (d) forest versus open basin classification.

**Figure 2:** Spatial information for the 513 selected CAMELS-US catchments, including the Köppen–Geiger climate classification based on: (a) the five main classes, (b) the 30 detailed classes, and (c) the aridity index.

**Figure 3:** Direct-graph representations of the HMCP units and architectures used in this study, including (a) the Mass-Conserving Perceptron (MCP) unit proposed in Wang & Gupta (2024ab) and Wang & Gupta (2025), (b) the SNOWMCP unit proposed in this study, and the two HYDROMCP architectures that couple SOILMCP and SNOWMCP using (c) routing (serial connection) and (d) bypass (parallel connection)

**Figure 4:** Skill metrics for HMCP(1,B) across 513 CONUS catchments, including (a) the KGE skill score ($KGE_{ss}$) and the three components of KGE: (b) linear correlation ($\gamma^{KGE}$), (c) flow variability ratio ($\alpha^{KGE}$) and (d) mass balance ratio ($\beta^{KGE}$).

**Figure 5:** Skill metrics for HMCP(1,MR) across 513 CONUS catchments, including (a) the KGE skill score ($KGE_{ss}$) and the three components of KGE : (b) linear correlation ($\gamma^{KGE}$), (c) flow variability ratio ( $\alpha^{KGE}$ ), and (d) mass balance ratio ( $\beta^{KGE}$ ). Subplots (e–h) show the corresponding differences in these metrics between HMCP(1,MR) and HMCP(1,B).

**Figure 6:** Spatial distribution of (a) the threshold $\tilde{C}_{MR}$ value learned in the HMCP(1,MR) architecture, and the associated gain ratio for the (b) testing period and (c) full time period.

**Figure 7:** KGE skill score ($KGE_{ss}$) across 513 CONUS catchments for three versions of the single-state SNOWMCP(1) architecture: (a) SNOWMCP(1,B), (b) SNOWMCP(1,G), and (c) SNOWMCP(1,C). Circles indicate snowy basins, and diamonds indicate non-snowy basins.

**Figure 8:** Skill metrics for SNOWMCP (1,BQ) across 513 CONUS catchments, including (a) the KGE skill score ($KGE_{ss}$) and the three components of KGE : (b) linear correlation ($\gamma^{KGE}$), (c) flow

variability ratio ($\alpha^{KGE}$) and (d) mass balance ratio ($\beta^{KGE}$). Circles indicate snowy basins, and diamonds indicate non-snowy basins.

**Figure 9:** Performance of SNOWMCP in simulating streamflow, shown as: (a) Boxplots of the KGE skill score ($KGE_{ss}$) distribution for the three SNOWMCP architectures — SNOWMCP(1,BQ), SNOWMCP(1,GQ), and SNOWMCP(1,CQ) — across 513 catchments, and the spatial distribution of skill metrics for SNOWMCP(1,GQ), including (b) KGE skill score ($KGE_{ss}$) and the three components of KGE : (c) linear correlation ($\gamma^{KGE}$), (d) flow variability ratio ($\alpha^{KGE}$), and (e) mass balance ratio ($\beta^{KGE}$). Circles indicate snowy basins, and diamonds indicate non-snowy basins in subplots (b)–(e).

**Figure 10:** Skill metrics for HYDROMCP (2,B) across 513 CONUS catchments, including (a) the KGE skill score ($KGE_{ss}$) and the three componentsof KGE : (b) linear correlation ($\gamma^{KGE}$), (c) flow variability ratio ($\alpha^{KGE}$) and (d) mass balance ratio ($\beta^{KGE}$). Circles indicate snowy basins, and diamonds indicate non-snowy basins.

**Figure 11:** Time series under different flow regimes for selected CAMELS-US basins. Panels (a–c) show simulated and observed streamflow for basin 02296500 during dry, median, and wet water years (WY 1989, 1993, and 1998, respectively). Panels (d–i) show simulated and observed streamflow for basin 13023000 during dry, median, and wet water years (WY 1992, 2000, and 1997, respectively). Panels (j–l) show corresponding snow water equivalent (SWE) simulations and observations for basin 13023000.

**Figure 12:** Optimal MCP-based architecture selection based on (a) the Akaike Information Criterion (AIC) and (b) the associated testing period KGE skill score ($KGE_{ss}$) across 513 basins. Circles indicate snowy basins, and diamonds indicate non-snowy basins.

**Figure 13:** Empirical cumulative density plots of performance metrics across 513 basins, including (a) KGE skill score (KGEss), (b) linear correlation ($r^{KGE}$), (c) flow variability ratio ($\alpha^{KGE}$), and (d) mass balance ratio ($\beta^{KGE}$). Note that HYDROMCP(2,P) and HYDROMCP(2,S) are derived from the "Basic" case, HYDROMCP(2,B).

**Figure 14:** Performance differences between the OPTMCP architectures and the single-layer 5-node distributed-state mass-conserving network, shown for (a) the KGE skill score ($KGE_{ss}$), and the three components of KGE : (b) linear correlation ($r^{KGE}$), (c) adjusted flow variability ratio ($\alpha_*^{KGE}$), and (d) mass balance ratio ($\beta_*^{KGE}$). Circles indicate snowy basins, and diamonds indicate non-snowy basins.

**Figure 15:** Optimal architecture selection among $OPTMCP$, $NMCP_{wsh}(5, DS)$, and $LSTM(5)$ based on (a) the Akaike Information Criterion (AIC) and (b) the associated testing period KGE skill score ($KGE_{ss}$) across 513 basins. Panels (c) and (d) show corresponding results when model selection is performed using KGE as the criterion instead of AIC. Circles indicate snowy basins, and diamonds indicate non-snowy basins.

## Figure Captions for Appendix

**Figure E1:** Performance of the PET-constrained $HCMP(1, B)$ model architecture, including: (a) spatial distribution of the KGE skill score ($KGE_{ss}$) for the PET-constrained $HCMP(1, B)$ model denoted as $HCMP(1, B^{(con)})$, and (b) box plot comparing PET-constrained $HCMP(1, B)$ model ($HCMP(1, B^{(con)})$) with the baseline $HCMP(1, B)$ model without PET constraint.

**Figure E2:** Performance of the PET-constrained architecture, including: (a) spatial distribution of the KGE skill score ($KGE_{ss}$) for the PET-constrained version of the $HCMP(1, AL)$ model denoted as $HCMP( 1, AL^{(con)})$ , and (b) box plot comparing PET-constrained $HCMP(1, AL)$ model ($HCMP(1, AL^{(con)}$) with the loss gate augmented $HCMP(1)$ model without PET constraint $HCMP(1, AL)$.

**Table 1.** Summary of geographic and environmental characteristics (wetness, elevation, snow cover, forest cover, and climatologic classification) for the regions analyzed in this study.

| | Appalachian Mountain (AM) | Rocky Mountain (RM) | Colorado River Basin (CRB) | Sierra Neveda (SN) | Cascades (CC) | Eastern CONUS | Western CONUS | CONUS |
|---|---|---|---|---|---|---|---|---|
| No. of Basins | 103 | 54 | 34 | 10 | 33 | 326 | 187 | 513 |
| Avg. Elevation [meter] | 538.54 | 2530.35 | 2338.40 | 1988.79 | 933.30 | 352.67 | 1484.45 | 765.23 |
| Median Aridity Index | 0.69 | 1.45 | 1.91 | 1.08 | 0.33 | 0.84 | 1.28 | 0.86 |
| No of. Snowy Basins | 91 | 54 | 28 | 9 | 32 | 212 | 148 | 360 |
| No of. Non-Snowy Basins | 12 | 0 | 6 | 1 | 1 | 114 | 39 | 153 |
| No of. Forest Basins | 84 | 25 | 8 | 7 | 32 | 172 | 110 | 282 |
| No of. Non-Forest (Open) Basins | 19 | 29 | 26 | 3 | 1 | 154 | 77 | 231 |
| No of. Arid Basins | 0 | 5 | 14 | 0 | 0 | 4 | 31 | 35 |
| No of. Temperate Basins | 43 | 0 | 3 | 3 | 18 | 175 | 67 | 242 |
| No of. Cold Basins | 60 | 47 | 16 | 7 | 15 | 147 | 87 | 234 |
| No of. Polar Basins | 0 | 2 | 1 | 0 | 0 | 0 | 2 | 2 |
| Dominant Climate Zone (All) | Dfb | Dfc | BSk | Dsb | Csb | Cfa | Csb | Cfa |

**Note.** Cold, no dry season, warm summer (Dfb); Cold, no dry season, cold summer (Dfc); Arid, steppe, cold (BSk); Cold, dry summer, warm summer (Dsb); Temperate, dry summer, warm summer (Csb); Temperate, no dry season, hot summer (Cfa); Temperate, dry summer, warm summer (Csb); no dry season, hot summer (Cfa);

**Table 2.** Summary of $KGE_{ss}$ metric for the MCP-based architectures across geographic regions in the CONUS

| Region | Percentile | Architecture Backbone | | | | | | | | | | |
|---|---|---|---|---|---|---|---|---|---|---|---|---|
| | | SOILMCP (HMCP) | | | SNOWMCP | | | HYDROMCP | | | | |
| | | $(1)$ | $(1, AL)$ | $(1, MR)$ | $(1, BQ)$ | $(1, GQ)$ | $(1, CQ)$ | $(1, B)$ | $(2, S)$ | $(2, P)$ | $(2, G)$ | $(2, C)$ |
| CONUS | 5 | 0.40 | 0.43 | 0.49 | 0.31 | 0.41 | 0.43 | -2.97 | -2.82 | -1.59 | -1.60 | -1.28 |
| | 25 | 0.60 | 0.62 | 0.67 | 0.67 | 0.70 | 0.70 | 0.25 | 0.09 | 0.61 | 0.62 | 0.64 |
| | 50 | 0.72 | 0.74 | 0.76 | 0.78 | 0.80 | 0.80 | 0.63 | 0.58 | 0.79 | 0.80 | 0.81 |
| | 75 | 0.81 | 0.82 | 0.82 | 0.85 | 0.86 | 0.86 | 0.77 | 0.76 | 0.87 | 0.87 | 0.87 |
| | 95 | 0.89 | 0.89 | 0.89 | 0.91 | 0.92 | 0.92 | 0.86 | 0.86 | 0.94 | 0.94 | 0.94 |
| Eastern CONUS | 50 | 0.75 | 0.77 | 0.76 | 0.76 | 0.77 | 0.77 | 0.67 | 0.67 | 0.78 | 0.78 | 0.79 |
| Western CONUS | | 0.64 | 0.67 | 0.73 | 0.85 | 0.86 | 0.86 | 0.53 | 0.46 | 0.85 | 0.86 | 0.86 |
| AM | | 0.73 | 0.75 | 0.74 | 0.78 | 0.78 | 0.80 | 0.67 | 0.67 | 0.78 | 0.78 | 0.77 |
| RM | | 0.53 | 0.55 | 0.72 | 0.89 | 0.89 | 0.90 | 0.40 | 0.38 | 0.92 | 0.93 | 0.93 |
| CRB | | 0.51 | 0.55 | 0.67 | 0.83 | 0.84 | 0.84 | 0.24 | 0.20 | 0.86 | 0.89 | 0.90 |
| SN | | 0.56 | 0.59 | 0.56 | 0.85 | 0.86 | 0.89 | 0.59 | 0.57 | 0.90 | 0.91 | 0.90 |
| CC | | 0.80 | 0.81 | 0.83 | 0.88 | 0.88 | 0.86 | 0.72 | 0.70 | 0.87 | 0.87 | 0.87 |
| Snowy | | 0.71 | 0.72 | 0.73 | 0.80 | 0.81 | 0.81 | 0.60 | 0.54 | 0.80 | 0.81 | 0.82 |
| Non-Snowy | | 0.79 | 0.81 | 0.81 | 0.68 | 0.74 | 0.76 | 0.69 | 0.69 | 0.74 | 0.76 | 0.76 |
| Forest | | 0.76 | 0.77 | 0.77 | 0.79 | 0.82 | 0.82 | 0.67 | 0.66 | 0.82 | 0.82 | 0.83 |
| Open | | 0.69 | 0.71 | 0.73 | 0.76 | 0.76 | 0.75 | 0.54 | 0.49 | 0.77 | 0.78 | 0.78 |
| Arid Climate | | 0.57 | 0.60 | 0.62 | 0.51 | 0.69 | 0.71 | 0.38 | 0.37 | 0.79 | 0.77 | 0.78 |
| Temperate Climate | | 0.79 | 0.81 | 0.80 | 0.77 | 0.80 | 0.79 | 0.71 | 0.71 | 0.76 | 0.78 | 0.79 |
| Cold Climate | | 0.68 | 0.68 | 0.72 | 0.80 | 0.81 | 0.81 | 0.57 | 0.49 | 0.82 | 0.82 | 0.82 |
| Polar Climate | Avg | 0.42 | 0.43 | 0.82 | 0.93 | 0.94 | 0.94 | 0.45 | 0.17 | 0.65 | 0.95 | 0.96 |

**Table 3.** Summary of streamflow and SWE performance metrics for the two selected basins.

| Metric | Flow Regime | Streamflow: Basin 02296500 | | | Flow Regime | Streamflow: Basin 13023000 | | | | | Snow Water Equivalent: Basin 13023000 | | |
|---|---|---|---|---|---|---|---|---|---|---|---|---|---|
| | | $HMCP(1,B)$ | $NMCP_{wsh}$ $(5,DS)$ | $LSTM(5)$ | | $HMCP(1,B)$ | $SNOWMCP$ $(1,BQ)$ | $HYDROMCP$ $(2,B)$ | $NMCP_{wsh}$ $(5,DS)$ | $LSTM(5)$ | $SNOWMCP$ $(1,B)$ | $SNOWMCP$ $(1,BQ)$ | $HYDROMCP$ $(2,B)$ |
| $KGE_{ss}$ | Dry (WY1989) | 0.72 | 0.64 | 0.65 | Dry (WY1992) | 0.50 | 0.54 | 0.49 | 0.78 | 0.66 | 0.93 | 0.08 | 0.87 |
| | Median (WY1993) | 0.76 | 0.77 | 0.75 | Median (WY2000) | 0.66 | 0.88 | 0.83 | 0.89 | 0.94 | 0.86 | 0.01 | 0.84 |
| | Wet (WY1998) | 0.95 | 0.93 | 0.95 | Wet (WY1997) | 0.53 | 0.87 | 0.90 | 0.85 | 0.97 | 0.88 | 0.48 | 0.90 |
| $RMSE$ | Dry (WY1989) | 0.25 | 0.22 | 0.24 | Dry (WY1992) | 0.61 | 0.48 | 0.42 | 0.28 | 0.31 | 15.17 | 140.59 | 19.68 |
| | Median (WY1993) | 0.49 | 0.36 | 0.41 | Median (WY2000) | 0.85 | 0.46 | 0.48 | 0.36 | 0.26 | 35.16 | 211.12 | 36.65 |
| | Wet (WY1998) | 0.96 | 0.79 | 0.55 | Wet (WY1997) | 1.97 | 0.88 | 0.68 | 0.73 | 0.50 | 53.08 | 275.63 | 54.99 |
| $MAE$ | Dry (WY1989) | 0.13 | 0.13 | 0.13 | Dry (WY1992) | 0.53 | 0.32 | 0.27 | 0.23 | 0.21 | 11.07 | 123.56 | 15.26 |
| | Median (WY1993) | 0.30 | 0.24 | 0.23 | Median (WY2000) | 0.63 | 0.36 | 0.33 | 0.27 | 0.17 | 20.11 | 167.60 | 21.67 |
| | Wet (WY1998) | 0.56 | 0.44 | 0.33 | Wet (WY1997) | 1.24 | 0.63 | 0.45 | 0.42 | 0.32 | 41.65 | 207.49 | 40.86 |

**Table 4.** Summary of basin counts and $KGE_{ss}$ metrics for the model architectures of HMCP, SNOWMCP, and HYDROMCP across geographic regions, comparing model selection outcomes based on the AIC criterion versus the maximum KGE metric.

| Regions | Model Selection Based on Min AIC | | | | Model Selection Based on Max KGE | | | |
|---|---|---|---|---|---|---|---|---|
| | No. of Basins | | | $KGE_{ss}^{50\%}$ | No. of Basins | | | $KGE_{ss}^{50\%}$ |
| | HMCP | SNOWMCP | HYDROMCP | OPTMCP | HMCP | SNOWMCP | HYDROMCP | OPTMCP |
| CONUS | 306 | 94 | 113 | 0.78 | 84 | 182 | 247 | 0.85 |
| Eastern CONUS | 216 | 55 | 55 | 0.77 | 69 | 119 | 138 | 0.83 |
| Western CONUS | 90 | 39 | 58 | 0.81 | 15 | 63 | 109 | 0.89 |
| AM | 68 | 21 | 14 | 0.75 | 7 | 58 | 38 | 0.82 |
| RM | 12 | 11 | 31 | 0.89 | 0 | 2 | 52 | 0.94 |
| CRB | 14 | 9 | 11 | 0.71 | 1 | 8 | 25 | 0.90 |
| SN | 3 | 1 | 6 | 0.83 | 0 | 3 | 7 | 0.92 |
| CC | 17 | 11 | 5 | 0.86 | 3 | 16 | 14 | 0.89 |
| Snowy | 192 | 80 | 88 | 0.77 | 34 | 140 | 186 | 0.85 |
| Non-Snowy | 114 | 14 | 25 | 0.79 | 50 | 42 | 61 | 0.85 |
| Forest | 170 | 50 | 62 | 0.78 | 36 | 109 | 137 | 0.86 |
| Open | 136 | 44 | 51 | 0.77 | 48 | 73 | 110 | 0.82 |
| Arid Climate | 17 | 3 | 15 | 0.67 | 3 | 7 | 25 | 0.81 |
| Temperate Climate | 163 | 39 | 40 | 0.80 | 62 | 89 | 91 | 0.85 |
| Cold Climate | 125 | 52 | 57 | 0.75 | 19 | 86 | 129 | 0.85 |
| Polar Climate | 1 | 0 | 1 | 0.63 | 0 | 0 | 2 | 0.96 |

**Table 5.** Summary of basin counts and $KGE_{ss}$ metrics for the model benchmarks of $OPTMCP$, $NMCP_{wsh}(5, DS)$, and $LSTM(5)$across geographic regions, comparing model selection outcomes based on the AIC criterion versus the maximum KGE metric.

| Regions | Model Selection Based on Min AIC | | | | Model Selection Based on Max KGE | | | |
|---|---|---|---|---|---|---|---|---|
| | No. of Basins | | | $KGE_{ss}^{50\%}$ | No. of Basins | | | $KGE_{ss}^{50\%}$ |
| | $OPTMCP$ | $NMCP_{wsh}(5, DS)$ | $LSTM(5)$ | Optimal | $OPTMCP$ | $NMCP_{wsh}(5, DS)$ | $LSTM(5)$ | Best |
| CONUS | 140 | 205 | 168 | 0.82 | 67 | 71 | 375 | 0.85 |
| Eastern CONUS | 88 | 159 | 79 | 0.80 | 38 | 58 | 230 | 0.84 |
| Western CONUS | 52 | 46 | 89 | 0.88 | 29 | 13 | 145 | 0.90 |
| AM | 22 | 49 | 32 | 0.78 | 9 | 11 | 83 | 0.83 |
| RM | 8 | 4 | 42 | 0.95 | 7 | 1 | 46 | 0.95 |
| CRB | 5 | 3 | 26 | 0.93 | 2 | 0 | 32 | 0.93 |
| SN | 9 | 0 | 1 | 0.89 | 5 | 0 | 5 | 0.90 |
| CC | 11 | 9 | 13 | 0.84 | 5 | 1 | 27 | 0.87 |
| Snowy | 89 | 138 | 133 | 0.82 | 40 | 38 | 282 | 0.85 |
| Non-Snowy | 51 | 67 | 35 | 0.83 | 27 | 33 | 93 | 0.85 |
| Forest | 77 | 114 | 91 | 0.84 | 40 | 37 | 205 | 0.87 |
| Open | 63 | 91 | 77 | 0.80 | 27 | 34 | 170 | 0.84 |
| Arid Climate | 13 | 7 | 15 | 0.79 | 7 | 3 | 25 | 0.84 |
| Temperate Climate | 75 | 106 | 61 | 0.82 | 39 | 46 | 157 | 0.85 |
| Cold Climate | 52 | 92 | 90 | 0.82 | 21 | 22 | 191 | 0.85 |
| Polar Climate | 0 | 0 | 2 | 0.97 | 0 | 0 | 2 | 0.97 |

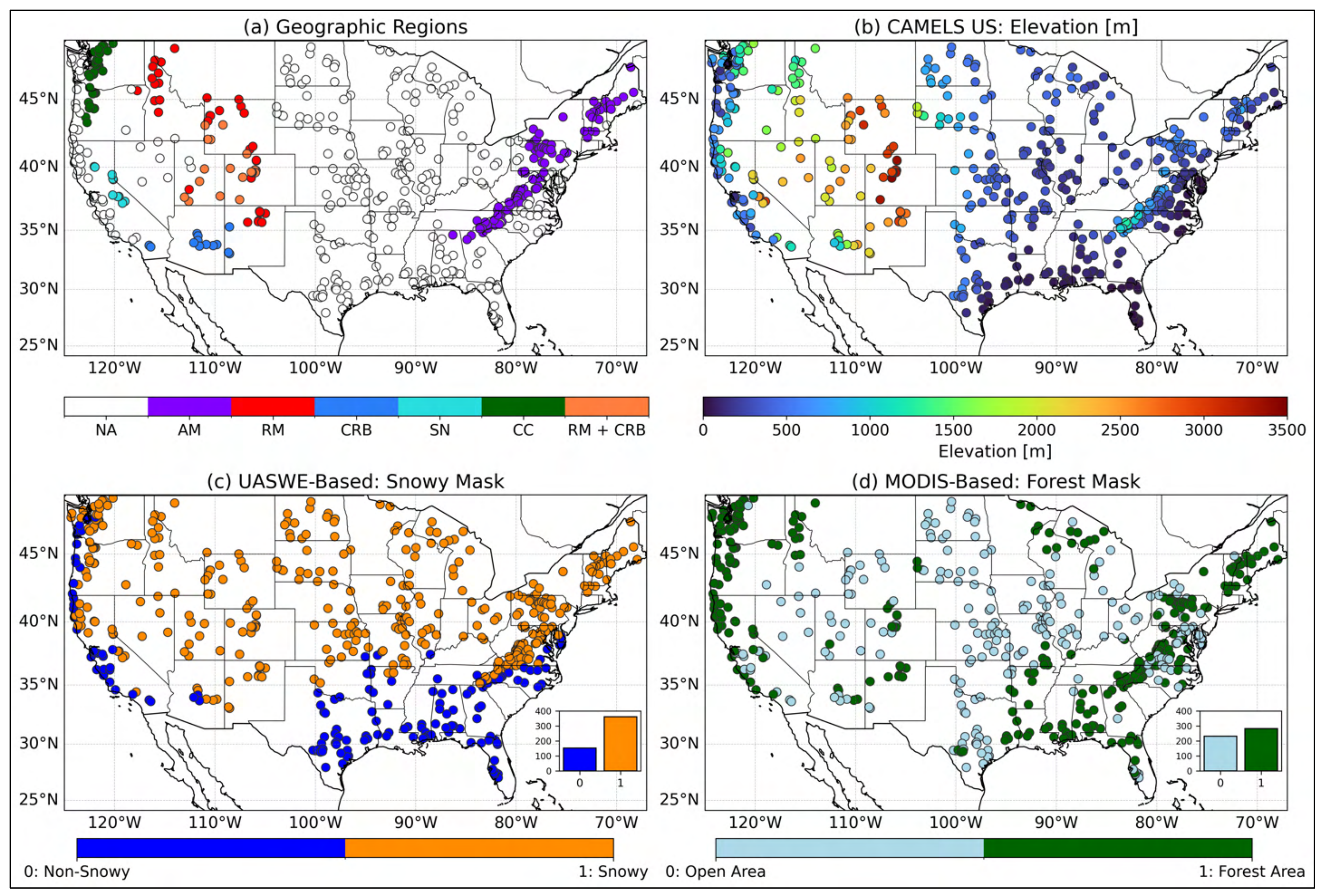

**Figure 1:** Geographical information for the 513 CAMELS-US catchments used in this study, including (a) selected geographic regions, (b) elevation (m), (c) snowy versus non-snowy basin delineation, and (d) forest versus open basin classification.

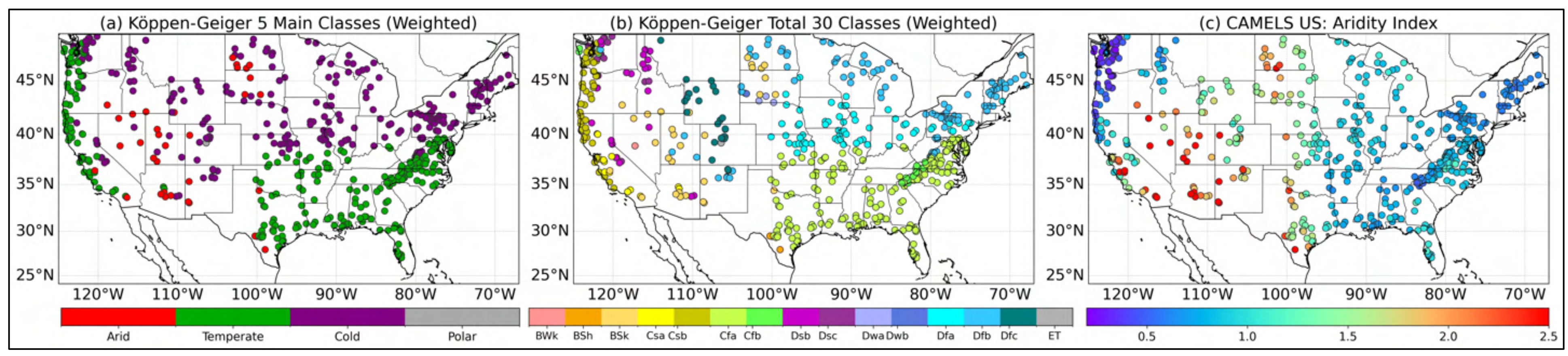

**Figure 2:** Spatial information for the 513 selected CAMELS-US catchments, including the Köppen–Geiger climate classification based on: (a) the five main classes, (b) the 30 detailed classes, and (c) the aridity index.

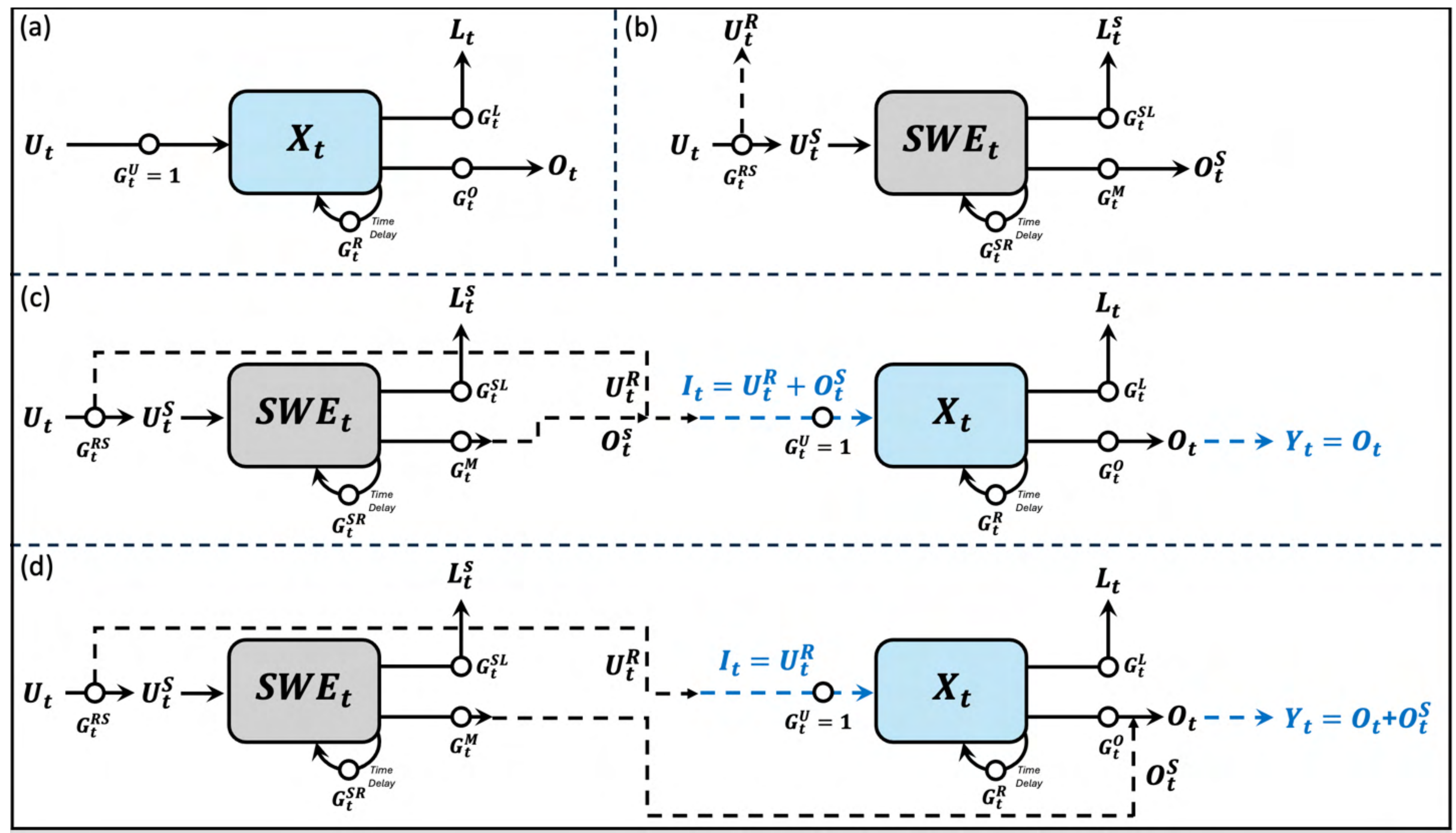

**Figure 3:** Direct-graph representations of the HMCP units and architectures used in this study, including (a) the Mass-Conserving Perceptron (MCP) unit proposed in Wang & Gupta (2024ab) and Wang & Gupta (2025), (b) the SNOWMCP unit proposed in this study, and the two HYDROMCP architectures that couple SOILMCP and SNOWMCP using (c) routing (serial connection) and (d) bypass (parallel connection)

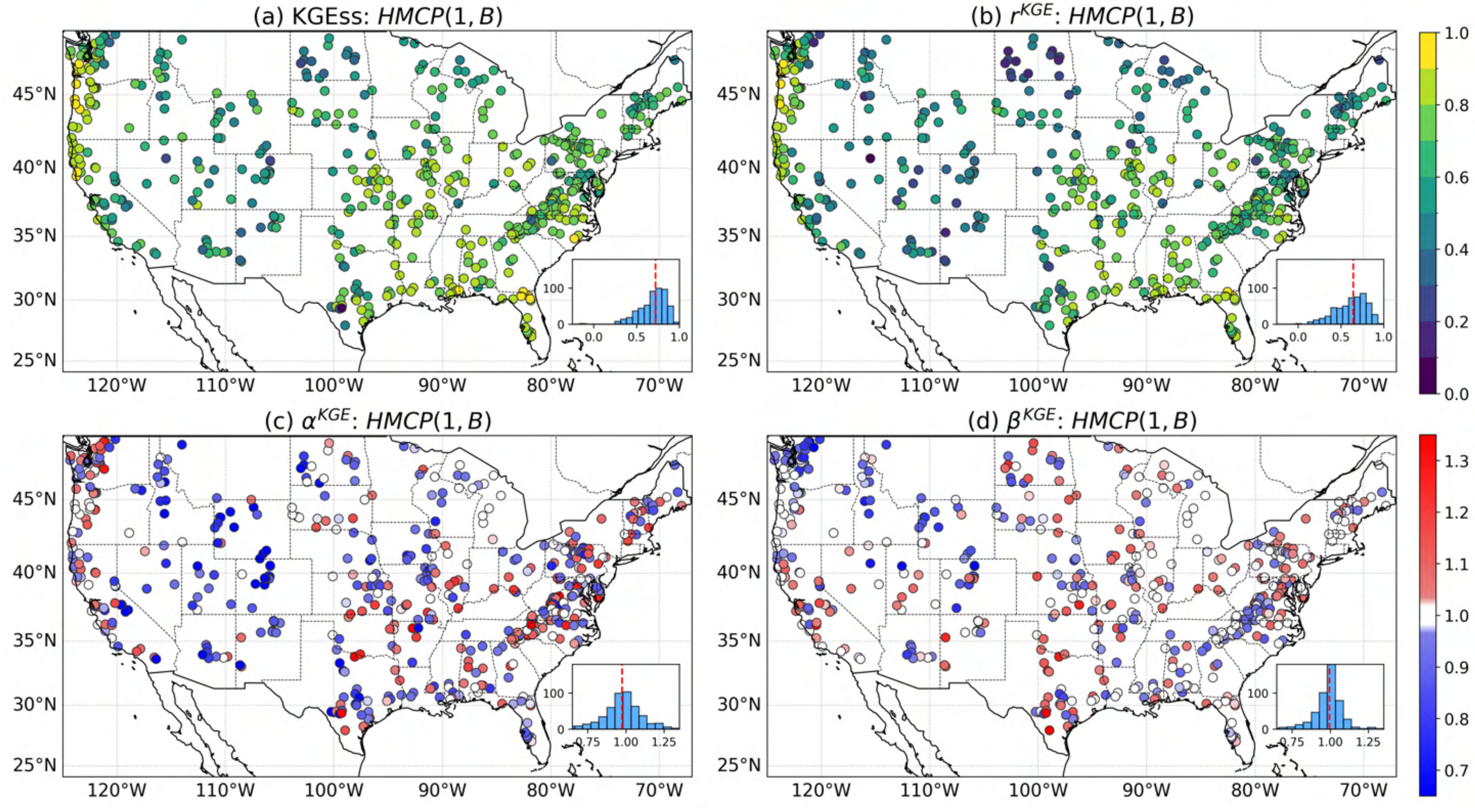

**Figure 4:** Skill metrics for HMCP(1,B) across 513 CONUS catchments, including (a) the KGE skill score ($KGE_{ss}$) and the three components of KGE: (b) linear correlation ($\gamma^{KGE}$), (c) flow variability ratio ($\alpha^{KGE}$) and (d) mass balance ratio ($\beta^{KGE}$).

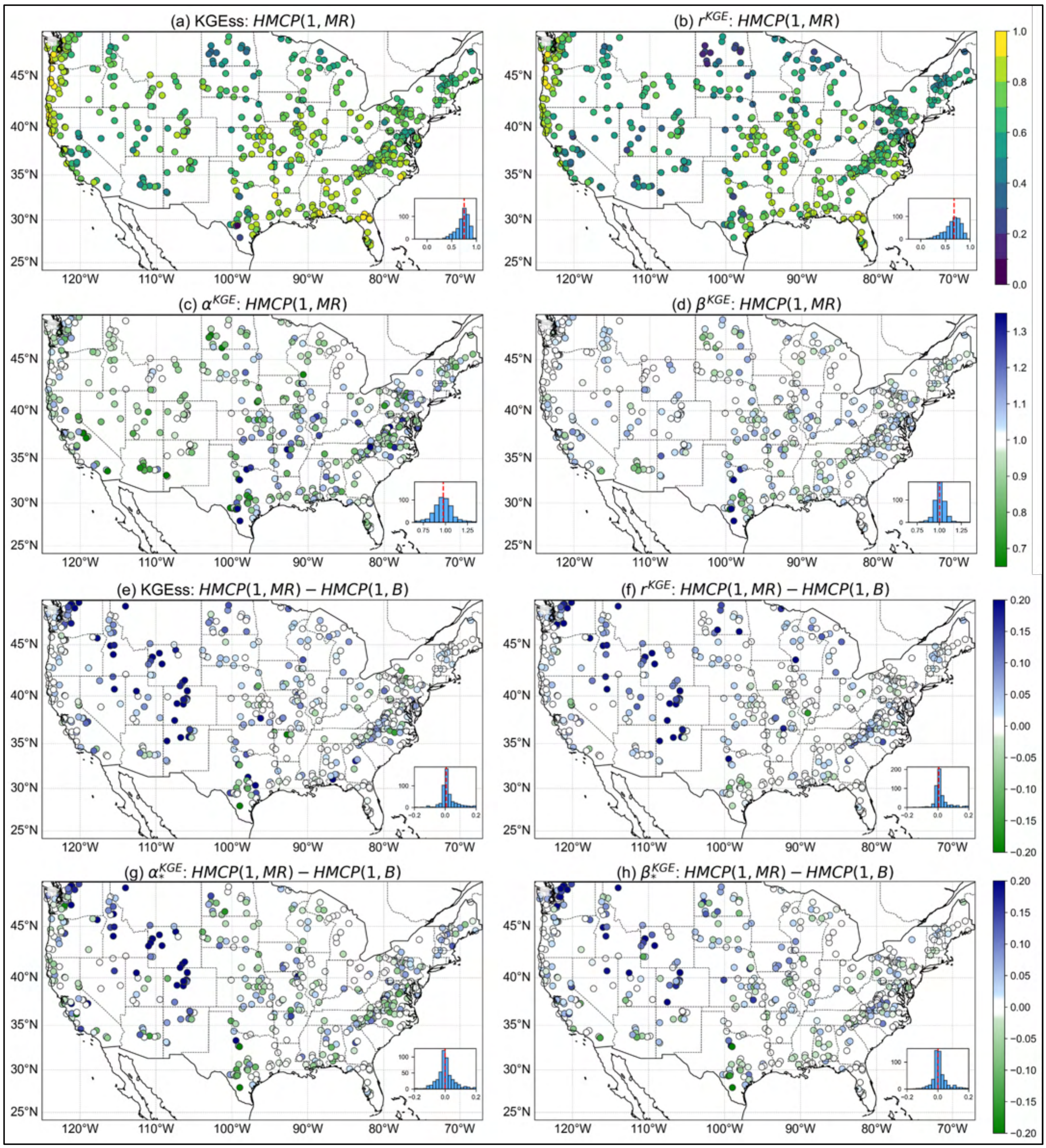

**Figure 5:** Skill metrics for HMCP(1,MR) across 513 CONUS catchments, including (a) the KGE skill score ($KGE_{ss}$) and the three components of KGE : (b) linear correlation ($\gamma^{KGE}$), (c) flow variability ratio ($\alpha^{KGE}$), and (d) mass balance ratio ($\beta^{KGE}$). Subplots (e–h) show the corresponding differences in these metrics between HMCP(1,MR) and HMCP(1,B).

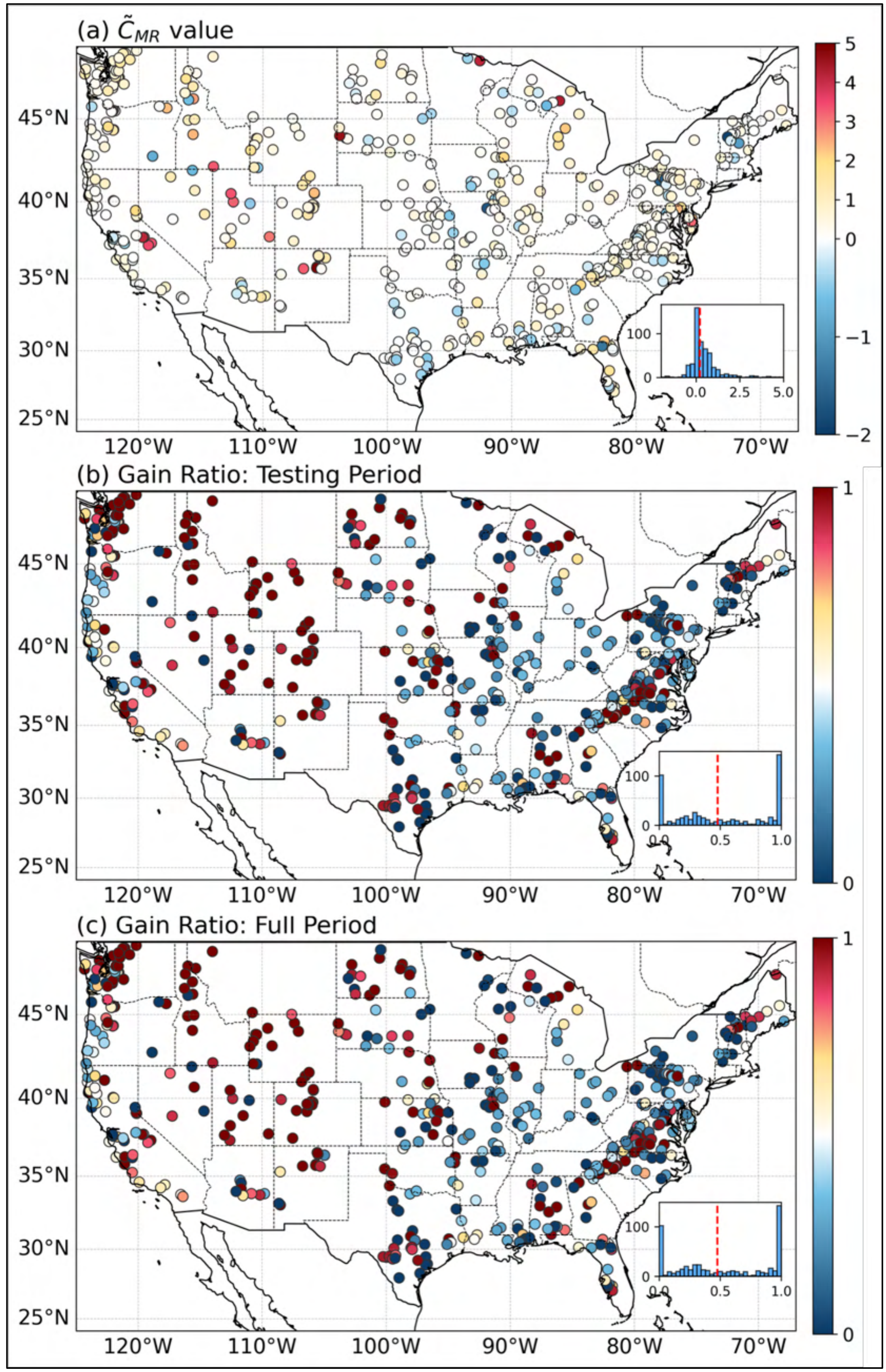

**Figure 6:** Spatial distribution of (a) the threshold $\tilde{C}_{MR}$ value learned in the HMCP(1,MR) architecture, and the associated gain ratio for the (b) testing period and (c) full time period.

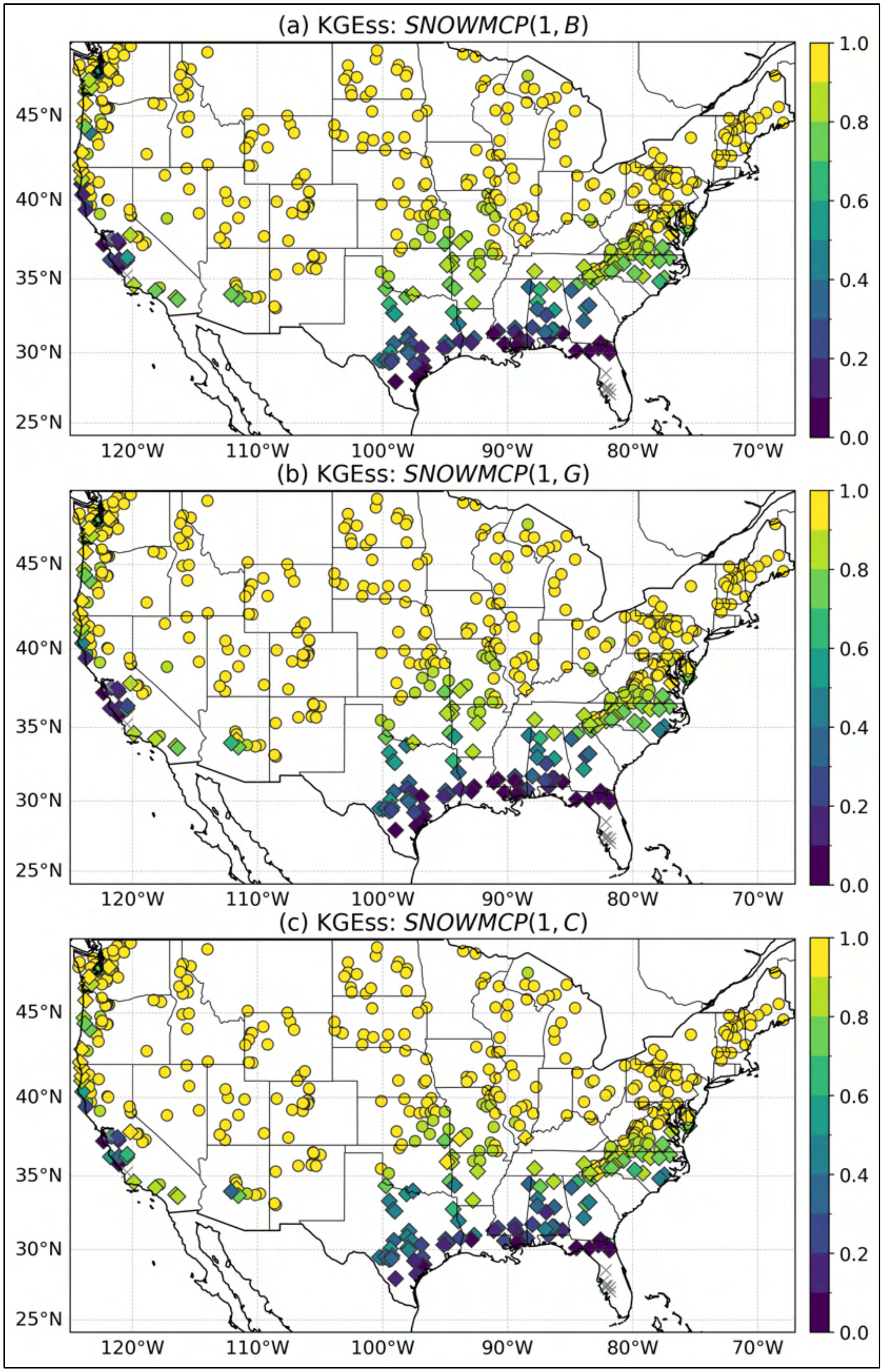

**Figure 7:** KGE skill score ( $KGE_{ss}$ ) across 513 CONUS catchments for three versions of the single-state SNOWMCP(1) architecture: (a) SNOWMCP(1,B), (b) SNOWMCP(1,G), and (c) SNOWMCP(1,C). Circles indicate snowy basins, and diamonds indicate non-snowy basins.

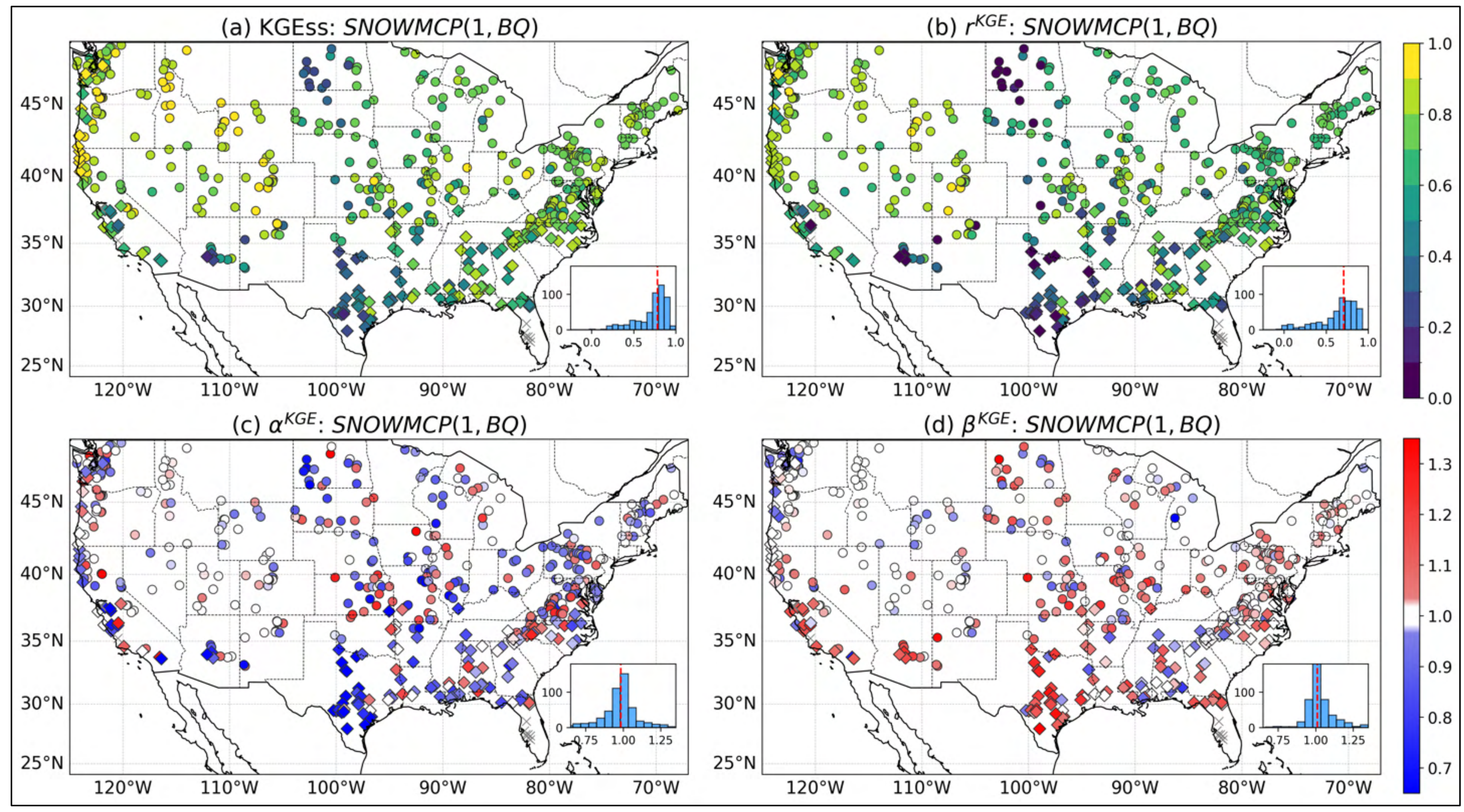

**Figure 8:** Skill metrics for SNOWMCP (1,BQ) across 513 CONUS catchments, including (a) the KGE skill score ($KGE_{ss}$) and the three components of KGE : (b) linear correlation ($\gamma^{KGE}$), (c) flow variability ratio ($\alpha^{KGE}$) and (d) mass balance ratio ($\beta^{KGE}$). Circles indicate snowy basins, and diamonds indicate non-snowy basins.

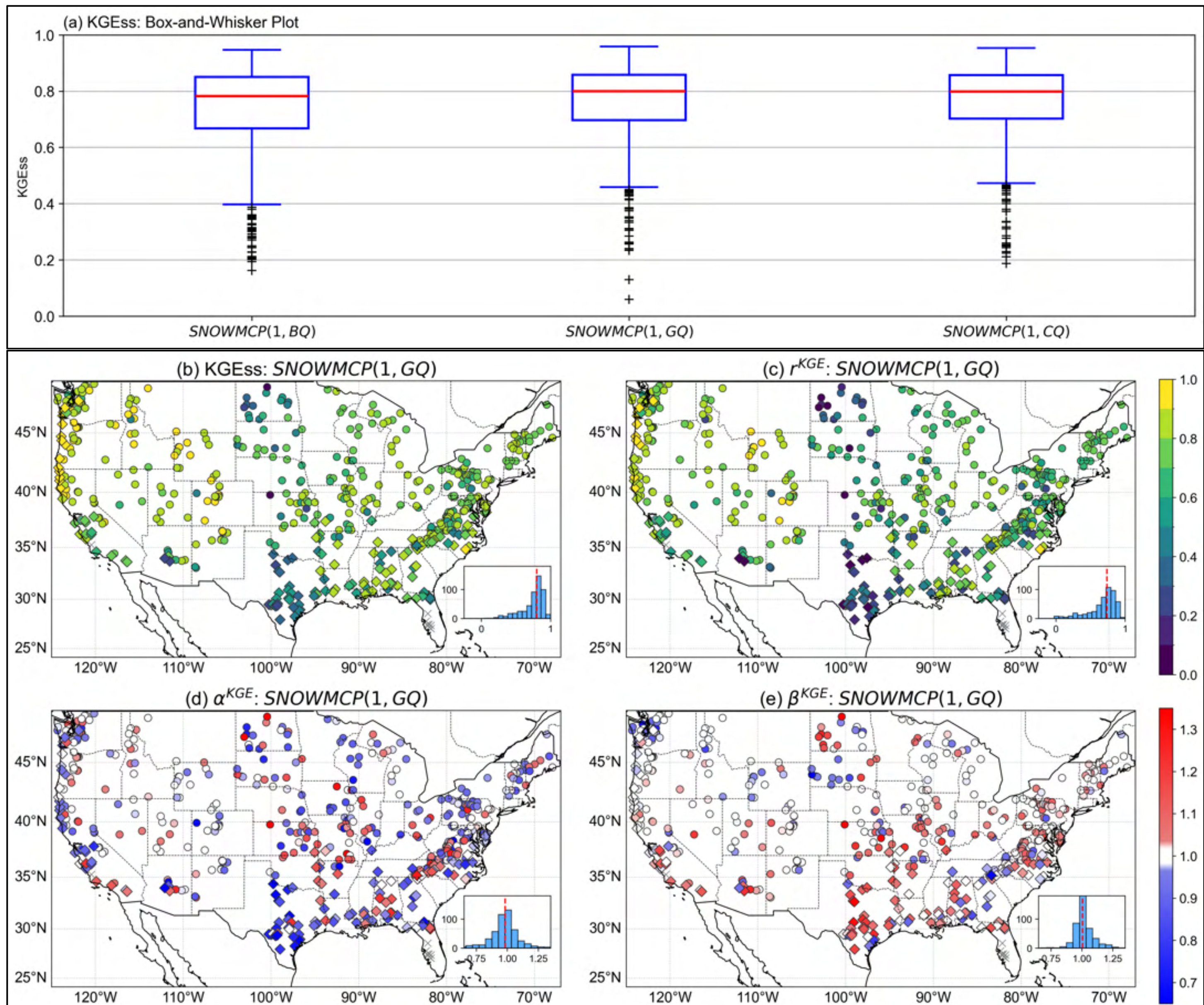

**Figure 9:** Performance of SNOWMCP in simulating streamflow, shown as: (a) Boxplots of the KGE skill score ($KGE_{ss}$) distribution for the three SNOWMCP architectures — SNOWMCP(1,BQ), SNOWMCP(1,GQ), and SNOWMCP(1,CQ) — across 513 catchments, and the spatial distribution of skill metrics for SNOWMCP(1,GQ), including (b) KGE skill score ($KGE_{ss}$) and the three components of KGE : (c) linear correlation ($\gamma^{KGE}$), (d) flow variability ratio ($\alpha^{KGE}$), and (e) mass balance ratio ($\beta^{KGE}$). Circles indicate snowy basins, and diamonds indicate non-snowy basins in subplots (b)–(e).

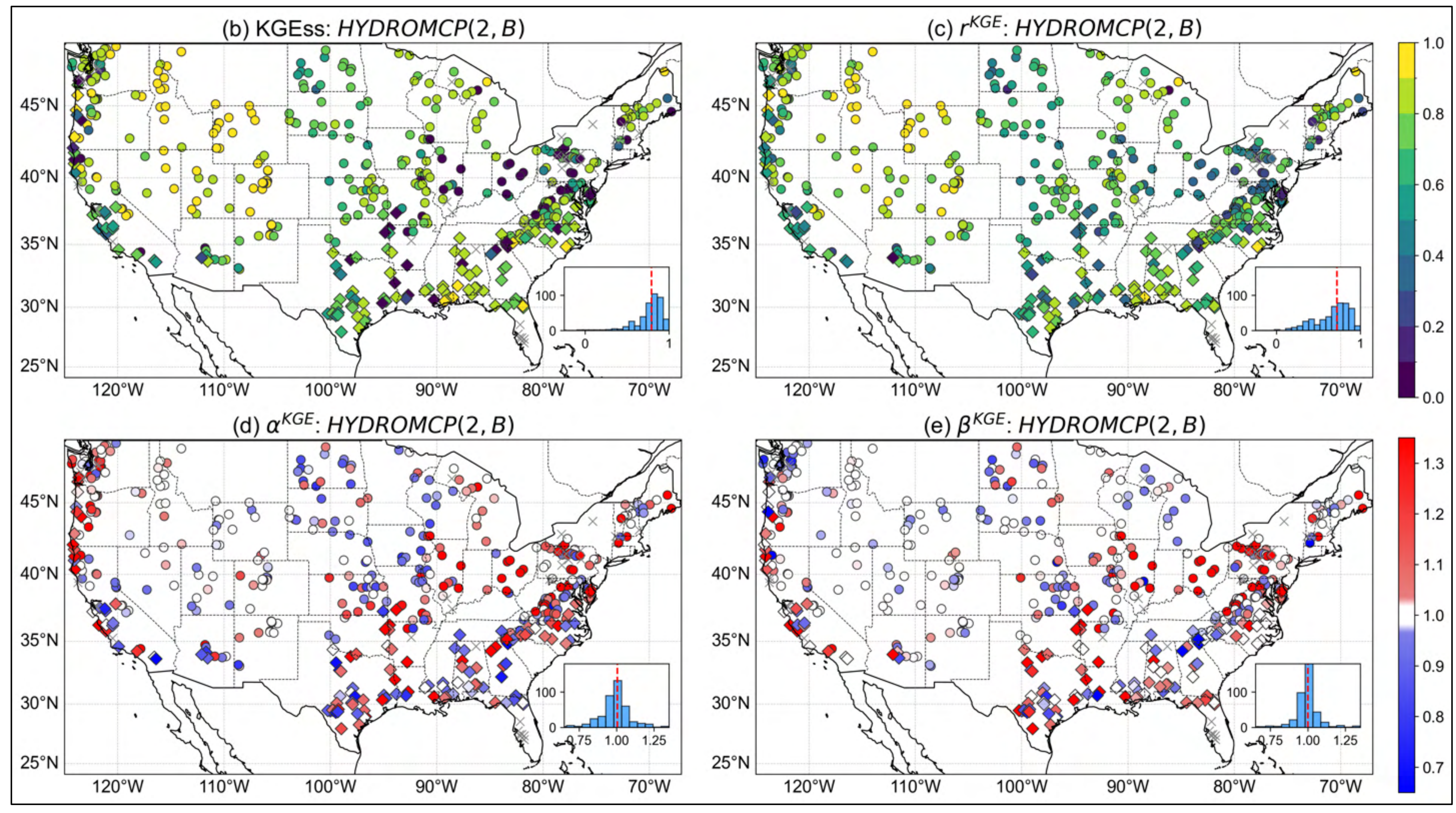

**Figure 10:** Skill metrics for HYDROMCP (2,B) across 513 CONUS catchments, including (a) the KGE skill score ($KGE_{ss}$) and the three componentsof KGE : (b) linear correlation ($\gamma^{KGE}$), (c) flow variability ratio ($\alpha^{KGE}$) and (d) mass balance ratio ($\beta^{KGE}$). Circles indicate snowy basins, and diamonds indicate non-snowy basins.

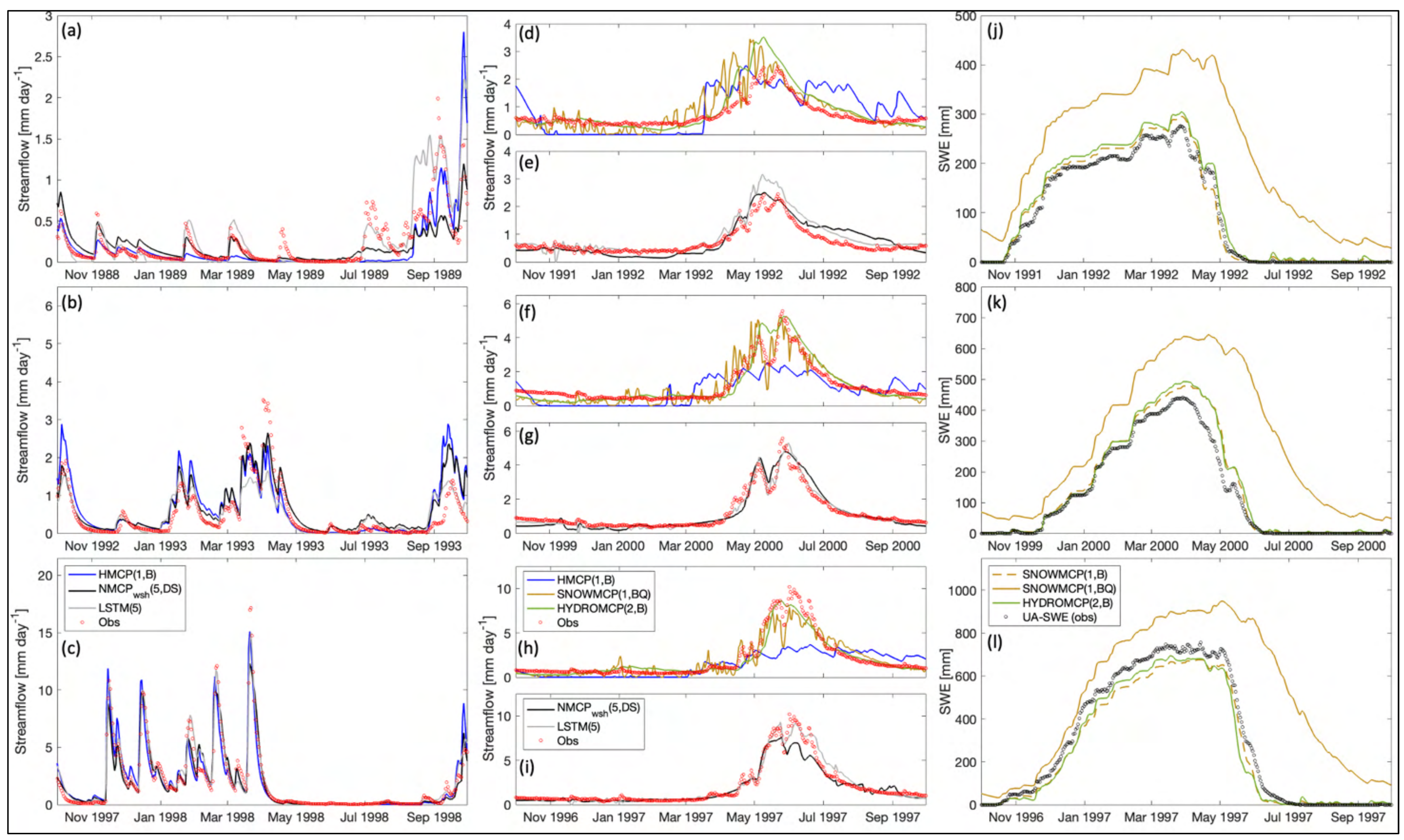

**Figure 11:** Time series under different flow regimes for selected CAMELS-US basins. Panels (a–c) show simulated and observed streamflow for basin 02296500 during dry, median, and wet water years (WY 1989, 1993, and 1998, respectively). Panels (d–i) show simulated and observed streamflow for basin 13023000 during dry, median, and wet water years (WY 1992, 2000, and 1997, respectively). Panels (j–l) show corresponding snow water equivalent (SWE) simulations and observations for basin 13023000.

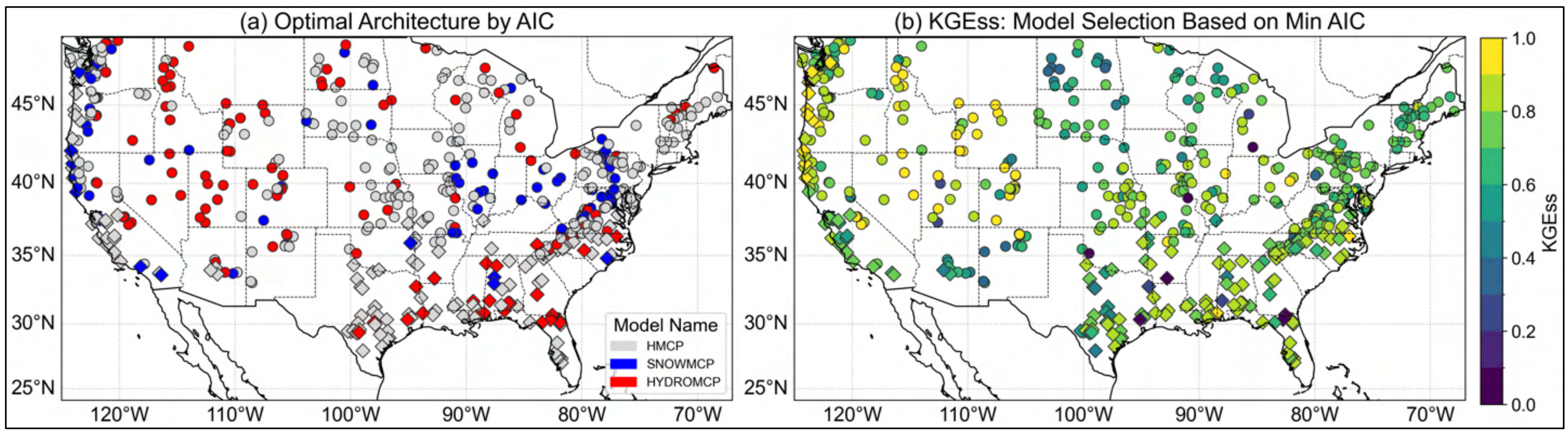

**Figure 12:** Optimal MCP-based architecture selection based on (a) the Akaike Information Criterion (AIC) and (b) the associated testing period KGE skill score ($KGE_{ss}$) across 513 basins. Circles indicate snowy basins, and diamonds indicate non-snowy basins.

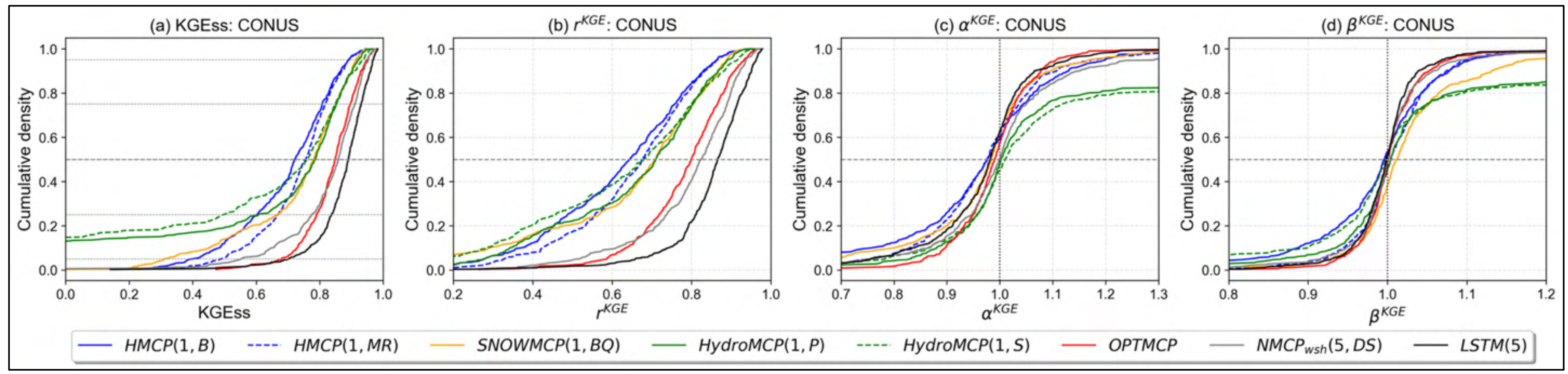

**Figure 13:** Empirical cumulative density plots of performance metrics across 513 basins, including (a) KGE skill score (KGEss), (b) linear correlation ($r^{KGE}$), (c) flow variability ratio ($\alpha^{KGE}$), and (d) mass balance ratio ($\beta^{KGE}$). Note that HYDROMCP(2,P) and HYDROMCP(2,S) are derived from the "Basic" case, HYDROMCP(2,B).

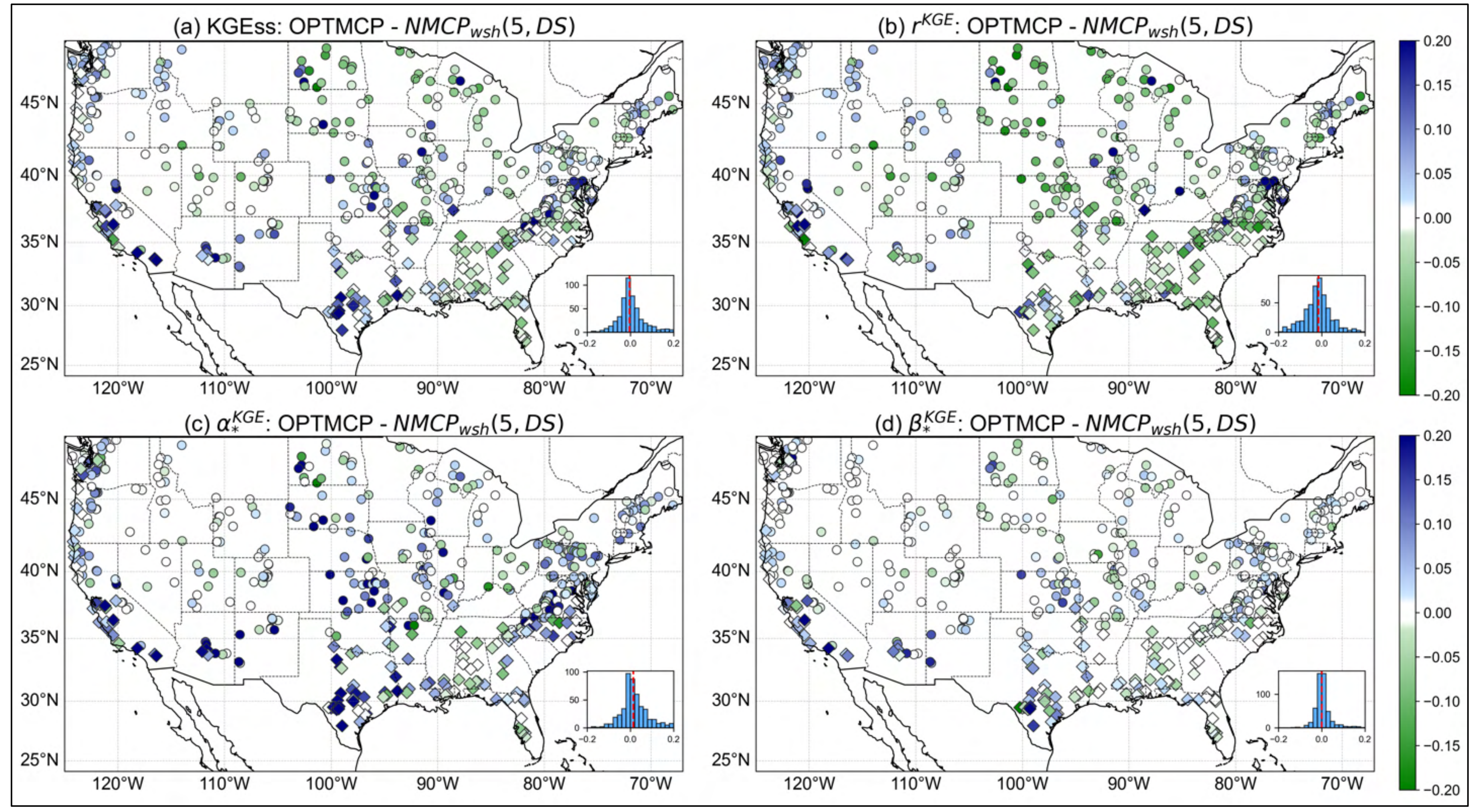

**Figure 14:** Performance differences between the OPTMCP architectures and the single-layer 5-node distributed-state mass-conserving network, shown for (a) the KGE skill score ($KGE_{ss}$), and the three components of KGE : (b) linear correlation ($r^{KGE}$), (c) adjusted flow variability ratio ($\alpha_*^{KGE}$), and (d) mass balance ratio ($\beta_*^{KGE}$). Circles indicate snowy basins, and diamonds indicate non-snowy basins.

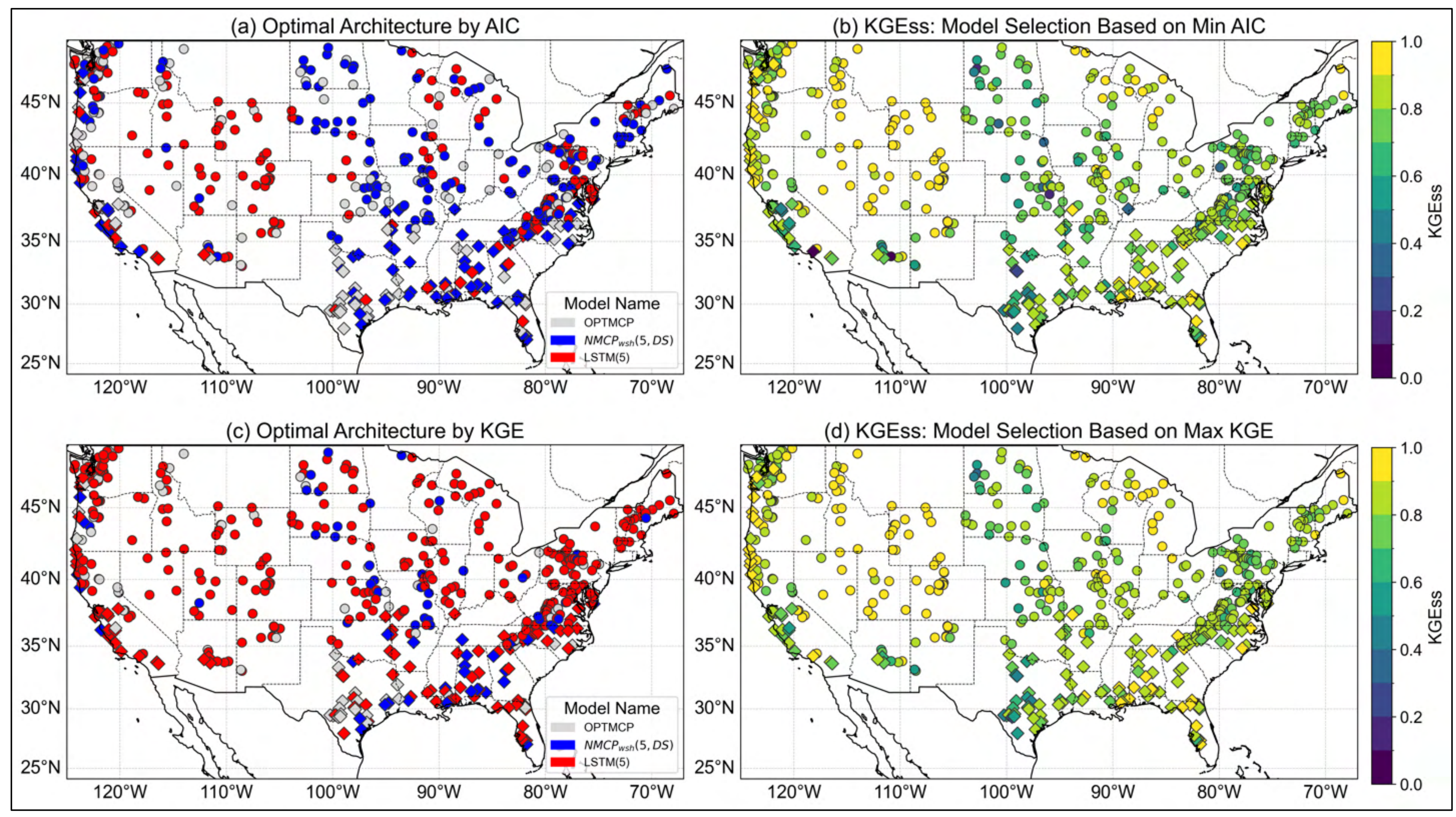

**Figure 15:** Optimal architecture selection among $OPTMCP$, $NMCP_{wsh}(5, DS)$, and $LSTM(5)$ based on (a) the Akaike Information Criterion (AIC) and (b) the associated testing period KGE skill score ($KGE_{ss}$) across 513 basins. Panels (c) and (d) show corresponding results when model selection is performed using KGE as the criterion instead of AIC. Circles indicate snowy basins, and diamonds indicate non-snowy basins.

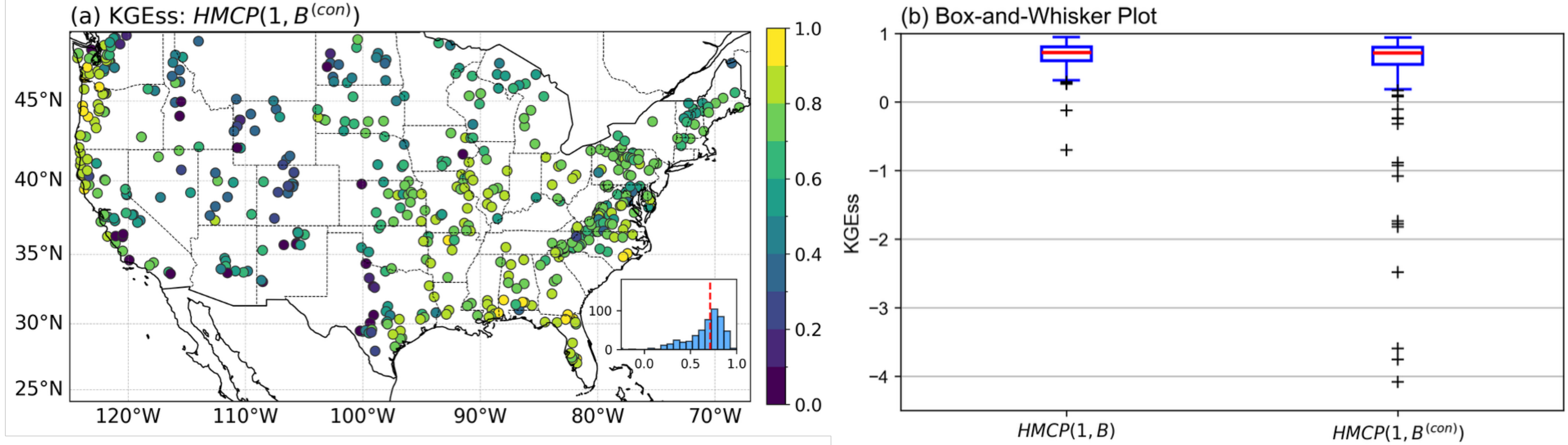

**Figure E1:** Performance of the PET-constrained $HCMP(1, B)$ model architecture, including: (a) spatial distribution of the KGE skill score ($KGE_{ss}$) for the PET-constrained $HCMP(1, B)$ model denoted as $HCMP(1, B^{(con)})$, and (b) box plot comparing PET-constrained $HCMP(1, B)$ model ($HCMP(1, B^{(con)})$) with the baseline $HCMP(1, B)$ model without PET constraint.

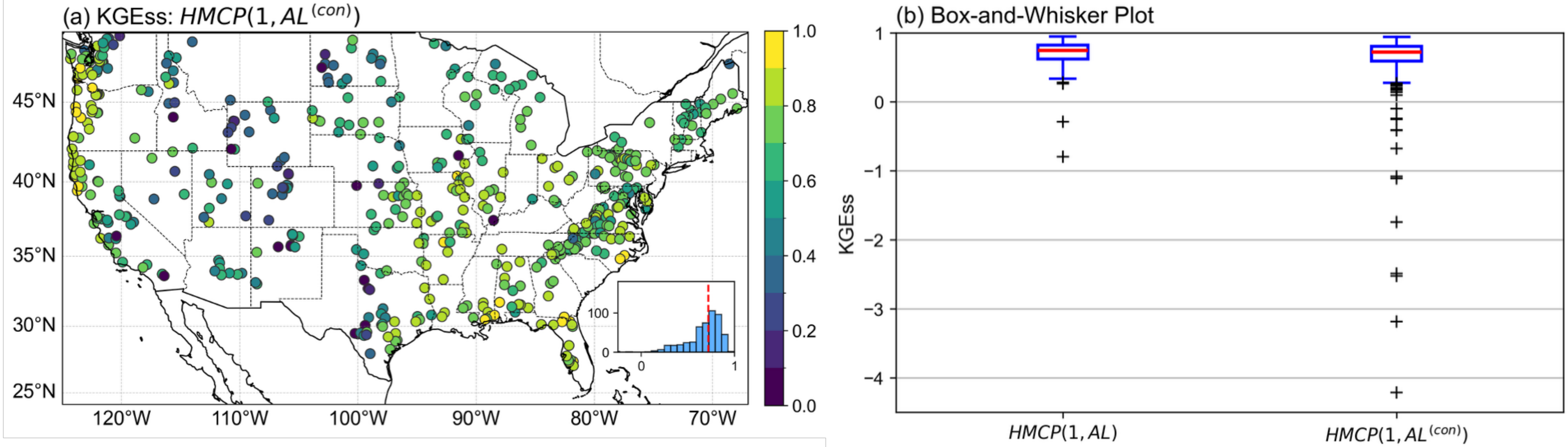

**Figure E2:** Performance of the PET-constrained architecture, including: (a) spatial distribution of the KGE skill score ($KGE_{ss}$) for the PET-constrained version of the $HCMP(1, AL)$ model denoted as $HCMP(1, AL^{(con)})$ , and (b) box plot comparing PET-constrained $HCMP(1, AL)$ model ($HCMP(1, AL^{(con)})$) with the loss gate augmented $HCMP(1)$ model without PET constraint $HCMP(1, AL)$.

Supporting Information for

# Towards CONUS-Wide ML-Augmented Conceptually-Interpretable Modeling of Catchment-Scale Precipitation-Storage-Runoff Dynamics

Yuan-Heng Wang[1,2], Yang Yang[3,4], Fabio Ciulla[1], Hoshin V Gupta[2], Charuleka Varadharajan[1]

[1] Environmental Sciences Area, Lawrence Berkeley National Laboratory, Berkeley, CA
[2] Department of Hydrology and Atmospheric Science, The University of Arizona, Tucson, AZ
[3] School for the Environment, University of Massachusetts Boston, MA
[4] Department of Civil Engineering, The University of Hong Kong, Hong Kong SAR, China

Corresponding Author: Yuan-Heng Wang, Ph.D.
Email: yhwang0730@gmail.com | YuanHengWang@lbl.gov | yhwang0730@arizona.edu

## Contents of this supplementary material:

Text S1 to S3
Table S1 to S9
Figures S1 to S22

## Introduction

This Supporting Information provides, 3 supplementary texts, 9 supplementary tables, and 22 supplementary figures to support the discussions in the main manuscript. The contents of these supplementary materials are as follows.

### Text

**Text S1.** Detailed Results for Basic Two-State-Variable Hydrologic Models (HYDROMCP)

**TextS2.** Detailed Results for Augmented Two-State-Variable Hydrologic Models (HYDROMCP)

**Text S3.** Summary of the AIC selection procedure conducted solely on the SOILMCP, SNOWMCP, and HYDROMCP model families.

### Table

**Table S1.** Summary of parameter initialization, learning rates and number of epochs used for each network configuration

**Table S2.** Summary of $r^{KGE}$ metric for the MCP-based architectures across geographic regions in the CONUS

**Table S3.** Summary of $\alpha_{*}^{KGE}$ metric for the MCP-based architectures across geographic regions in the CONUS

**Table S4.** Summary of $\beta_{*}^{KGE}$ metric for the MCP-based architectures across geographic regions in the CONUS

**Table S5.** Summary of streamflow and SWE performance metrics for the two selected basins, focusing on the additional $SNOWMCP(1, BQ)^{*}$ case and comparing it with other SNOWMCP and HYDROMCP cases.

**Table S6.** Summary of the number of basins selected by the AIC criterion in individual experiments with SOILMCP, SNOWMCP, and HYDRO-MCP.

**Table S7.** Summary of $KGE_{ss}$ selected by the AIC criterion in individual experiments with SOILMCP, SNOWMCP, and HYDRO-MCP.

**Table S8.** Summary of the number of basins selected by the KGE criterion in individual experiments with SOILMCP, SNOWMCP, and HYDRO-MCP.

**Table S9.** Summary of $KGE_{ss}$ selected by the KGE criterion in individual experiments with SOILMCP, SNOWMCP, and HYDRO-MCP.

**Figure**

**Figure S1:** Geographic regions highlighted for discussion in this study, including the Appalachian Mountains (AM), Rocky Mountains (RM), Colorado River Basin (CRB), Sierra Nevada (SN), and the Cascades (CC).

**Figure S2:** Information used to derive the Snowy Mask: (a) University of Arizona (UA) snow product (snow water equivalent; SWE; Broxton et al., 2019)–based mask derived using a 3 mm annual maximum SWE threshold, (b) the 3 mm annual maximum SWE threshold itself, (c) CAMELS-US snow fraction, and (d) the final combined criterion, where non-snowy areas are defined as having an annual maximum SWE ≤ 3 mm and a CAMELS-US snow fraction < 5%.

**Figure S3:** Information used to derive the Forest-Covered Mask: (a) MODIS-based land cover classes (Broxton et al., 2014), (b) majority-upscaled binary forest cover derived from (a), (c) forest fraction derived from (a), and (d) CAMELS-US forest fraction.

**Figure S4:** Information used to derive the Köppen–Geiger Climatologic Mask (Beck et al., 2023): (a) classification for 1961–1990, (b) classification for 1991–2020, and (c) weighted classification derived from (a) and (b) using linear proportional weighting, consistent with the simulation period (WY 1982–2008).

**Figure S5:** Skill metrics for HMCP(1,AL) across 513 CONUS catchments, including (a) the KGE skill score ($KGE_{ss}$) and the three components of KGE: (b) linear correlation ($\gamma^{KGE}$), (c) flow variability ratio ($\alpha^{KGE}$), and (d) mass balance ratio ($\beta^{KGE}$). Subplots (e–h) show the corresponding differences in these metrics between HMCP(1,AL) and HMCP(1,B).

**Figure S6:** UA SWE–derived snow signatures for the study period (WY 1982–2008), including the annual median of (a) maximum SWE (mm), (b) April 1 SWE, (c) date of maximum SWE occurrence, (d) first day of SWE, (e) last day of SWE, and (f) snowy season length (difference between first and last day of SWE).

**Figure S7:** Performance differences between SNOWMCP(1,GQ) and SNOWMCP(1,BQ) for the following metrics: (a) KGE skill score (KGEss) and the three components of KGE, (b) linear correlation ($r^{KGE}$), (c) adjusted flow variability ratio ($\alpha_*^{KGE}$), and (d) mass balance ratio ($\beta_*^{KGE}$). Circles indicate snowy basins, and diamonds indicate non-snowy basins. The symbol × indicates where the model failed to finish training.

**Figure S8:** Performance of SNOWMCP(1,CQ) for the following metrics: (a) KGE skill score (KGEss) and the three components of KGE, (b) linear correlation ($r^{KGE}$), (c) adjusted flow variability ratio ($\alpha^{KGE}$), and (d) mass balance ratio ($\beta^{KGE}$). Panels (e)–(h) and (i)–(l) summarize the performance differences of SNOWMCP(1,CQ) against SNOWMCP(1,BQ) and SNOWMCP(1,GQ), respectively. Circles indicate snowy basins, and diamonds indicate non-snowy basins. The symbol × indicates where the model failed to finish training.

**Figure S9:** Success and failure flags for HYDRO-MCP(2) model training under three strategies: (a, c, e) serial (S) configuration and (b, d, f) parallel (P) configuration. S and P also refer to routing and bypass in the context of coupling HMCP and SNOWMCP. Gray circles indicate successful training and blue crosses indicate failed training.

**Figure S10:** Spatial maps of performance metrics, including KGE skill score (KGEss) and the three components of KGE: linear correlation ($r^{KGE}$), flow variability ratio ($\alpha^{KGE}$), and mass balance ratio ($\beta^{KGE}$). Subplots (a–d) summarize HYDRO-MCP(2,S) under the first training strategy, (e–h) HYDRO-MCP(2,P) under the first strategy, (i–l) HYDRO-MCP(2,S) under the second strategy, (m–p) HYDRO-MCP(2,P) under the second strategy, (q–t) HYDRO-MCP(2,S) under the third strategy, and (u–x) HYDRO-MCP(2,P) under the third strategy. Here, S and P denote Serial and Parallel configurations, respectively. Circles indicate snowy basins, diamonds indicate non-snowy basins, and blue crosses (×) mark catchments where model training failed to complete.

**Figure S11:** Spatial information for the HYDRO-MCP(2,B) model showing (a) the best training strategy, (b) the better-performing series (S) versus parallel (P) hypothesis, and (c) the best model hypothesis and training strategy for each catchment. Roman numerals denote training strategies. Circles indicate snowy basins, diamonds indicate non-snowy basins, and blue crosses (×) mark catchments where model training failed to complete.

**Figure S12:** Spatial information on the performance differences between HYDRO-MCP(2,$B^{+}$) and HYDRO-MCP(2,$B$), with locations with large differences highlighted in color.

**Figure S13:** Skill metrics and component-wise comparisons for alternative $HYDROMCP$ architectures across 513 CONUS catchments. Panel (a) shows box-and-whisker plots of the KGE skill score ($KGE_{ss}$) for $HYDROMCP(2,B)$, $HYDROMCP(2,G)$, and $HYDROMCP(2,C)$. Panels (b)–(e) show the spatial distributions of $KGE_{ss}$, the linear correlation component ($r^{KGE}$), the flow variability ratio ($\alpha^{KGE}$), and the mass balance ratio ($\beta^{KGE}$) for $HYDROMCP(2,G)$. Panels (f)–(i) show the corresponding differences between $HYDROMCP(2,G)$ and $HYDROMCP(2,B)$. Panels (j)–(m) show the same four metrics for $HYDROMCP(2,C)$, and panels (n)–(q) show the corresponding differences between $HYDROMCP(2,C)$ and $HYDROMCP(2,B)$. Circles indicate snowy basins, and diamonds indicate non-snowy basins.

**Figure S14:** Spatial information on the performance differences between (a) HYDRO-MCP(2,G) and HYDRO-MCP(2,B), (b) HYDRO-MCP(2,C) and HYDRO-MCP(2,B), and (c) HYDRO-MCP(2,C) and HYDRO-MCP(2,G), with locations with large differences highlighted in color.

**Figure S15:** Time series generated by selected SNOWMCP-based models under different flow regimes for CAMELS-US basin 13023000. Panels (a–c) show simulated and observed streamflow during dry, median, and wet water years (WY 2007, WY 1993, and WY 1998, respectively). Panels (d–f) show the corresponding snow water equivalent (SWE) simulations and UA-SWE observations for the same water years. The models include SNOWMCP(1,B), SNOWMCP(1,BQ), and SNOWMCP(1,BQ)*, where SNOWMCP(1,BQ)* denotes the model trained jointly against streamflow and SWE using KGE.

**Figure S16:** Box–whisker plots of the $KGE_{ss}$ metric for SWE simulations across CAMELS-US basins under increasingly restrictive snow-dominance criteria. Results are grouped by (a) annual maximum SWE thresholds and (b) snowy-season-length (SSL) thresholds. SSL denotes snowy season length, defined here as the median annual difference between the last snowy day and the first snowy day within WY1982–2008, where a snowy day is defined as SWE > 3 mm.

**Figure S17:** Spatial distribution of (a) model selection results based on the minimum-AIC metric for the HMCP family of models and their associated $KGE_{ss}$ values in (b), and (c) model selection results based purely on the maximum-KGE metric and their associated KGEss values in (d). Corresponding results for SNOWMCP are shown in subplots (e–h), and for HYDRO-MCP in subplots (i–l).

**Figure S18:** Box–whisker plots of the $KGE_{ss}$ metric across 513 basins for (a) HMCP, (b) SNOWMCP, and (c) HYDRO-MCP. For each model category, results are shown for both minimum-AIC–based selection and maximum-KGE–based selection

**Figure S19:** Box–whisker plots of $KGE_{ss}$ across CAMELS-US basin groups defined by geographical region, mountain region, elevation, snow condition, land cover, climate class, and aridity regime. Each panel shows the distribution of $KGE_{ss}$ for the corresponding basin group, with the number of basins indicated in parentheses.

**Figure S20:** Box–whisker plots of $\alpha_*^{KGE}$ across CAMELS-US basin groups defined by geographical region, mountain region, elevation, snow condition, land cover, climate class, and aridity regime. Each panel shows the distribution of $\alpha_*^{KGE}$ for the corresponding basin group, with the number of basins indicated in parentheses. Note that $\alpha_*^{KGE} = 1 - |1 - \alpha^{KGE}|$, which is optimized at 1 and is unbounded below.

**Figure S21:** Box–whisker plots of $\beta_*^{KGE}$ across CAMELS-US basin groups defined by geographical region, mountain region, elevation, snow condition, land cover, climate class, and aridity regime. Each panel shows the distribution of $\beta_*^{KGE}$ for the corresponding basin group, with the number of basins indicated in parentheses. Note that $\beta_*^{KGE} = 1 - |1 - \beta^{KGE}|$, which is optimized at 1 and is unbounded below.

**Figure S22:** Box–whisker plots of $\gamma^{KGE}$ across CAMELS-US basin groups defined by geographical region, mountain region, elevation, snow condition, land cover, climate class, and aridity regime. Each panel shows the distribution of $\gamma^{KGE}$ for the corresponding basin group, with the number of basins indicated in parentheses.

## Text S1. Detailed Results for Basic Two-State-Variable Hydrologic Models (HYDROMCP)

The purpose of developing the full suite of HYDROMCP models — comprising both soil moisture storage and snowpack components—is to account for the influence of snowpack on streamflow, including early spring snowmelt and rain-on-snow events, which can significantly affect the shape and timing of the downstream hydrograph in snowy catchments. It should be noted that the metrics reported below for the HYDROMCP modeling exercises exclude basins in which training did not successfully converge.

For all six cases discussed in the **Section 3.4.1** of main text, not all of the $513$ basins successfully completed training. In the model symbol below, *S* and *P* stand for "serial" routing and "parallel" bypass, respectively. As reported in **Figure S9**, we found that the number of catchments for which the model successfully completed training includes $443$ and $466$ catchments for $HYDROMCP(2, S_I)$ and $HYDROMCP(2, P_I)$, $445$ and $467$ for $HYDROMCP(2, S_{II})$ and $HYDROMCP(2, P_{II})$, and $453$ and $471$ for $HYDROMCP(2, S_{III})$ and $HYDROMCP(2, P_{III})$, respectively. The Roman numerals used in the superscripts denote the training strategy type. Among the six coupled cases, we found that $474$ out of $513$ catchments were able to complete at least one of the coupled models runs. The "parallel" hypothesis appears to cause fewer issues during training than the "serial" hypothesis, supporting earlier findings that surface routing does not notably improve performance (*Wang & Gupta, 2024b*) and that, from a neural network perspective, placing the nodes in parallel is more ideal than placing them in series (*Wang & Gupta, 2025*).

Note that $HYDROMCP(2, B)$ refers to the two-state coupled version of the standard SOILMCP component $HMCP(1, B)$ and the simplest SNOWMCP component $SNOWMCP(1, B)$. The $HYDROMCP(2, B)$ is composed of six cases mentioned above including $HYDROMCP(2, S_I)$, $HYDROMCP(2, P_I)$, $HYDROMCP(2, S_{II})$, $HYDROMCP(2, P_{II})$, $HYDROMCP(2, S_{III})$, and $HYDROMCP(2, P_{III})$.

The $HYDROMCP(2, S_I)$ and $HYDROMCP(2, P_I)$ models, in which the snow component parameters are fixed during training against streamflow, shows overall lower skill than the results derived using second and third strategies. This is because they assume that the SNOW-MCP can perfectly represent the contribution of snowmelt to streamflow. As a result, the $KGE_{ss}^{50\%} = 0.65$ for $HYDROMCP(2, S_I)$ and $0.59$ for $HYDROMCP(2, P_I)$ are lower than those derived using second and third strategies, where the $KGE_{ss}^{50\%} \geq 0.73$. The performance of $HYDROMCP(2, S_I)$ and $HYDROMCP(2, P_I)$ (**Figure S10a-h**) is notably worse than that of $HYDROMCP(2, S_{II})$ and $HYDROMCP(2, P_{II})$ (**Figure S10i-p**) and $HYDROMCP(2, S_{III})$ and $HYDROMCP(2, P_{III})$ (**Figure S10q-x**) over the Missouri region and the mountainous areas of the U.S.

The $HYDROMCP(2, S_{II})$ and $HYDROMCP(2, P_{II})$ models show slightly higher accuracy than that of $HYDROMCP(2, S_{III})$ and $HYDROMCP(2, P_{III})$. When comparing $KGE_{ss}^{50\%}$ skill, $HYDROMCP(2, S_{II})$ equals $0.75$, which is higher than the $0.73$ for $HYDROMCP(2, S_{III})$. Similarly, the $KGE_{ss}^{50\%}$ skill for $HYDROMCP(2, P_{II})$ equals $0.78$ which is higher than the $0.74$ for $HYDROMCP(2, P_{III})$. Performance over snowy pixels further improves in all four cases, with $KGE_{ss}^{50\%}$ increasing to $0.84$ for $HYDROMCP(2, S_{II})$, $0.85$ for $HYDROMCP(2, P_{II})$, $0.82$ for $HYDROMCP(2, S_{III})$, and $0.83$ for $HYDROMCP(2, P_{III})$. This suggests that training solely on streamflow can still enhance model performance compared to cases in which streamflow and snowpack are weighted equally in the objective function. We refer readers to **Figure S11** for additional details on the spatial distribution of performance.

Furthermore, the $HYDROMCP(2, S_{II})$ and $HYDROMCP(2, P_{II})$ model achieved the best testing period $KGE_{ss}$ skill, accounting for 341 out of 474 successful runs. In contrast, the $HDROMCP(2, S_I)$, $HYDROMCP(2, P_I)$, $YDROMCP(2, S_{III})$ and $HYDROMCP(2, P_{III})$ models only performed well at 59 and 74 catchments, respectively (**Figure S11a**). Among the best cases, 272 out of 474 catchments show higher support for the hypothesis of snowmelt bypass (parallel configuration), while the remaining 202 catchments support using snowmelt as input to the SOILMCP (series configuration). So, overall, the two hypotheses are supported by a comparable number of catchments (**Figure S11b**). No further insights were found regarding why a specific training method or the series/parallel hypothesis works for certain locations (**Figure S11c**). We recommend that future studies should further investigate this topic as one of the key focal directions concerning model architectural hypotheses.

We summarize the corresponding best achieved testing period $KGE_{ss}$ in **Figure 10**. Across the $474$ catchments, the largest median $KGE_{ss}$ achieved among all six of the cases was $0.79$ (**Figure 10a**), with corresponding values of $0.80$ for snowy basins and $0.74$ for non-snowy basins. Here, we observe an error pattern in which the poorly performing purple dots in **Figure 8a** (where $KGE_{ss}$ is between $0$ to $0.1$) are associated with a combination of low to moderate linear correlation ($0.2$ - $0.6$; **Figure 10b**), high flow variability ratio (**Figure 10c**), and overestimation of flow mass balance. Presumably, this is due to misrepresentation of snowpack dynamics (**Figure 10d**).

Furthermore, while the median $KGE_{ss} = 0.82$ in the forest area is higher than in the open area of $0.77$, we observe that model training failed to complete in certain locations, notably in Florida and parts of the eastern CONUS. On average, $69\%$ of the basins that failed across the six models are located in forested regions, with a substantial number in the SN and CC, and particularly in the AM (see blue labels in **Figure S9**). This suggests that our simple HYDROMCP architecture is not adequate to the task of characterizing the full spectrum of land surface processes in forested areas, and highlights the need for future work to improve the representation of vegetation dynamics.

## Text S2. Detailed Results for Augmented Two-State-Variable Hydrologic Models (HYDROMCP)

Given that the $HMCP(1,AL)$ demonstrates better performance than the original $HMCP(1,B)$ as summarized in **Section 3.1**, we examine the performance of coupling this loss-gate-augmented $HMCP(1,AL)$ with the basic SNOWMCP $SNOWMCP(1,B)$ (**Section 3.3.1**), whereby this HYDROMCP version of architecture is denoted as $HYDROMCP(2,B^{+})$. The "+" symbol represents the cell-state is incorporated into the loss gate as opposed to only PET in the $HMCP(1,B)$ .

**Figure ST1a** shows the best achievable $KGE_{ss}$ score across the country obtained by the $HYDROMCP(2,B_{I}^{+})$, $HYDROMCP(2,B_{II}^{+})$ and $HYDROMCP(2,B_{III}^{+})$ architectures, along with the three associated $KGE$ components (**Figure ST1b-d**). Overall, $HYDROMCP(2,B^{+})$ has a similar spatial distribution of KGE-based skill compared to $HYDROMCP(2,B)$ . We observe both increases and decreases in skill. Specifically, 78 and 156 catchments have $KGE_{ss}$ differences greater than $0.05$ and $0.01$ , respectively, whereas 86 and 157 catchments have $KGE_{ss}$ differences smaller than $-0.05$ and $-0.01$, respectively.

As shown in **Figure S12**, performance improvements are located mostly in the eastern US with several Midwest basins showing substantial improvement (blue dots in **Figure ST1e–f**). For example, 52 basins show a $KGE_{ss}$ skill improvement exceeding $0.1$, with 44 of these —having an average aridity index of 0.66— located east of the 100th meridian.

Although using $HMCP(1,AL)$ as the SOILMCP component improved model performance — increasing the number of catchments completing training from 474 to 484 — the $KGE_{ss}$ box-whisker plot indicates that the number of the poor-performed basins increased (**Figure ST2**). Therefore, given the broad focus of this manuscript, we did not pursue further exploration of the improvement produced by the $HMCP(1,AL)$ SOILMCP architecture.

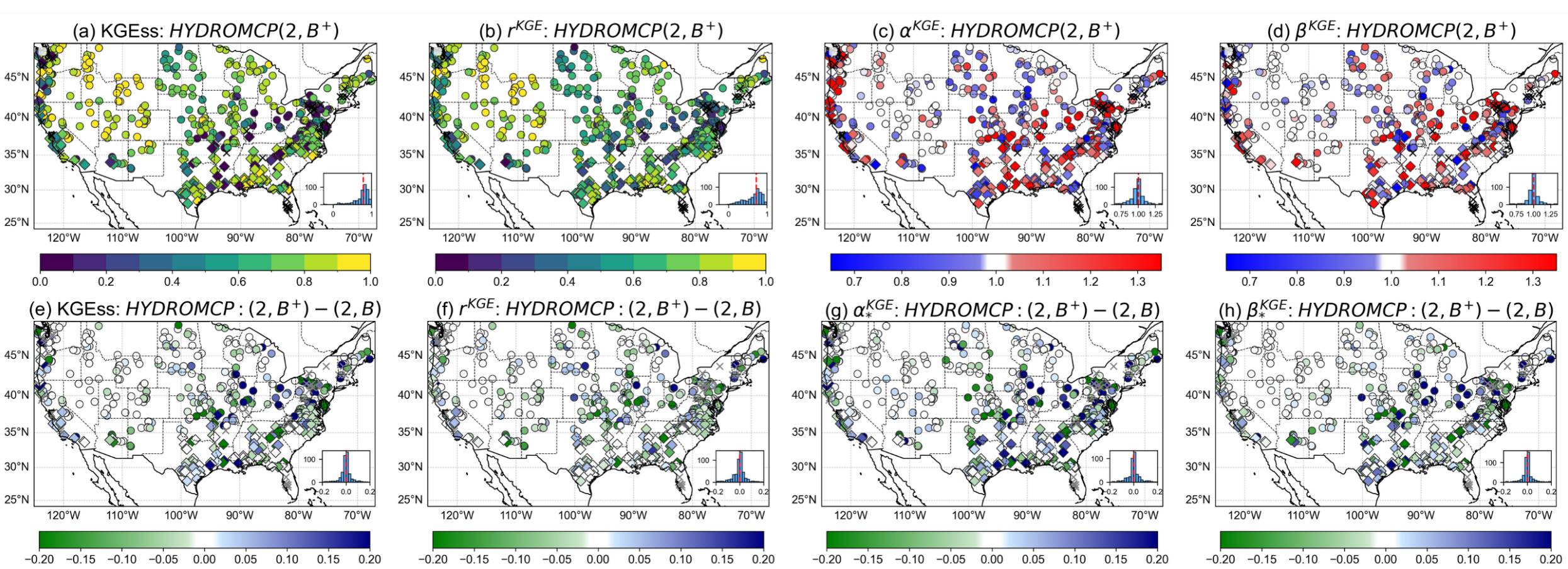

**Figure ST1:** Skill metrics for $HYDROMCP(2,B^{+})$across 513 CONUS basins, including (a) the KGE skill score ($KGE_{ss}$) and the three components of KGE : (b) linear correlation ($\gamma^{KGE}$), (c) flow variability ratio ($\alpha^{KGE}$), and (d) mass balance ratio ($\beta^{KGE}$). Subplots (e–h) show the corresponding differences in these metrics between $HYDROMCP(2,B^{+})$ and $HYDROMCP(2,B)$.

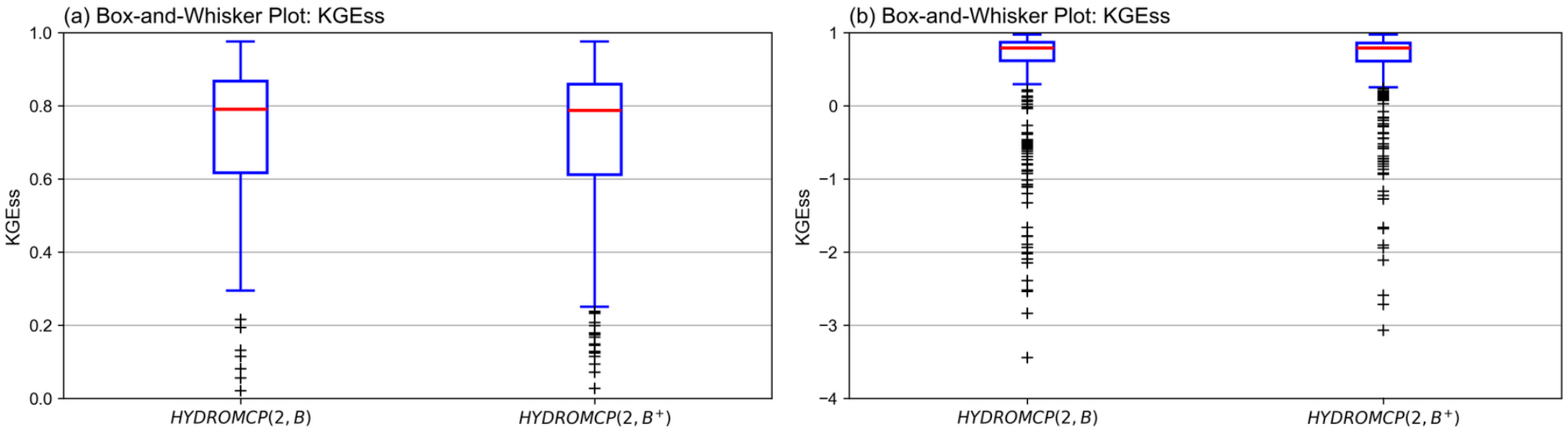

**Figure ST2:** Box–whisker plots of the $KGE_{ss}$ metric for the $HYDROMCP(2,B^{+})$ and $HYDROMCP(2,B)$ architectures across 513 basins for (a) skill ranging on 0 to 1, and (b) skill ranging on -4 to 1

## Text S3. Summary of the AIC selection procedure conducted solely on the SOILMCP, SNOWMCP, and HYDROMCP model families.

We conducted the AIC (and KGE) model selection procedure not only by comparing architectures across all candidates, but also separately within each architecture class, as summarized below for the HMCP, SNOWMCP, and HYDROMCP candidates. Detailed counts of selected models and their associated $KGE_{ss}$ metrics are provided in the supplementary materials (**Tables S6–S9**), to which we refer the reader for further information.

### *Minimum Architecture for HMCP (SOILMCP) Architectures*

Results for the AIC-selected optimal models using the SOIL-MCP architecture are shown in **Figure S17a**. Since all three candidate models have a single cell-state and a single flow-path, we use color coding to distinguish between them: the simplest model, $HMCP(1,B)$, is shown in gray; the intermediate model, $HMCP(1,AL)$ in blue; and the most complex model, $HMCP(1,MR)$, in red. We found that the simplest $HMCP(1,B)$ architecture was among the most frequently selected at AM, RM, CRB, and SN, except at CC, where the $HMCP(1,AL)$ was chosen most often. In contrast, the most complex $HMCP(1,MR)$ architecture was selected more frequently only at RM. Overall, the testing period median $KGE_{ss}$ reaches $0.75$ with values of $0.72$ and $0.81$ for non-snowy and snowy basins, respectively (**Figure S17b**). AIC penalizes models with the larger numbers of parameters, with $HMCP(1,B)$, $HMCP(1,AL)$, and $HMCP(1,MR)$ being selected at 246, 203, and 64 basins respectively.

**Figure S17c** shows architectural selection based solely on $KGE$ performance, without penalizing for model complexity. $HMCP(1,B)$, $HMCP(1,AL)$, and $HMCP(1,MR)$ are selected at $82$, $205$, and $226$ basins respectively. Nearly half of the basins exhibit an increase in model complexity, with 76 shifting from $HMCP(1)$ to $HMCP(1,AL)$, 88 from $HMCP(1)$ to $HMCP(1,MR)$, and 91 from $HMCP(1,AL)$ to $HMCP(1,MR)$. In addition, about $44\%$ of the forest-covered basins (124 out of 282) exhibit a change in optimal architecture.

Accordingly, AIC appears to be an effective approach for identifying parsimonious architectures. For instance, the AIC more often selects for $HMCP(1,B)$ instead of $HMCP(1,MR)$ at mountainous US basins. Although the additional mass-relaxation mechanism can account for snow accumulation and melt — thereby improving accuracy – which is not represented by the SOILMCP architecture, the improvement does not arise for the right reason.

**Figure S18a** shows the box–whisker plot of testing period $KGE_{ss}$ across all 513 basins, comparing all SOILMCP candidate architectures with the AIC- and KGE-selected optimal architectures. The AIC-selected architecture falls between the two extremes, whereas the KGE-selected architecture serves as the upper benchmark. The two approaches yield testing period median $KGE_{ss}$ values of $0.75$ and $0.77$, respectively. Overall, performance of the SOILMCP models suggests that their ability to achieve desirable results is largely confined to the western mountainous regions of the US, the southwestern US, and the northern portion of the eastern US (**Figures S17b & S17d**).

### *Minimum Architecture for SNOWMCP Architectures*

**Figure S17e** presents the AIC-selected optimal architecture for the SNOWMCP-based architecture. In all 502 basins for which SNOW-MCP was successfully trained, the numbers of catchments corresponding to the simplest architecture (coded in gray) through to the most complex architecture (coded in red) are 350, 134, and 18, respectively. The testing period median $KGE_{ss}$ reaches $0.80$ and $0.73$, at snowy and non-snow basins respectively (**Figure S17f**).

We observed a relatively large AIC penalty for the SNOWMCP architecture compared to the SOILMCP when evaluated against the KGE-based selection (**Figure S17g**), with 107 basins shifting from $SNOWMCP(1,BQ)$ to $SNOWMCP(1,GQ)$, 152 from $SNOWMCP(1,BQ)$ to $SNOWMCP(1,CQ)$, and 74 from $SNOWMCP(1,GQ)$ to $SNOWMCP(1,CQ)$. This increase in complexity appears to be uniform across the country. The major skill improvement of the KGE-selected architecture over the AIC-selected optimal architecture occurs in the non-snowy catchments (**Figure S17h**), where testing period median $KGE_{ss}$ increases from $0.73$ to $0.79$. In contrast, the snowy catchments show only a marginal improvement, from $0.80$ to $0.82$.

Overall, while the KGE-selected approach does improve skill, it raises a similar question of whether the improvement is for the right reasons, given that most of the gains occur in non-snowy basins. The performance of the AIC-selected model complexity is comparable to that of any SNOWMCP candidate (**Figure S18b**). Notably, $270$ out of the $360$ snowy basins (75%) point to the selection of the simplest $SNOWMCP(1, GQ)$, with a testing period median $KGE_{ss}$ of $0.77$ (only $39$ basins have testing period median $KGE_{ss} < 0.50$). This suggests that one can be reasonably confident in recommending the use of $SNOWMCP(1, BQ)$ as the basis, before introducing additional model complexity, unless there is a better understanding of the local hydrologic regime.

### *Minimum Architecture for HYDROMCP Architectures*

In all $476$ successfully completed HYDRO-MCP basins, the AIC metric effectively penalized model complexity, with $282$, $134$ and $60$ basins selecting $HYDROMCP(2, B)$, $HYDROMCP(2, G)$, and $HYDROMCP(2, C)$, respectively (**Figure S17i**). Notably, AIC rarely selected the most complex snowpack architecture. The choice of $HYDROMCP(2, C)$ in non-snowy basins across the southeastern US likely reflects a lack of an adequate RR component, where additional parameters reduce the mismatch between simulation and observation. By contrast, KGE-based selection favored more complex models, yielding $121$, $142$ and $213$ basins (**Figure S17k**).

However, poorly performing basins (e.g., those in purple with $KGE_{ss} < 0.1$) persisted under both criteria and were more frequent than in the SOILMCP and SNOWMCP cases for both non-snowy and snowy basins (**Figure S17j** and **S17l**). The boxplot in **Figure S18c** further confirms these unusually low levels of skill. This suggests the possible unsuitability of coupled architectures for streamflow modeling, even in snowy catchments. Overall, the results provide hydrologists with a preliminary understanding of modeling strategies when the site-specific hydrologic regime is unknown, and they help to reformulate or rethink approaches for developing bottom-up modeling tools.

**Table S1.** Summary of parameter initialization, learning rates and number of epochs used for each network configuration

<table>
<tr><th>Model Name</th><th>Number of Epoch</th><th>Learning Rate</th><th>Parameter Initialization/Inheritance</th></tr>
<tr><td>$HMCP(1,B)$</td><td>5000</td><td rowspan="2">0.025 initially; 0.0125 after 500 epochs</td><td>Randomly initialize all parameters from -1 to 1</td></tr>
<tr><td>$HMCP(1,AL)$</td><td>3000</td><td>Inherited parameters from $HMCP(1,B)$, with the extra parameter initialized randomly between −1 and 1</td></tr>
<tr><td>$HMCP(1,MR)$</td><td>3000</td><td>0.025 initially; 0.0125 after 500 epochs; 0.005 after 2000 epochs</td><td>Inherited parameters from $HMCP(1,B)$ with the extra parameter initialized randomly between −1 and 1</td></tr>
<tr><td>$SNOWMCP(1,B)$</td><td>1500</td><td rowspan="3">0.05 initially; 0.025 after 1000 epochs</td><td rowspan="3">Randomly initialized parameters between −1 and 1</td></tr>
<tr><td>$SNOWMCP(1,G)$</td><td>1500</td></tr>
<tr><td>$SNOWMCP(1,C)$</td><td>1500</td></tr>
<tr><td>$SNOWMCP(1,BQ)$</td><td>3000</td><td rowspan="6">0.025 initially; 0.0125 after 500 epochs</td><td>Inherited parameters from $SNOWMCP(1,B)$</td></tr>
<tr><td>$SNOWMCP(1,GQ)$</td><td>3000</td><td>Inherited parameters from $SNOWMCP(1,G)$</td></tr>
<tr><td>$SNOWMCP(1,CQ)$</td><td>3000</td><td>Inherited parameters from $SNOWMCP(1,C)$</td></tr>
<tr><td>$HYDROMCP(2,B)$</td><td>3000</td><td>Inherited parameters from $HMCP(1,B)$and $SNOWMCP(1,B)$</td></tr>
<tr><td>$HYDROMCP(2,G)$</td><td>3000</td><td>Inherited parameters from $HMCP(1,B)$ and $SNOWMCP(1,G)$</td></tr>
<tr><td>$HYDROMCP(2,C)$</td><td>3000</td><td>Inherited parameters from $HMCP(1,B)$ and $SNOWMCP(1,C)$</td></tr>
<tr><td colspan="4">In Appendix</td></tr>
<tr><td>$HMCP(1,B^{con})$</td><td>2500</td><td>0.025 initially; 0.0125 after 500 epochs</td><td>Inherited parameters from $HMCP(1,B)$</td></tr>
<tr><td>$HMCP(1,AL^{con})$</td><td>2500</td><td>0.025 initially; 0.0125 after 500 epochs</td><td>Inherited parameters from $HMCP(1,AL)$</td></tr>
<tr><td colspan="4">In Supplementary Material</td></tr>
<tr><td>$HYDROMCP(2,B^{+})$</td><td>3000</td><td>0.025 initially; 0.0125 after 500 epochs</td><td>Inherited parameters from $HMCP(1,AL)$ and $SNOWMCP(1,B)$</td></tr>
</table>

**Table S2.** Summary of $r^{KGE}$ metric for the MCP-based architectures across geographic regions in the CONUS

| Region | Percentile | Architecture Backbone | | | | | | | | | | |
|---|---|---|---|---|---|---|---|---|---|---|---|---|
| | | SOILMCP (HMCP) | | | SNOWMCP | | | HYDROMCP | | | | |
| | | $(1)$ | $(1, AL)$ | $(1, MR)$ | $(1, BQ)$ | $(1, GQ)$ | $(1, CQ)$ | $(1, B)$ | $(1, S)$ | $(1, P)$ | $(1, G)$ | $(1, C)$ |
| | 5 | 0.28 | 0.28 | 0.33 | 0.09 | 0.23 | 0.24 | 0.15 | 0.15 | 0.29 | 0.28 | 0.29 |
| | 25 | 0.49 | 0.51 | 0.56 | 0.57 | 0.61 | 0.62 | 0.35 | 0.35 | 0.55 | 0.55 | 0.55 |
| CONUS | 50 | 0.64 | 0.67 | 0.68 | 0.71 | 0.74 | 0.73 | 0.56 | 0.54 | 0.72 | 0.74 | 0.74 |
| | 75 | 0.76 | 0.77 | 0.77 | 0.80 | 0.81 | 0.81 | 0.70 | 0.70 | 0.82 | 0.83 | 0.83 |
| | 95 | 0.86 | 0.87 | 0.87 | 0.89 | 0.90 | 0.90 | 0.81 | 0.82 | 0.92 | 0.93 | 0.93 |
| Eastern CONUS | | 0.67 | 0.70 | 0.69 | 0.68 | 0.69 | 0.71 | 0.59 | 0.58 | 0.70 | 0.71 | 0.71 |
| Western CONUS | | 0.53 | 0.56 | 0.63 | 0.80 | 0.81 | 0.82 | 0.51 | 0.51 | 0.80 | 0.80 | 0.82 |
| AM | | 0.64 | 0.67 | 0.66 | 0.70 | 0.70 | 0.73 | 0.55 | 0.56 | 0.69 | 0.70 | 0.69 |
| RM | | 0.46 | 0.46 | 0.61 | 0.85 | 0.85 | 0.86 | 0.40 | 0.44 | 0.90 | 0.90 | 0.91 |
| CRB | | 0.45 | 0.45 | 0.55 | 0.76 | 0.78 | 0.79 | 0.32 | 0.32 | 0.80 | 0.84 | 0.86 |
| SN | | 0.48 | 0.49 | 0.48 | 0.81 | 0.83 | 0.85 | 0.57 | 0.55 | 0.87 | 0.89 | 0.86 |
| CC | | 0.75 | 0.75 | 0.76 | 0.84 | 0.84 | 0.82 | 0.65 | 0.64 | 0.82 | 0.82 | 0.83 |
| Snowy | 50 | 0.60 | 0.62 | 0.65 | 0.73 | 0.75 | 0.75 | 0.53 | 0.51 | 0.74 | 0.75 | 0.76 |
| Non-Snowy | | 0.74 | 0.76 | 0.76 | 0.57 | 0.67 | 0.69 | 0.61 | 0.61 | 0.66 | 0.69 | 0.68 |
| Forest | | 0.67 | 0.70 | 0.69 | 0.72 | 0.76 | 0.76 | 0.60 | 0.60 | 0.76 | 0.76 | 0.77 |
| Open | | 0.60 | 0.62 | 0.65 | 0.69 | 0.69 | 0.69 | 0.50 | 0.47 | 0.70 | 0.71 | 0.70 |
| Arid Climate | | 0.46 | 0.46 | 0.50 | 0.39 | 0.61 | 0.59 | 0.32 | 0.33 | 0.71 | 0.72 | 0.74 |
| Temperate Climate | | 0.74 | 0.76 | 0.75 | 0.70 | 0.75 | 0.74 | 0.64 | 0.63 | 0.70 | 0.71 | 0.71 |
| Cold Climate | | 0.57 | 0.59 | 0.62 | 0.72 | 0.74 | 0.74 | 0.48 | 0.47 | 0.76 | 0.75 | 0.77 |
| Polar Climate | Avg | 0.44 | 0.44 | 0.75 | 0.91 | 0.91 | 0.92 | 0.74 | 0.35 | 0.56 | 0.94 | 0.94 |

**Table S3.** Summary of $\alpha_*^{KGE}$ metric for the MCP-based architectures across geographic regions in the CONUS

| Region | Percentile | Architecture Backbone | | | | | | | | | | |
|---|---|---|---|---|---|---|---|---|---|---|---|---|
| | | SOILMCP (HMCP) | | | SNOWMCP | | | HYDROMCP | | | | |
| | | $(1)$ | $(1, AL)$ | $(1, MR)$ | $(1, BQ)$ | $(1, GQ)$ | $(1, CQ)$ | $(1, B)$ | $(1, S)$ | $(1, P)$ | $(1, G)$ | $(1, C)$ |
| CONUS | 5 | 0.59 | 0.60 | 0.70 | 0.64 | 0.70 | 0.72 | -3.47 | -3.39 | -1.91 | -2.15 | -1.78 |
| | 25 | 0.85 | 0.84 | 0.88 | 0.88 | 0.88 | 0.89 | 0.50 | 0.31 | 0.83 | 0.84 | 0.85 |
| | 50 | 0.93 | 0.92 | 0.93 | 0.95 | 0.95 | 0.95 | 0.88 | 0.87 | 0.94 | 0.94 | 0.95 |
| | 75 | 0.97 | 0.96 | 0.97 | 0.98 | 0.98 | 0.98 | 0.95 | 0.95 | 0.98 | 0.98 | 0.98 |
| | 95 | 0.99 | 0.99 | 0.99 | 1.00 | 1.00 | 1.00 | 0.99 | 0.99 | 1.00 | 1.00 | 0.99 |
| Eastern CONUS | 50 | 0.93 | 0.93 | 0.93 | 0.94 | 0.93 | 0.93 | 0.89 | 0.89 | 0.92 | 0.92 | 0.93 |
| Western CONUS | | 0.93 | 0.90 | 0.95 | 0.97 | 0.97 | 0.97 | 0.83 | 0.79 | 0.97 | 0.97 | 0.97 |
| AM | | 0.92 | 0.93 | 0.93 | 0.96 | 0.95 | 0.94 | 0.88 | 0.88 | 0.91 | 0.91 | 0.91 |
| RM | | 0.79 | 0.81 | 0.96 | 0.99 | 0.98 | 0.98 | 0.81 | 0.70 | 0.98 | 0.98 | 0.98 |
| CRB | | 0.81 | 0.79 | 0.94 | 0.98 | 0.97 | 0.98 | 0.52 | 0.35 | 0.97 | 0.97 | 0.96 |
| SN | | 0.80 | 0.81 | 0.83 | 0.98 | 0.97 | 0.96 | 0.84 | 0.83 | 0.98 | 0.98 | 0.98 |
| CC | | 0.94 | 0.94 | 0.95 | 0.96 | 0.96 | 0.97 | 0.92 | 0.92 | 0.96 | 0.97 | 0.97 |
| Snowy | | 0.92 | 0.92 | 0.94 | 0.96 | 0.96 | 0.96 | 0.87 | 0.85 | 0.96 | 0.95 | 0.96 |
| Non-Snowy | | 0.94 | 0.93 | 0.93 | 0.92 | 0.91 | 0.92 | 0.88 | 0.90 | 0.91 | 0.91 | 0.92 |
| Forest | | 0.93 | 0.93 | 0.94 | 0.96 | 0.96 | 0.95 | 0.90 | 0.90 | 0.94 | 0.94 | 0.95 |
| Open | | 0.91 | 0.90 | 0.93 | 0.93 | 0.94 | 0.93 | 0.85 | 0.83 | 0.94 | 0.95 | 0.95 |
| Arid Climate | | 0.92 | 0.89 | 0.91 | 0.92 | 0.93 | 0.92 | 0.77 | 0.75 | 0.95 | 0.94 | 0.94 |
| Temperate Climate | | 0.94 | 0.93 | 0.93 | 0.94 | 0.93 | 0.93 | 0.89 | 0.90 | 0.92 | 0.91 | 0.93 |
| Cold Climate | | 0.91 | 0.91 | 0.94 | 0.96 | 0.96 | 0.96 | 0.87 | 0.85 | 0.96 | 0.96 | 0.96 |
| Polar Climate | Avg | 0.47 | 0.47 | 0.98 | 0.99 | 0.99 | 0.98 | 0.26 | 0.29 | 0.84 | 0.99 | 0.99 |

**Table S4.** Summary of $\beta_*^{KGE}$ metric for the MCP-based architectures across geographic regions in the CONUS

| Region | Percentile | Architecture Backbone | | | | | | | | | | |
|---|---|---|---|---|---|---|---|---|---|---|---|---|
| | | SOILMCP (HMCP) | | | SNOWMCP | | | HYDROMCP | | | | |
| | | $(1)$ | $(1, AL)$ | $(1, MR)$ | $(1, BQ)$ | $(1, GQ)$ | $(1, CQ)$ | $(1, B)$ | $(1, S)$ | $(1, P)$ | $(1, G)$ | $(1, C)$ |
| CONUS | 5 | 0.79 | 0.80 | 0.87 | 0.79 | 0.80 | 0.82 | -1.04 | -1.27 | -0.60 | -0.51 | -0.60 |
| | 25 | 0.92 | 0.93 | 0.95 | 0.93 | 0.94 | 0.94 | 0.54 | 0.40 | 0.91 | 0.91 | 0.92 |
| | 50 | 0.96 | 0.96 | 0.97 | 0.97 | 0.97 | 0.97 | 0.92 | 0.91 | 0.97 | 0.97 | 0.97 |
| | 75 | 0.99 | 0.99 | 0.99 | 0.99 | 0.99 | 0.99 | 0.97 | 0.97 | 0.99 | 0.99 | 0.99 |
| | 95 | 1.00 | 1.00 | 1.00 | 1.00 | 1.00 | 1.00 | 0.99 | 0.99 | 1.00 | 1.00 | 1.00 |
| Eastern CONUS | 50 | 0.97 | 0.97 | 0.97 | 0.97 | 0.97 | 0.97 | 0.94 | 0.93 | 0.96 | 0.96 | 0.96 |
| Western CONUS | | 0.95 | 0.95 | 0.98 | 0.98 | 0.98 | 0.98 | 0.88 | 0.84 | 0.98 | 0.98 | 0.98 |
| AM | | 0.97 | 0.97 | 0.97 | 0.98 | 0.98 | 0.98 | 0.93 | 0.93 | 0.96 | 0.97 | 0.96 |
| RM | | 0.92 | 0.91 | 0.97 | 0.99 | 0.99 | 0.99 | 0.83 | 0.74 | 0.99 | 0.99 | 0.99 |
| CRB | | 0.92 | 0.92 | 0.97 | 0.97 | 0.97 | 0.98 | 0.83 | 0.59 | 0.98 | 0.98 | 0.99 |
| SN | | 0.96 | 0.96 | 0.97 | 0.97 | 0.98 | 0.99 | 0.77 | 0.67 | 0.99 | 0.99 | 0.98 |
| CC | | 0.96 | 0.96 | 0.99 | 0.98 | 0.98 | 0.98 | 0.87 | 0.85 | 0.98 | 0.98 | 0.99 |
| Snowy | | 0.96 | 0.96 | 0.97 | 0.98 | 0.98 | 0.98 | 0.92 | 0.90 | 0.97 | 0.98 | 0.98 |
| Non-Snowy | | 0.97 | 0.96 | 0.96 | 0.94 | 0.96 | 0.95 | 0.93 | 0.93 | 0.95 | 0.95 | 0.95 |
| Forest | | 0.97 | 0.97 | 0.98 | 0.98 | 0.98 | 0.98 | 0.93 | 0.92 | 0.97 | 0.98 | 0.97 |
| Open | | 0.95 | 0.96 | 0.96 | 0.96 | 0.97 | 0.97 | 0.91 | 0.90 | 0.96 | 0.97 | 0.97 |
| Arid Climate | | 0.94 | 0.93 | 0.94 | 0.95 | 0.94 | 0.96 | 0.89 | 0.85 | 0.97 | 0.96 | 0.97 |
| Temperate Climate | | 0.97 | 0.97 | 0.97 | 0.96 | 0.97 | 0.97 | 0.94 | 0.94 | 0.96 | 0.96 | 0.97 |
| Cold Climate | | 0.96 | 0.96 | 0.97 | 0.98 | 0.98 | 0.98 | 0.91 | 0.87 | 0.97 | 0.98 | 0.97 |
| Polar Climate | Avg | 0.79 | 0.78 | 0.99 | 0.99 | 0.98 | 0.98 | 0.97 | 0.51 | 0.87 | 0.99 | 0.99 |

**Table S5.** Summary of streamflow and SWE performance metrics for the two selected basins, focusing on the additional $SNOWMCP(1, BQ)^{*}$ case and comparing it with other SNOWMCP and HYDROMCP cases.

| Metric | Flow Regime | Streamflow | | | | Snow Water Equivalent | | | |
|---|---|---|---|---|---|---|---|---|---|
| | | $SNOWMCP$ $(1, B)$ | $SNOWMCP$ $(1, BQ)$ | $SNOWMCP$ $(1, BQ)^{*}$ | $HYDROMCP$ $(2, B)$ | $SNOWMCP$ $(1, B)$ | $SNOWMCP$ $(1, BQ)$ | $SNOWMCP$ $(1, BQ)^{*}$ | $HYDROMCP$ $(2, B)$ |
| $KGE_{ss}$ | Dry (WY1992) | -0.51 | 0.54 | 0.52 | 0.49 | 0.93 | 0.08 | 0.87 | 0.87 |
| | Median (WY2000) | -0.02 | 0.88 | 0.84 | 0.83 | 0.86 | 0.01 | 0.86 | 0.84 |
| | Wet (WY1997) | 0.35 | 0.87 | 0.89 | 0.90 | 0.88 | 0.48 | 0.94 | 0.90 |
| $RMSE$ | Dry (WY1992) | 1.46 | 0.48 | 0.60 | 0.42 | 15.17 | 140.59 | 40.70 | 19.68 |
| | Median (WY2000) | 2.27 | 0.46 | 0.60 | 0.48 | 35.16 | 211.12 | 30.55 | 36.65 |
| | Wet (WY1997) | 3.64 | 0.88 | 0.91 | 0.68 | 53.08 | 275.63 | 44.96 | 54.99 |
| $MAE$ | Dry (WY1992) | 0.85 | 0.32 | 0.43 | 0.27 | 11.07 | 123.56 | 29.95 | 15.26 |
| | Median (WY2000) | 1.32 | 0.36 | 0.43 | 0.33 | 20.11 | 167.60 | 25.02 | 21.67 |
| | Wet (WY1997) | 2.05 | 0.63 | 0.61 | 0.45 | 41.65 | 207.49 | 32.73 | 40.86 |

**Table S6.** Summary of the number of basins selected by the AIC criterion in individual experiments with SOILMCP, SNOWMCP, and HYDROMCP.

| Regions | Architecture Backbone | | | | | | | | |
|---|---|---|---|---|---|---|---|---|---|
| | Experiment 1 | | | Experiment 2 | | | Experiment 3 | | |
| | SOILMCP | | | SNOWMCP | | | HYDROMCP | | |
| | $(1)$ | $(1, AL)$ | $(1, MR)$ | $(1, BQ)$ | $(1, GQ)$ | $(1, CQ)$ | $(1, B)$ | $(1, G)$ | $(1, C)$ |
| CONUS | 233 | 202 | 78 | 340 | 142 | 20 | 286 | 124 | 66 |
| Eastern CONUS | 157 | 130 | 39 | 208 | 95 | 15 | 190 | 74 | 33 |
| Western CONUS | 76 | 72 | 39 | 132 | 47 | 5 | 96 | 50 | 33 |
| AM | 49 | 47 | 7 | 83 | 19 | 1 | 52 | 25 | 10 |
| RM | 27 | 8 | 19 | 36 | 16 | 2 | 29 | 13 | 12 |
| CRB | 17 | 10 | 7 | 19 | 11 | 4 | 21 | 9 | 4 |
| SN | 4 | 4 | 2 | 6 | 4 | 0 | 4 | 5 | 1 |
| CC | 10 | 21 | 2 | 29 | 4 | 0 | 14 | 10 | 7 |
| Snowy | 163 | 144 | 53 | 259 | 91 | 10 | 195 | 97 | 48 |
| Non-Snowy | 70 | 58 | 25 | 81 | 51 | 10 | 91 | 27 | 18 |
| Forest | 121 | 120 | 41 | 197 | 70 | 11 | 148 | 74 | 33 |
| Open | 112 | 82 | 37 | 143 | 72 | 9 | 138 | 50 | 33 |
| Arid Climate | 15 | 13 | 7 | 16 | 18 | 1 | 15 | 9 | 11 |
| Temperate Climate | 100 | 106 | 36 | 156 | 64 | 11 | 142 | 54 | 24 |
| Cold Climate | 116 | 83 | 35 | 166 | 60 | 8 | 128 | 60 | 31 |
| Polar Climate | 2 | 0 | 0 | 2 | 0 | 0 | 1 | 1 | 0 |

**Table S7.** Summary of $KGE_{ss}$ selected by the AIC criterion in individual experiments with SOILMCP, SNOWMCP, and HYDROMCP.

| | Architecture Backbone | | | | | | | | |
|---|---|---|---|---|---|---|---|---|---|
| | Experiment 1 | | | Experiment 2 | | | Experiment 3 | | |
| | SOILMCP | | | SNOWMCP | | | HYDROMCP | | |
| Regions | $(1)$ | $(1, AL)$ | $(1, MR)$ | $(1, BQ)$ | $(1, GQ)$ | $(1, CQ)$ | $(1, B)$ | $(1, G)$ | $(1, C)$ |
| CONUS | 0.73 | 0.76 | 0.74 | 0.80 | 0.78 | 0.77 | 0.70 | 0.72 | 0.80 |
| Eastern CONUS | 0.76 | 0.77 | 0.77 | 0.77 | 0.76 | 0.77 | 0.69 | 0.70 | 0.80 |
| Western CONUS | 0.60 | 0.73 | 0.71 | 0.86 | 0.84 | 0.85 | 0.75 | 0.77 | 0.85 |
| AM | 0.74 | 0.75 | 0.67 | 0.79 | 0.79 | 0.58 | 0.68 | 0.67 | 0.63 |
| RM | 0.53 | 0.64 | 0.72 | 0.90 | 0.88 | 0.89 | 0.89 | 0.85 | 0.91 |
| CRB | 0.55 | 0.66 | 0.60 | 0.85 | 0.80 | 0.86 | 0.63 | 0.85 | 0.87 |
| SN | 0.51 | 0.62 | 0.80 | 0.88 | 0.86 | NaN | 0.82 | 0.81 | 0.95 |
| CC | 0.84 | 0.79 | 0.81 | 0.88 | 0.89 | NaN | 0.64 | 0.83 | 0.79 |
| Snowy | 0.70 | 0.74 | 0.72 | 0.80 | 0.79 | 0.80 | 0.71 | 0.74 | 0.80 |
| Non-Snowy | 0.82 | 0.82 | 0.78 | 0.73 | 0.71 | 0.74 | 0.68 | 0.67 | 0.82 |
| Forest | 0.75 | 0.78 | 0.75 | 0.81 | 0.80 | 0.78 | 0.75 | 0.74 | 0.81 |
| Open | 0.70 | 0.72 | 0.74 | 0.78 | 0.74 | 0.66 | 0.64 | 0.69 | 0.80 |
| Arid Climate | 0.61 | 0.67 | 0.56 | 0.63 | 0.70 | 0.26 | 0.61 | 0.69 | 0.88 |
| Temperate Climate | 0.82 | 0.81 | 0.79 | 0.80 | 0.78 | 0.74 | 0.70 | 0.72 | 0.81 |
| Cold Climate | 0.69 | 0.72 | 0.72 | 0.80 | 0.79 | 0.81 | 0.74 | 0.75 | 0.80 |
| Polar Climate | 0.42 | NaN | NaN | 0.93 | NaN | NaN | 0.01 | 0.94 | NaN |

**Table S8.** Summary of the number of basins selected by the KGE criterion in individual experiments with SOILMCP, SNOWMCP, and HYDROMCP.

| | Architecture Backbone | | | | | | | | |
|---|---|---|---|---|---|---|---|---|---|
| | Experiment 1 | | | Experiment 2 | | | Experiment 3 | | |
| | SOILMCP | | | SNOWMCP | | | HYDROMCP | | |
| Regions | $(1)$ | $(1, AL)$ | $(1, MR)$ | $(1, BQ)$ | $(1, GQ)$ | $(1, CQ)$ | $(1, B)$ | $(1, G)$ | $(1, C)$ |
| CONUS | 82 | 205 | 226 | 91 | 167 | 244 | 121 | 142 | 213 |
| Eastern CONUS | 63 | 157 | 106 | 65 | 112 | 141 | 79 | 98 | 120 |
| Western CONUS | 19 | 48 | 120 | 26 | 55 | 103 | 42 | 44 | 93 |
| AM | 15 | 58 | 30 | 20 | 36 | 47 | 23 | 29 | 35 |
| RM | 0 | 6 | 48 | 5 | 13 | 36 | 9 | 13 | 32 |
| CRB | 2 | 7 | 25 | 3 | 8 | 23 | 12 | 5 | 17 |
| SN | 2 | 5 | 3 | 1 | 3 | 6 | 0 | 4 | 6 |
| CC | 5 | 8 | 20 | 7 | 8 | 18 | 6 | 6 | 19 |
| Snowy | 46 | 146 | 168 | 71 | 114 | 175 | 77 | 95 | 168 |
| Non-Snowy | 36 | 59 | 58 | 20 | 53 | 69 | 44 | 47 | 45 |
| Forest | 39 | 115 | 128 | 36 | 89 | 153 | 59 | 80 | 116 |
| Open | 43 | 90 | 98 | 55 | 78 | 91 | 62 | 62 | 97 |
| Arid Climate | 4 | 10 | 21 | 4 | 12 | 19 | 9 | 9 | 17 |
| Temperate Climate | 52 | 102 | 88 | 41 | 84 | 106 | 64 | 73 | 83 |
| Cold Climate | 26 | 93 | 115 | 45 | 71 | 118 | 48 | 60 | 111 |
| Polar Climate | 0 | 0 | 2 | 1 | 0 | 1 | 0 | 0 | 2 |

**Table S9.** Summary of $KGE_{ss}$ selected by the KGE criterion in individual experiments with SOILMCP, SNOWMCP, and HYDROMCP.

| | Architecture Backbone | | | | | | | | |
|---|---|---|---|---|---|---|---|---|---|
| | Experiment 1 | | | Experiment 2 | | | Experiment 3 | | |
| | SOILMCP | | | SNOWMCP | | | HYDROMCP | | |
| Regions | $(1)$ | $(1, AL)$ | $(1, MR)$ | $(1, BQ)$ | $(1, GQ)$ | $(1, CQ)$ | $(1, B)$ | $(1, G)$ | $(1, C)$ |
| CONUS | 0.81 | 0.79 | 0.74 | 0.81 | 0.82 | 0.82 | 0.74 | 0.81 | 0.85 |
| Eastern CONUS | 0.81 | 0.79 | 0.76 | 0.80 | 0.81 | 0.80 | 0.74 | 0.80 | 0.82 |
| Western CONUS | 0.83 | 0.76 | 0.73 | 0.87 | 0.87 | 0.87 | 0.73 | 0.86 | 0.90 |
| AM | 0.70 | 0.77 | 0.74 | 0.78 | 0.81 | 0.81 | 0.80 | 0.81 | 0.74 |
| RM | NaN | 0.70 | 0.72 | 0.94 | 0.90 | 0.90 | 0.91 | 0.95 | 0.94 |
| CRB | 0.56 | 0.69 | 0.69 | 0.80 | 0.88 | 0.85 | 0.81 | 0.75 | 0.92 |
| SN | 0.56 | 0.61 | 0.75 | 0.79 | 0.87 | 0.92 | NaN | 0.83 | 0.93 |
| CC | 0.86 | 0.84 | 0.81 | 0.88 | 0.89 | 0.87 | 0.63 | 0.88 | 0.88 |
| Snowy | 0.78 | 0.77 | 0.73 | 0.80 | 0.82 | 0.83 | 0.77 | 0.81 | 0.85 |
| Non-Snowy | 0.81 | 0.82 | 0.81 | 0.82 | 0.81 | 0.78 | 0.71 | 0.79 | 0.82 |
| Forest | 0.81 | 0.79 | 0.77 | 0.84 | 0.84 | 0.82 | 0.74 | 0.85 | 0.85 |
| Open | 0.81 | 0.78 | 0.72 | 0.80 | 0.78 | 0.81 | 0.73 | 0.78 | 0.83 |
| Arid Climate | 0.57 | 0.67 | 0.60 | 0.79 | 0.69 | 0.79 | 0.66 | 0.77 | 0.88 |
| Temperate Climate | 0.83 | 0.82 | 0.81 | 0.83 | 0.83 | 0.80 | 0.73 | 0.80 | 0.83 |
| Cold Climate | 0.70 | 0.76 | 0.72 | 0.77 | 0.82 | 0.82 | 0.79 | 0.83 | 0.85 |
| Polar Climate | NaN | NaN | 0.82 | 0.94 | NaN | 0.93 | NaN | NaN | 0.96 |

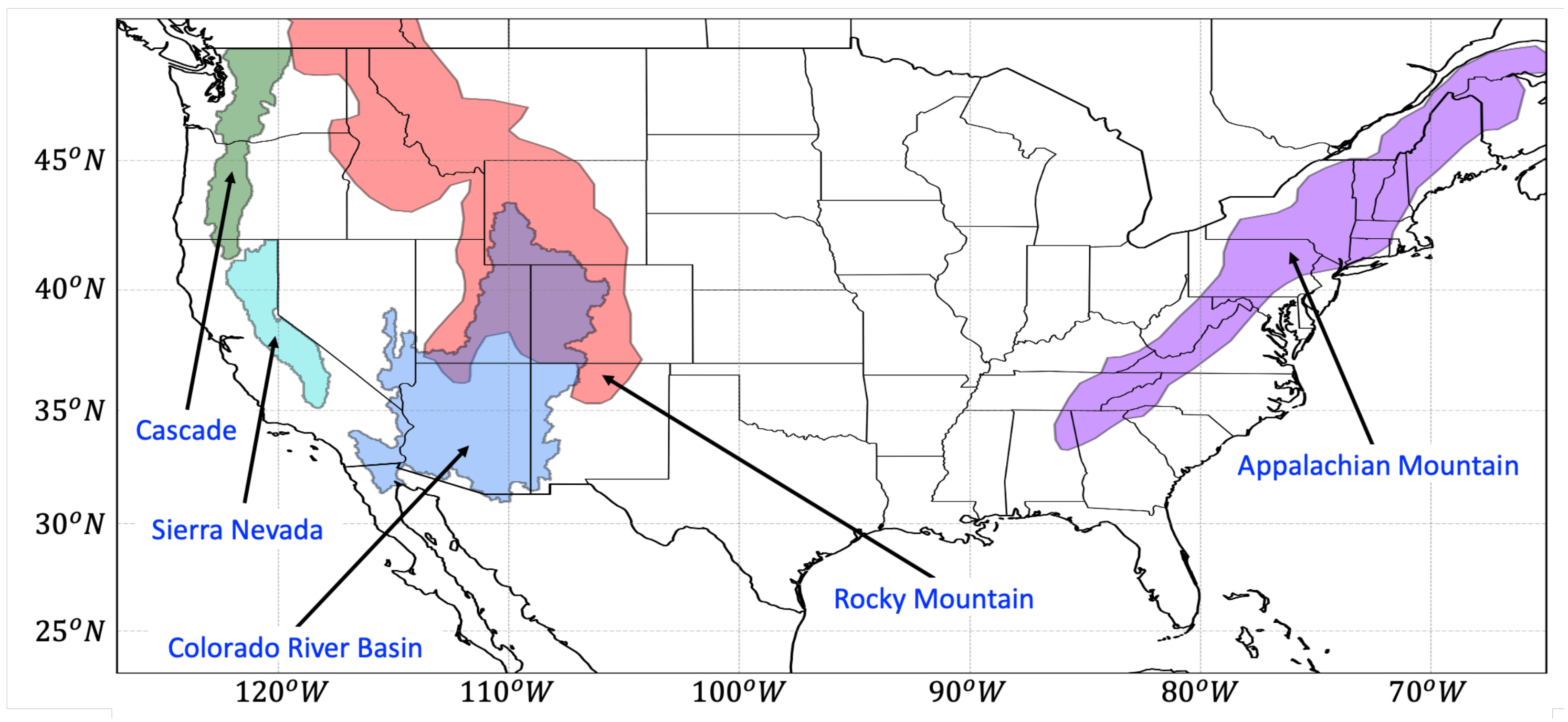

**Figure S1:** Geographic regions highlighted for discussion in this study, including the Appalachian Mountains (AM), Rocky Mountains (RM), Colorado River Basin (CRB), Sierra Nevada (SN), and the Cascades (CC).

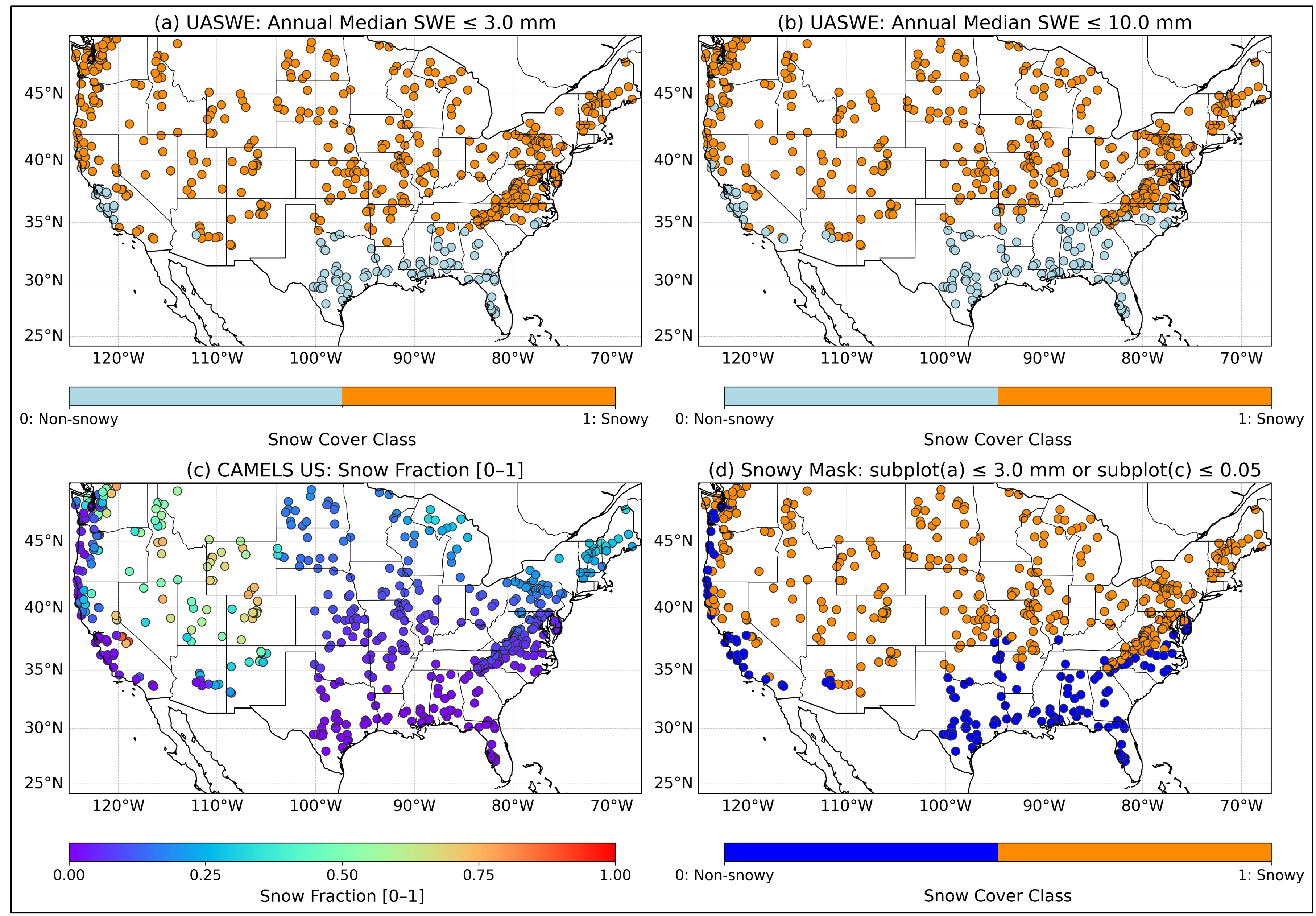

**Figure S2:** Information used to derive the Snowy Mask: (a) University of Arizona (UA) snow product (snow water equivalent; SWE; Broxton et al., 2019)–based mask derived using a 3 mm annual maximum SWE threshold, (b) the 3 mm annual maximum SWE threshold itself, (c) CAMELS-US snow fraction, and (d) the final combined criterion, where non-snowy areas are defined as having an annual maximum SWE ≤ 3 mm and a CAMELS-US snow fraction < 5%.

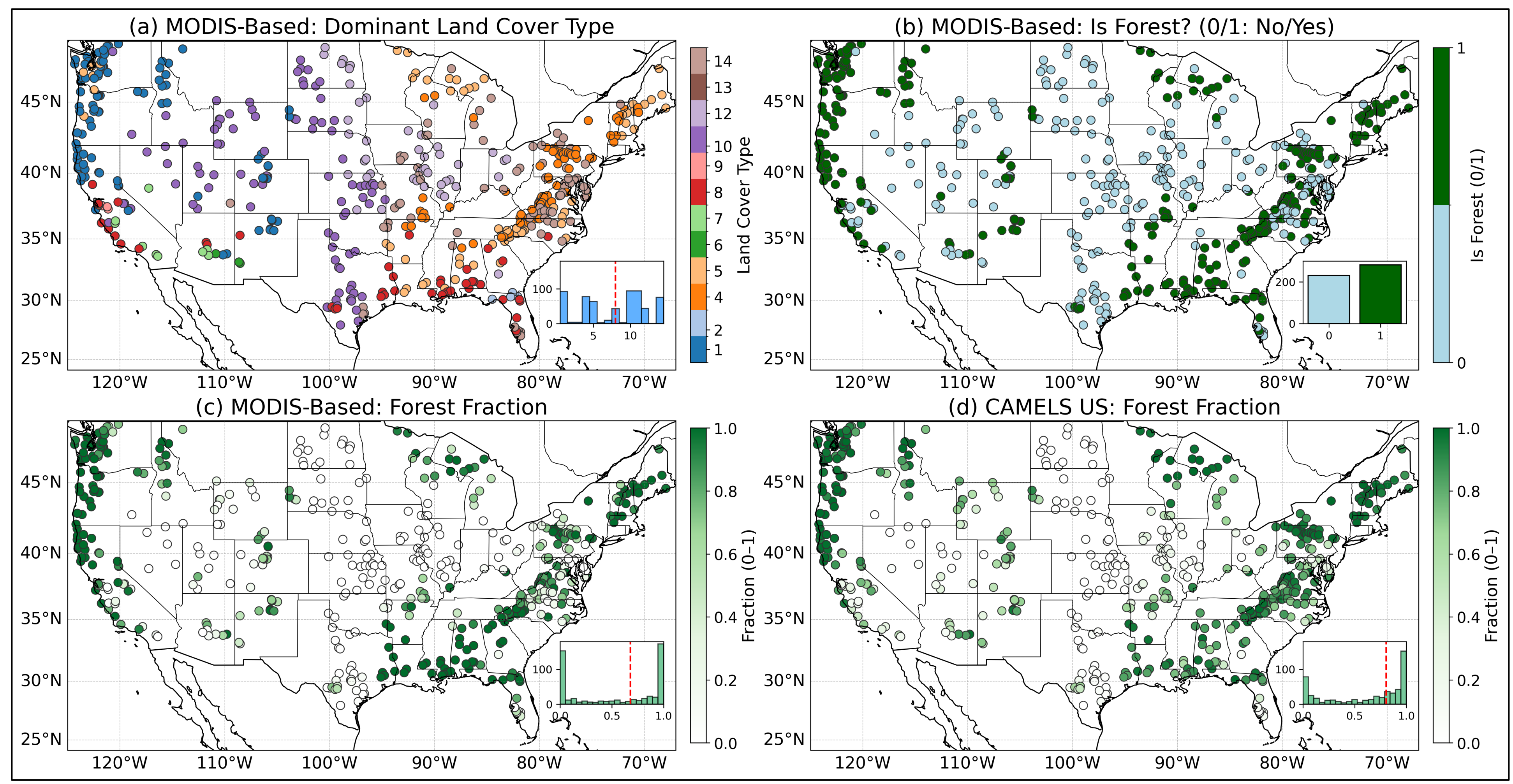

**Figure S3:** Information used to derive the Forest-Covered Mask: (a) MODIS-based land cover classes (Broxton et al., 2014), (b) majority-upscaled binary forest cover derived from (a), (c) forest fraction derived from (a), and (d) CAMELS-US forest fraction.

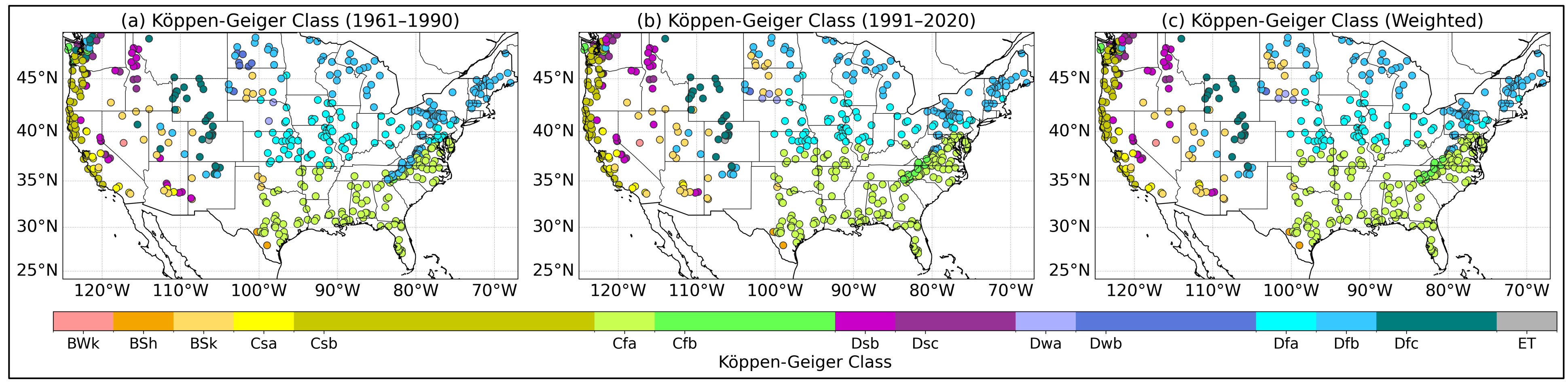

**Figure S4:** Information used to derive the Köppen–Geiger Climatologic Mask (Beck et al., 2023): (a) classification for 1961–1990, (b) classification for 1991–2020, and (c) weighted classification derived from (a) and (b) using linear proportional weighting, consistent with the simulation period (WY 1982–2008).

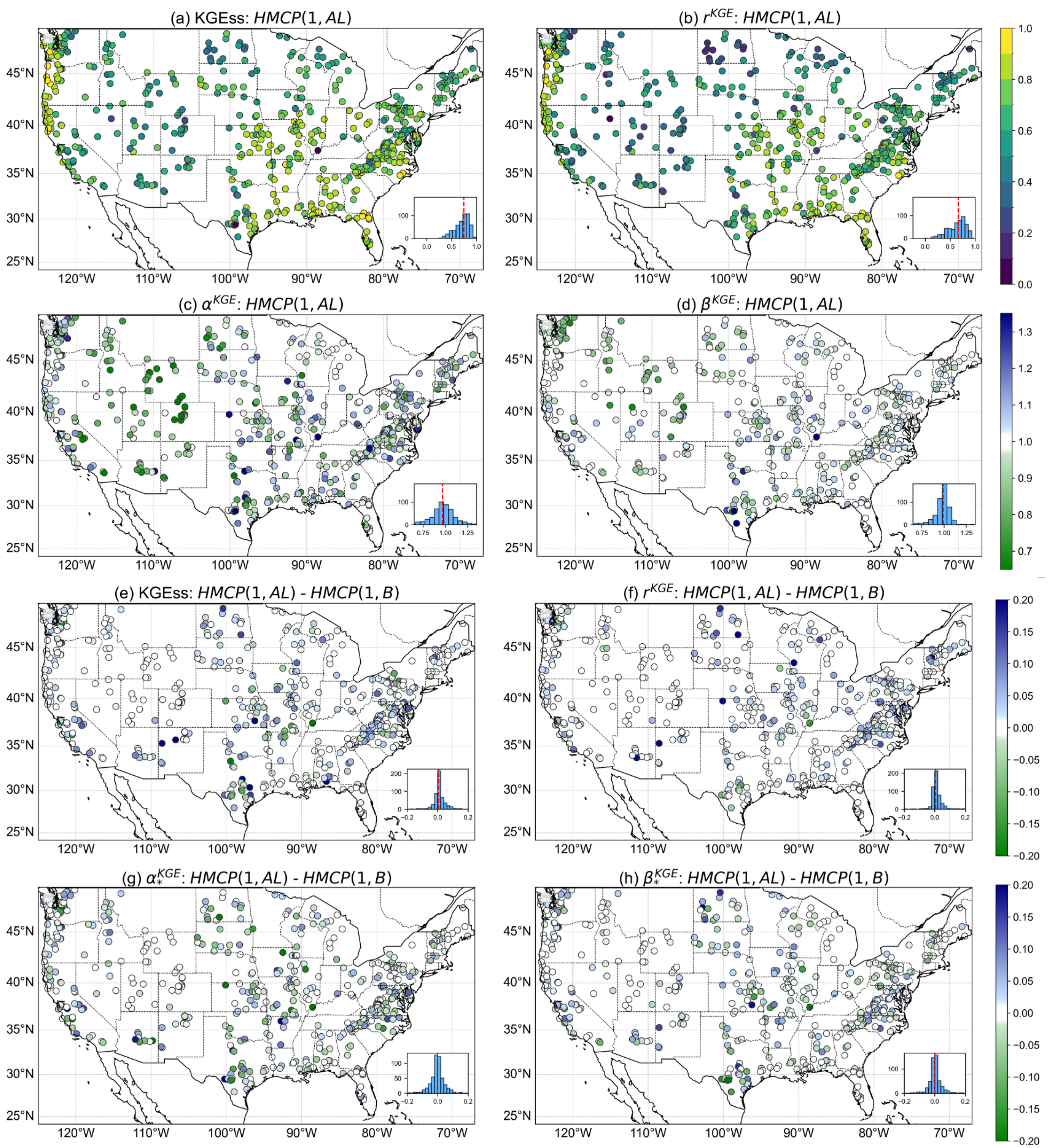

**Figure S5:** Skill metrics for HMCP(1,AL) across 513 CONUS catchments, including (a) the KGE skill score ($KGE_{ss}$) and the three components of KGE: (b) linear correlation ($\gamma^{KGE}$), (c) flow variability ratio ($\alpha^{KGE}$), and (d) mass balance ratio ($\beta^{KGE}$). Subplots (e–h) show the corresponding differences in these metrics between HMCP(1,AL) and HMCP(1,B).

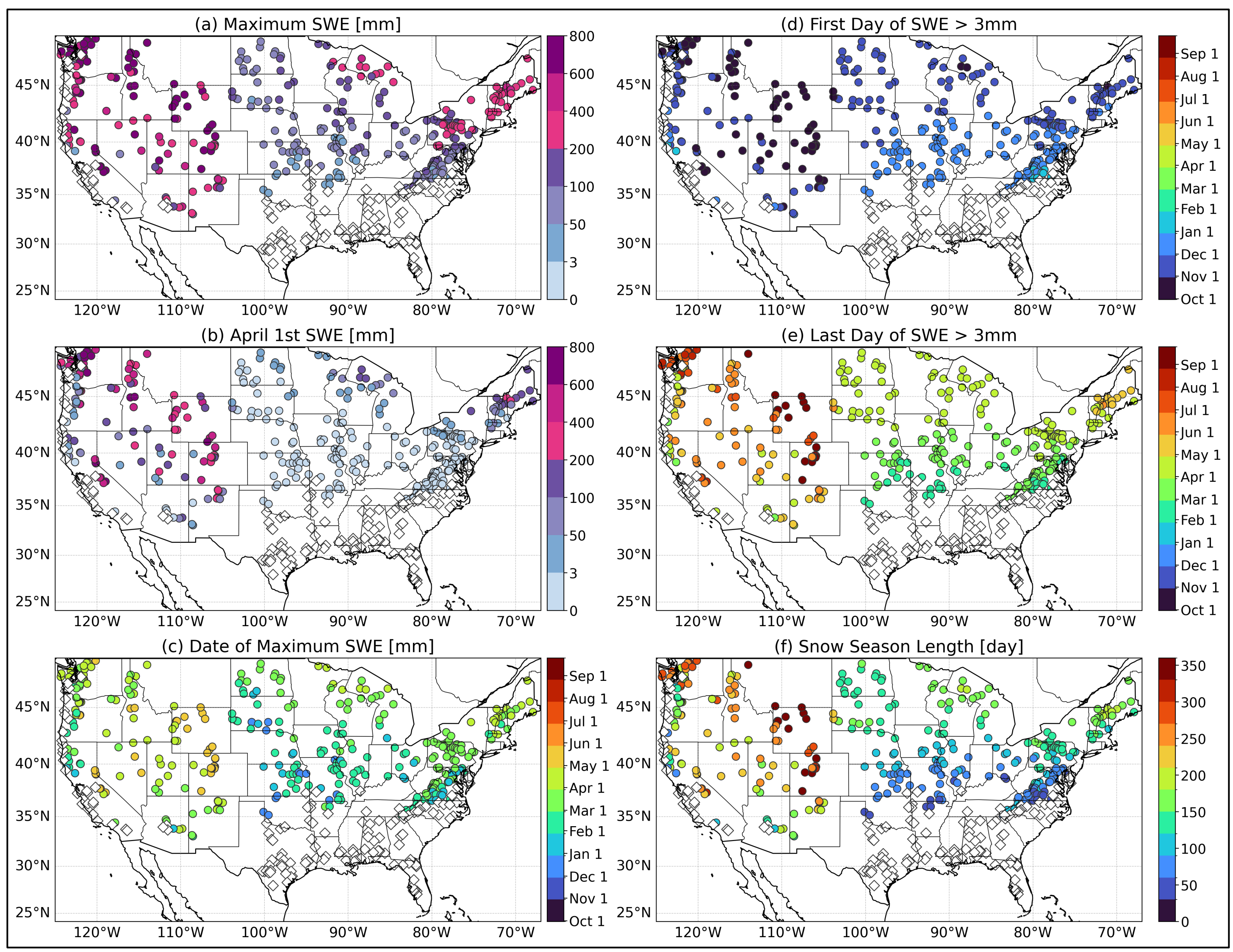

**Figure S6:** UA SWE–derived snow signatures for the study period (WY 1982–2008), including the annual median of (a) maximum SWE (mm), (b) April 1 SWE, (c) date of maximum SWE occurrence, (d) first day of SWE, (e) last day of SWE, and (f) snowy season length (difference between first and last day of SWE).

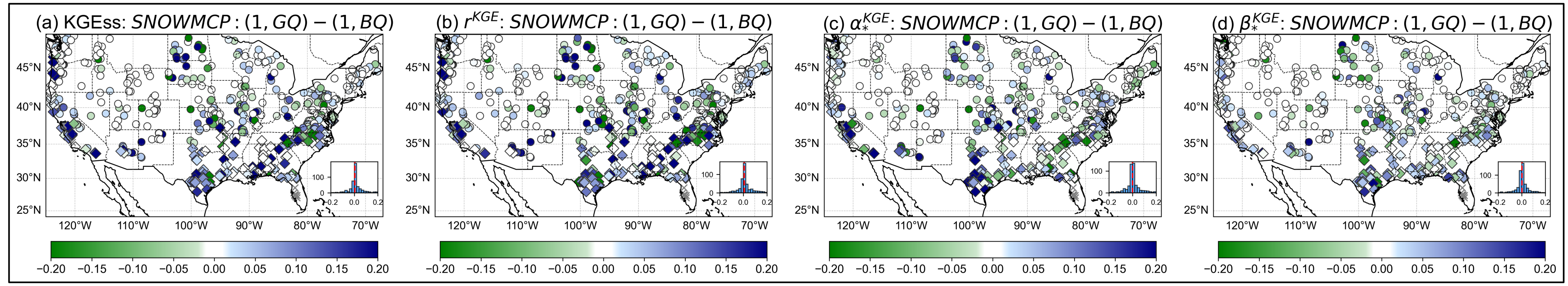

**Figure S7:** Performance differences between SNOWMCP(1,GQ) and SNOWMCP(1,BQ) for the following metrics: (a) KGE skill score ($KGE_{ss}$) and the three components of KGE, (b) linear correlation ($r^{KGE}$), (c) adjusted flow variability ratio ($\alpha_*^{KGE}$), and (d) mass balance ratio ($\beta_*^{KGE}$). Circles indicate snowy basins, and diamonds indicate non-snowy basins. The symbol × indicates where the model failed to finish training.

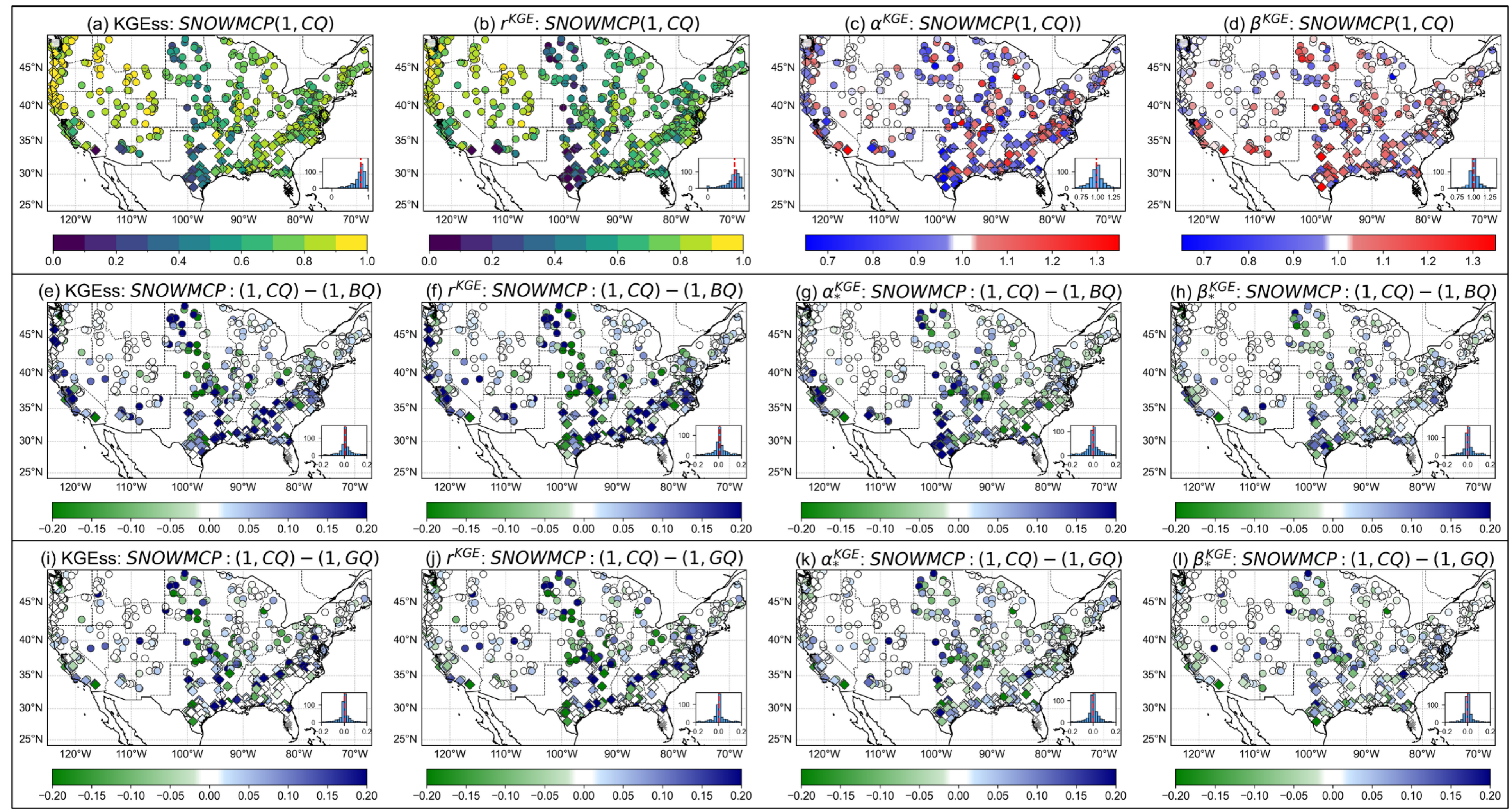

**Figure S8:** Performance of SNOWMCP(1,CQ) for the following metrics: (a) KGE skill score (KGEss) and the three components of KGE, (b) linear correlation ($r^{KGE}$), (c) adjusted flow variability ratio ($\alpha^{KGE}$), and (d) mass balance ratio ($\beta^{KGE}$). Panels (e)–(h) and (i)–(l) summarize the performance differences of SNOWMCP(1,CQ) against SNOWMCP(1,BQ) and SNOWMCP(1,GQ), respectively. Circles indicate snowy basins, and diamonds indicate non-snowy basins. The symbol × indicates where the model failed to finish training.

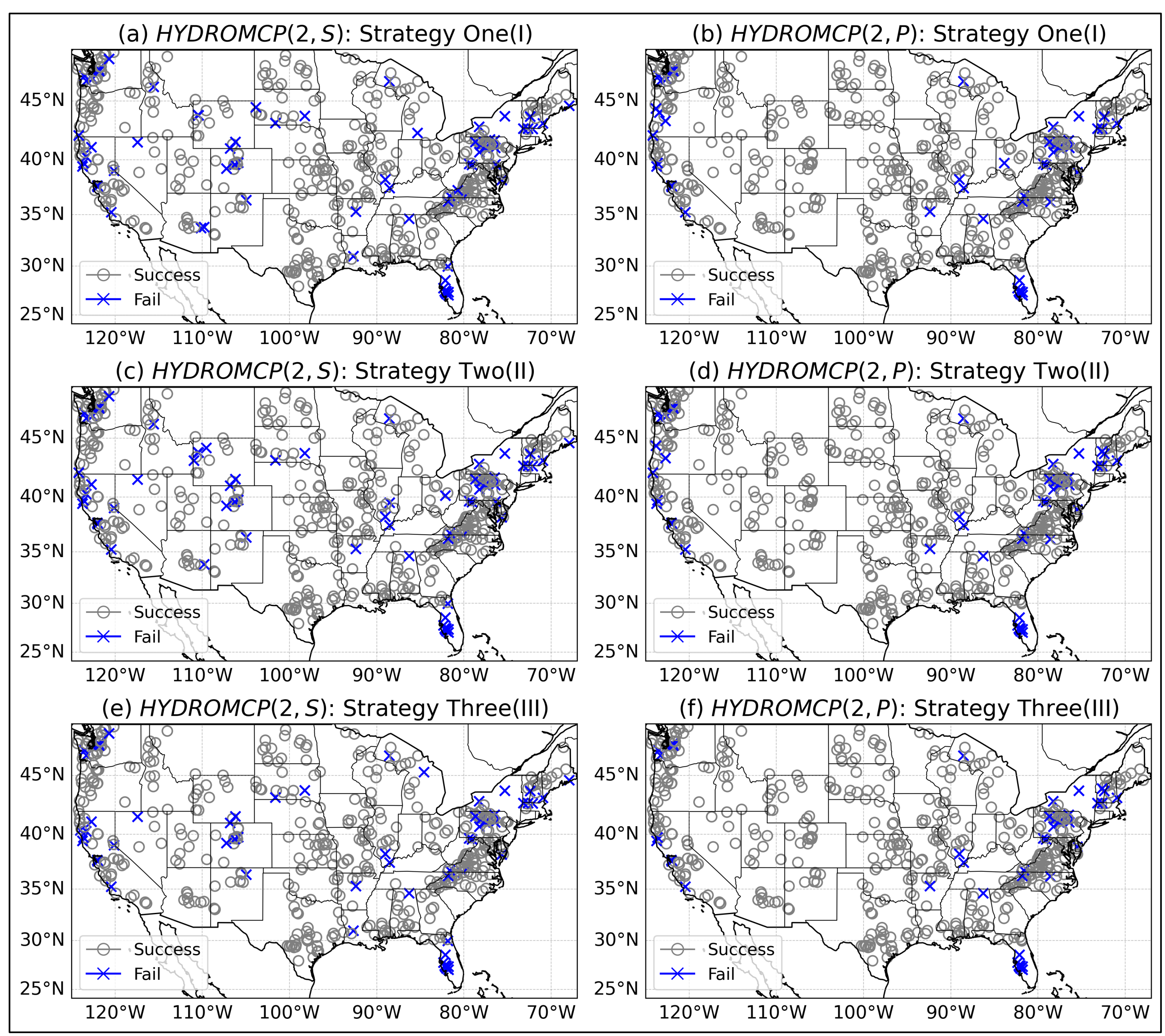

**Figure S9:** Success and failure flags for HYDROMCP(2) model training under three strategies: (a, c, e) serial (S) configuration and (b, d, f) parallel (P) configuration. S and P also refer to routing and bypass in the context of coupling HMCP and SNOWMCP. Gray circles indicate successful training and blue crosses indicate failed training.

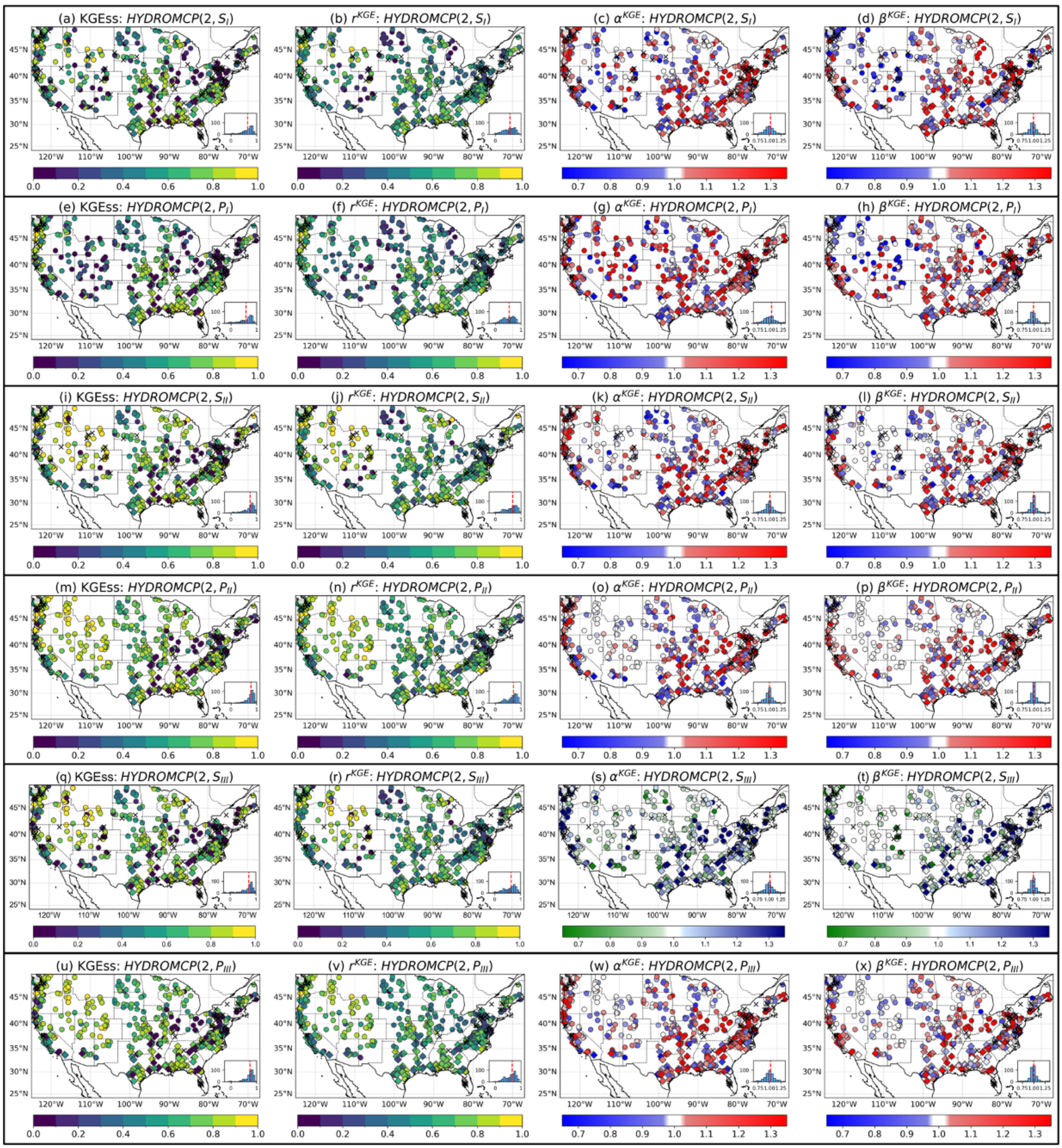

**Figure S10:** Spatial maps of performance metrics, including KGE skill score ($KGE_{ss}$) and the three components of KGE: linear correlation ($r^{KGE}$), flow variability ratio ($\alpha^{KGE}$), and mass balance ratio ($\beta^{KGE}$). Subplots (a–d) summarize HYDROMCP(2,S) under the first training strategy, (e–h) HYDROMCP(2,P) under the first strategy, (i–l) HYDROMCP(2,S) under the second strategy, (m–p) HYDROMCP(2,P) under the second strategy, (q–t) HYDROMCP(2,S) under the third strategy, and (u–x) HYDROMCP(2,P) under the third strategy. Here, S and P denote Serial and Parallel configurations, respectively. Circles indicate snowy basins, diamonds indicate non-snowy basins, and blue crosses (×) mark catchments where model training failed to complete.

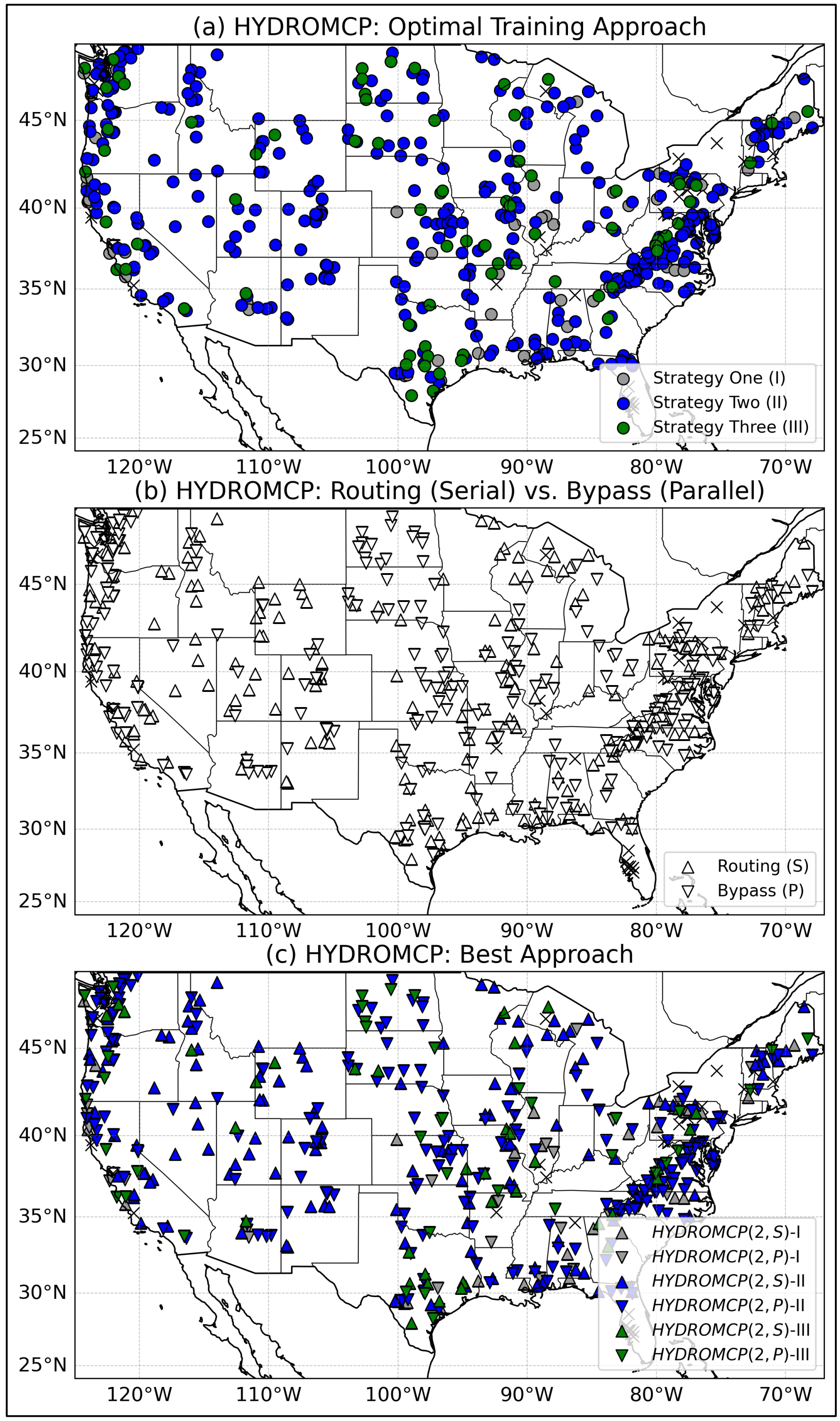

**Figure S11:** Spatial information for the HYDROMCP(2,B) model showing (a) the best training strategy, (b) the better-performing series (S) versus parallel (P) hypothesis, and (c) the best model hypothesis and training strategy for each catchment. Roman numerals denote training strategies. Circles indicate snowy basins, diamonds indicate non-snowy basins, and blue crosses (×) mark catchments where model training failed to complete.

**Figure S12:** Spatial information on the performance differences between HYDROMCP(2,$B^{+}$) and HYDROMCP(2,$B$), with locations with large differences highlighted in color

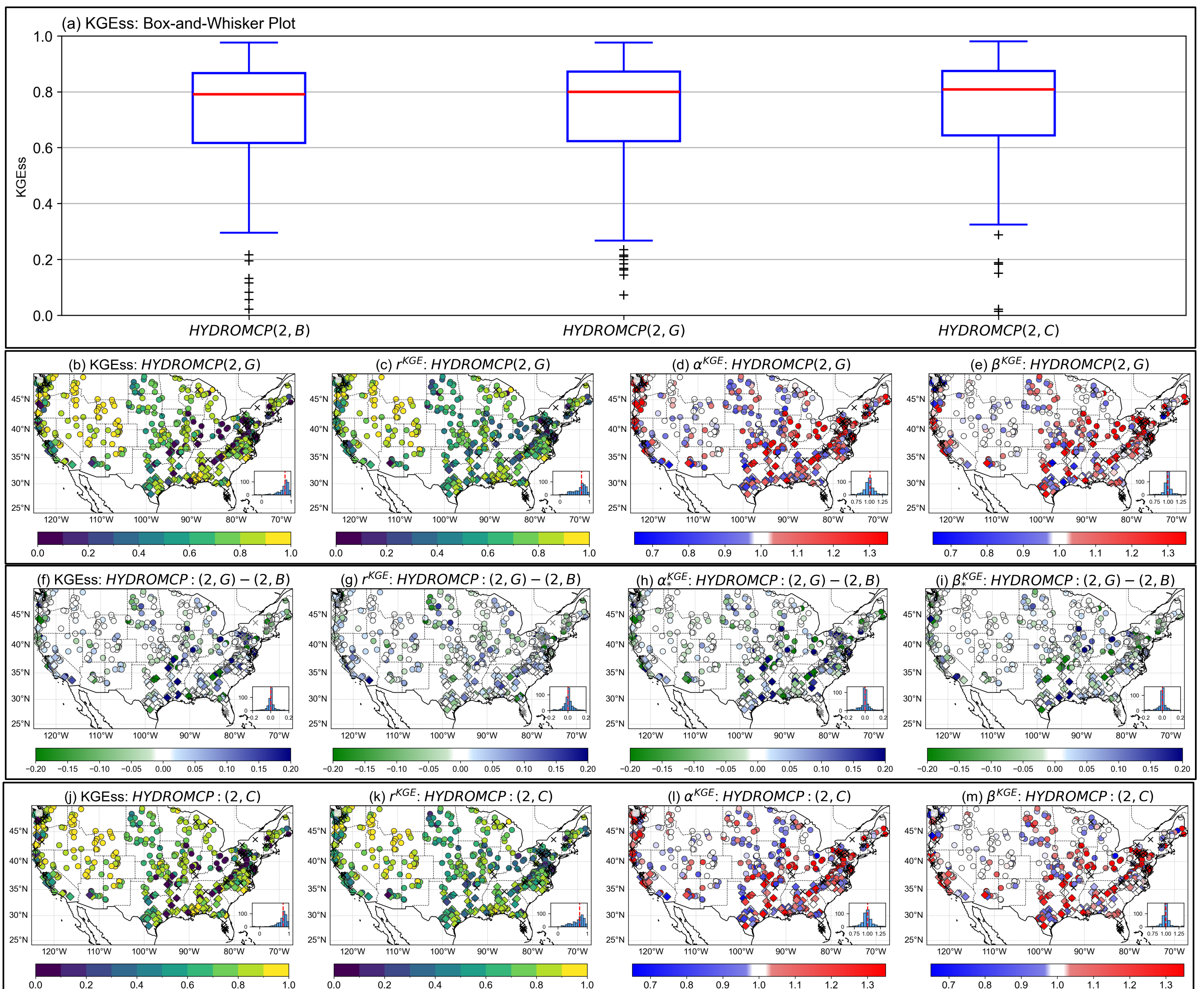

**Figure S13:** Skill metrics and component-wise comparisons for alternative $HYDROMCP$ architectures across 513 CONUS catchments. Panel (a) shows box-and-whisker plots of the KGE skill score ( $KGE_{ss}$ ) for $HYDROMCP(2, B)$, $HYDROMCP(2, G)$, and $HYDROMCP(2, C)$. Panels (b)–(e) show the spatial distributions of $KGE_{ss}$, the linear correlation component ($r^{KGE}$), the flow variability ratio ($\alpha^{KGE}$), and the mass balance ratio ($\beta^{KGE}$) for $HYDROMCP(2, G)$. Panels (f)–(i) show the corresponding differences between $HYDROMCP(2, G)$ and $HYDROMCP(2, B)$. Panels (j)–(m) show the same four metrics for $HYDROMCP(2, C)$, and panels (n)–(q) show the corresponding differences between $HYDROMCP(2, C)$ and $HYDROMCP(2, B)$. Circles indicate snowy basins, and diamonds indicate non-snowy basins.

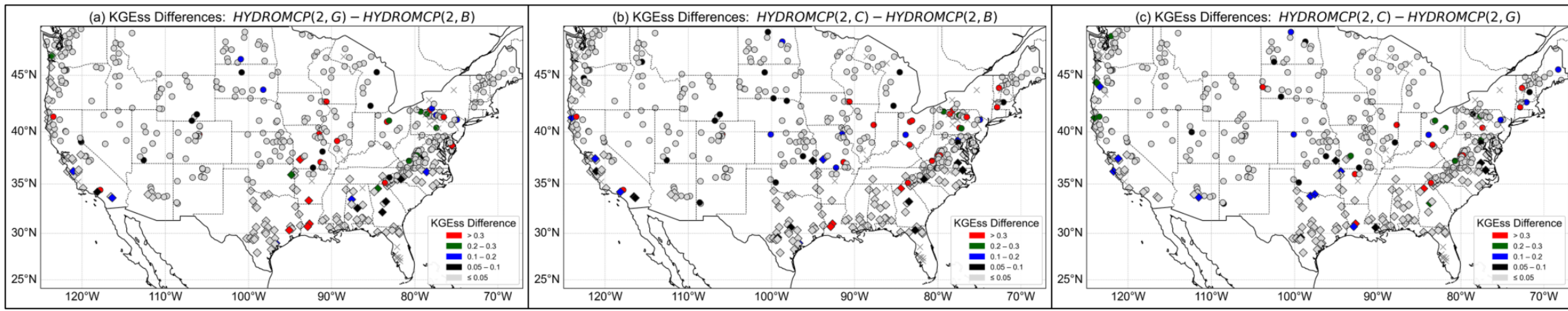

**Figure S14:** Spatial information on the performance differences between (a) HYDROMCP(2,G) and HYDROMCP(2,B), (b) HYDROMCP(2,C) and HYDROMCP(2,B), and (c) HYDROMCP(2,C) and HYDROMCP(2,G), with locations with large differences highlighted in color.

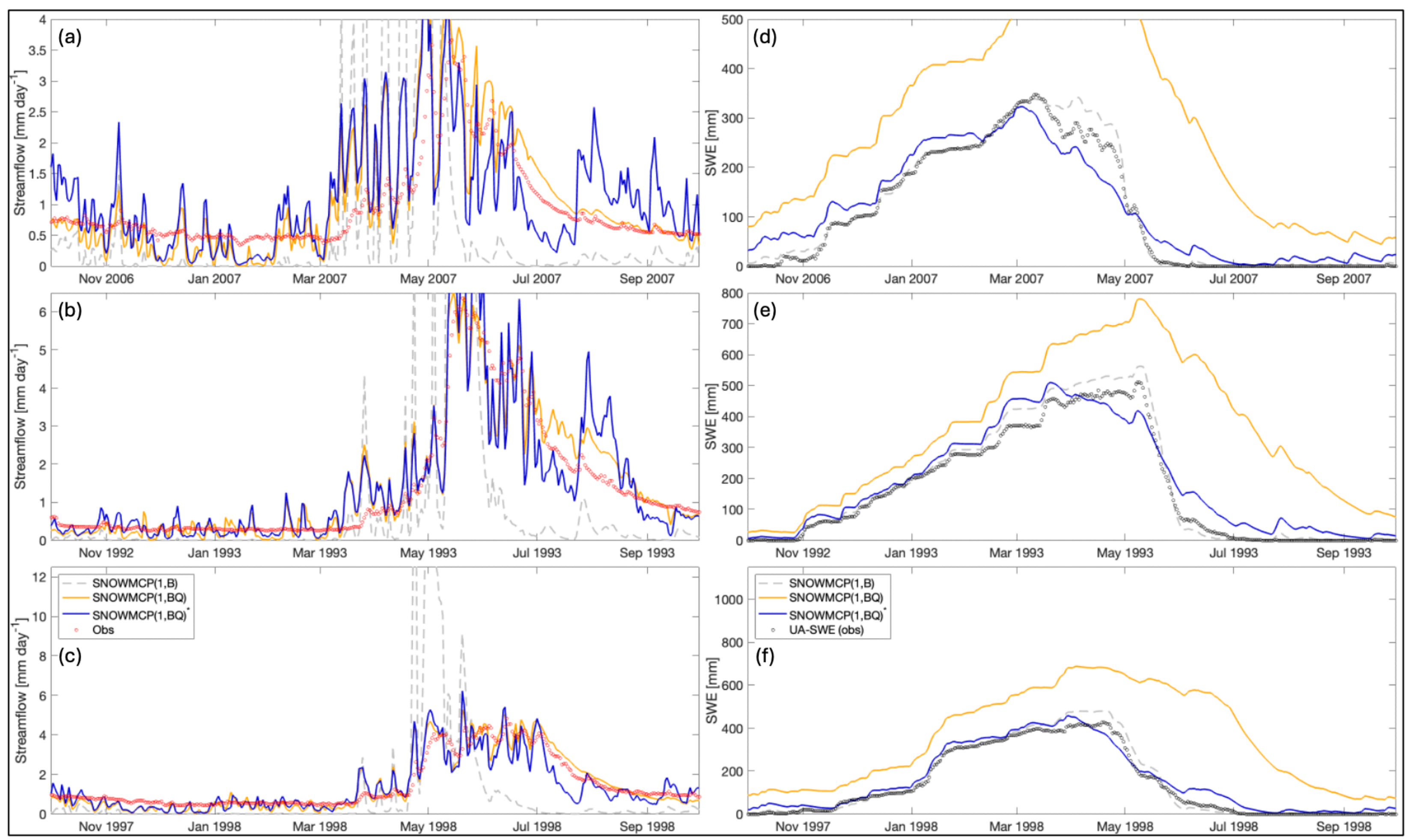

**Figure S15:** Time series generated by selected SNOWMCP-based models under different flow regimes for CAMELS-US basin 13023000. Panels (a–c) show simulated and observed streamflow during dry, median, and wet water years (WY 2007, WY 1993, and WY 1998, respectively). Panels (d–f) show the corresponding snow water equivalent (SWE) simulations and UA-SWE observations for the same water years. The models include SNOWMCP(1,B), SNOWMCP(1,BQ), and SNOWMCP(1,BQ)*, where SNOWMCP(1,BQ)* denotes the model trained jointly against streamflow and SWE.

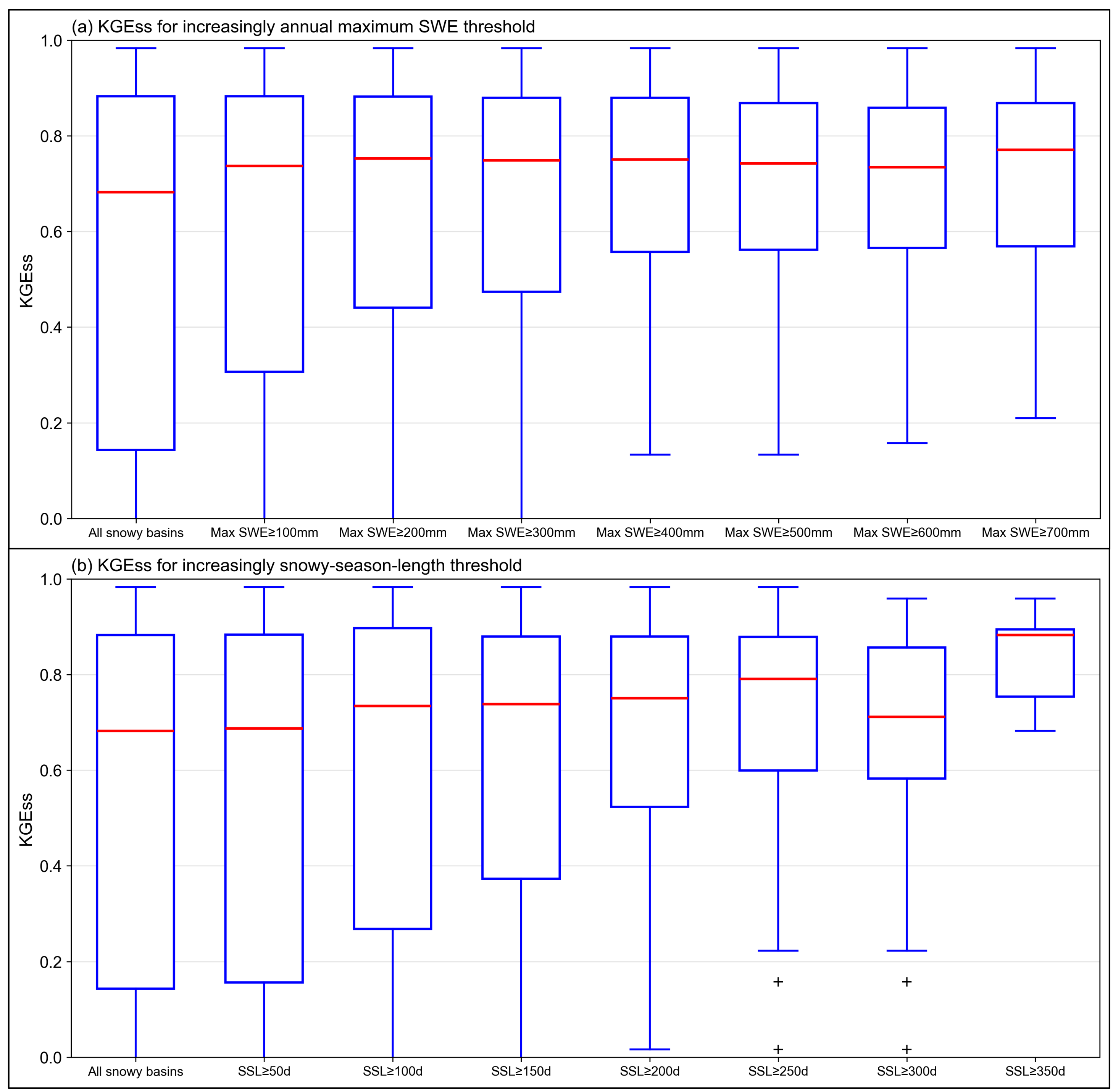

**Figure S16:** Box–whisker plots of the $KGE_{ss}$ metric for SWE simulations across CAMELS-US basins under increasingly restrictive snow-dominance criteria. Results are grouped by (a) annual maximum SWE thresholds and (b) snowy-season-length (SSL) thresholds. SSL denotes snowy season length, defined here as the median annual difference between the last snowy day and the first snowy day within WY1982–2008, where a snowy day is defined as SWE > 3 mm.

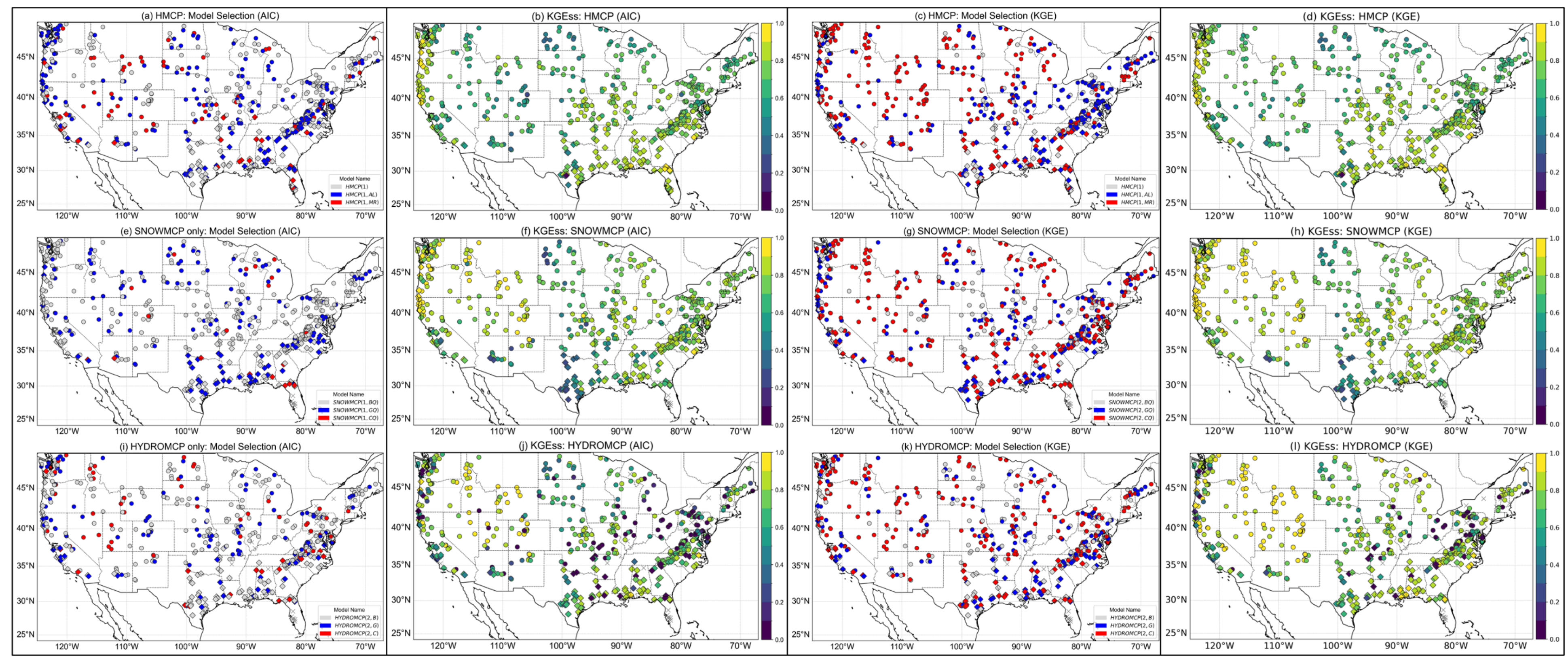

**Figure S17:** Spatial distribution of (a) model selection results based on the minimum-AIC metric for the HMCP family of models and their associated $KGE_{ss}$values in (b), and (c) model selection results based purely on the maximum-KGE metric and their associated $KGE_{ss}$ values in (d). Corresponding results for SNOWMCP are shown in subplots (e–h), and for HYDROMCP in subplots (i–l).

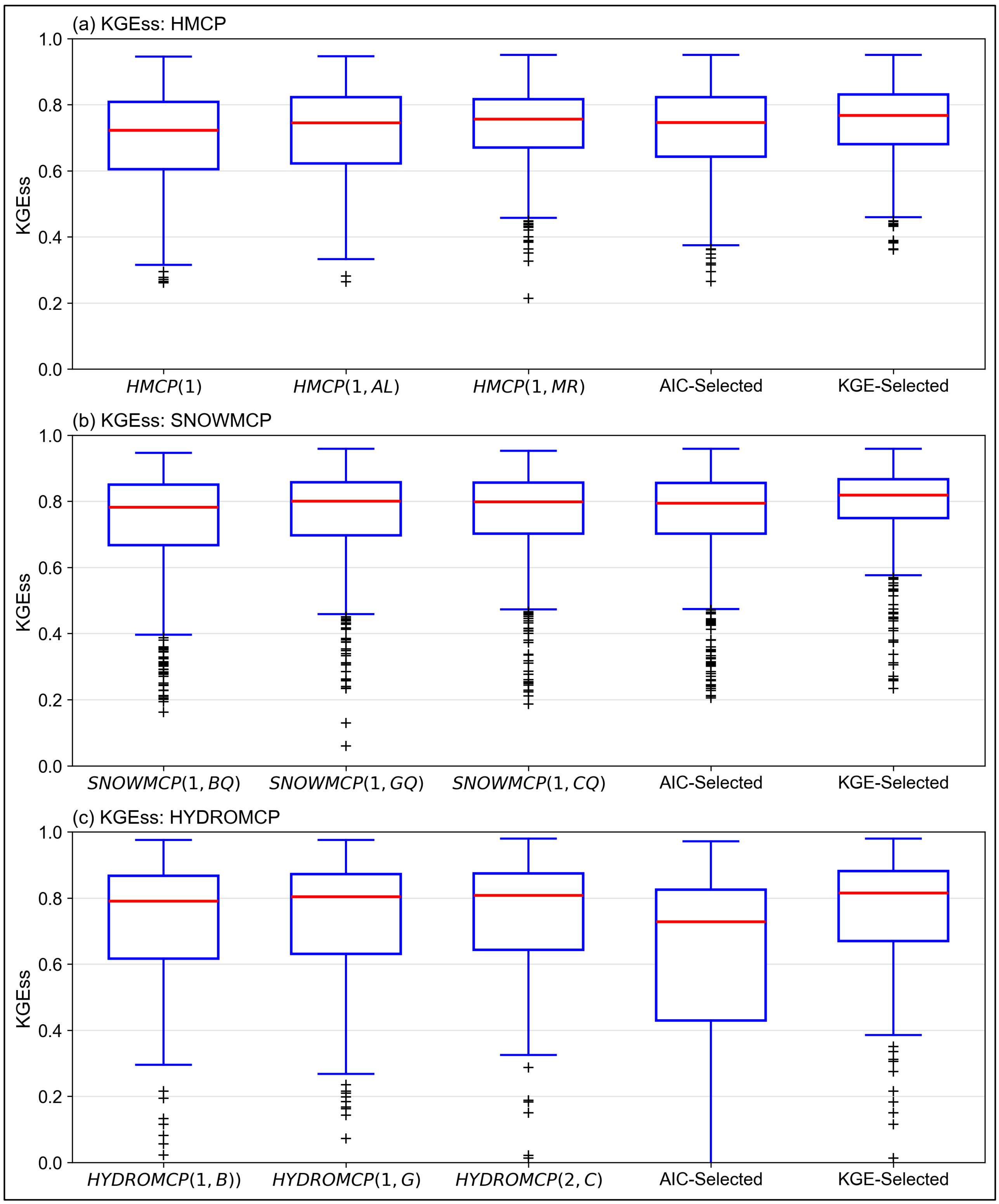

**Figure S18:** Box–whisker plots of the $KGE_{ss}$ metric across 513 basins for (a) HMCP, (b) SNOWMCP, and (c) HYDRO-MCP. For each model category, results are shown for both minimum-AIC–based selection and maximum-KGE–based selection.

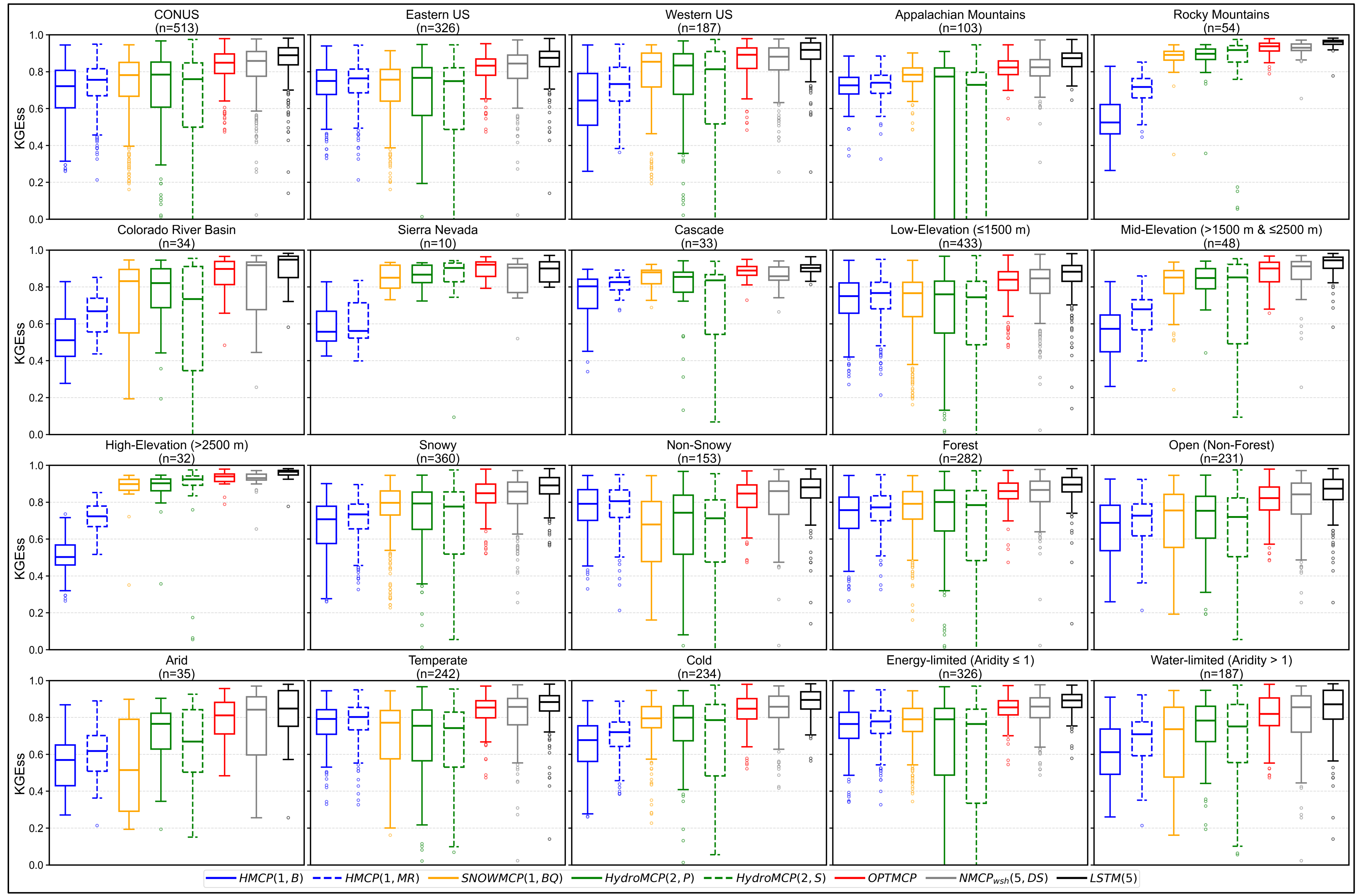

**Figure S19:** Box–whisker plots of $KGE_{ss}$ across CAMELS-US basin groups defined by geographical region, mountain region, elevation, snow condition, land cover, climate class, and aridity regime. Each panel shows the distribution of $KGE_{ss}$ for the corresponding basin group, with the number of basins indicated in parentheses. Only two basins, 07083000 and 09034900, are classified as polar-climate basins; their corresponding $KGE_{ss}$ scores, ordered from left to right according to the legend, are (0.51, 0.32), (0.84, 0.80), (0.94, 0.92), (0.95, 0.36), (0.95, NA), (0.95, 0.96), (0.95, 0.95), and (0.98, 0.98), respectively.

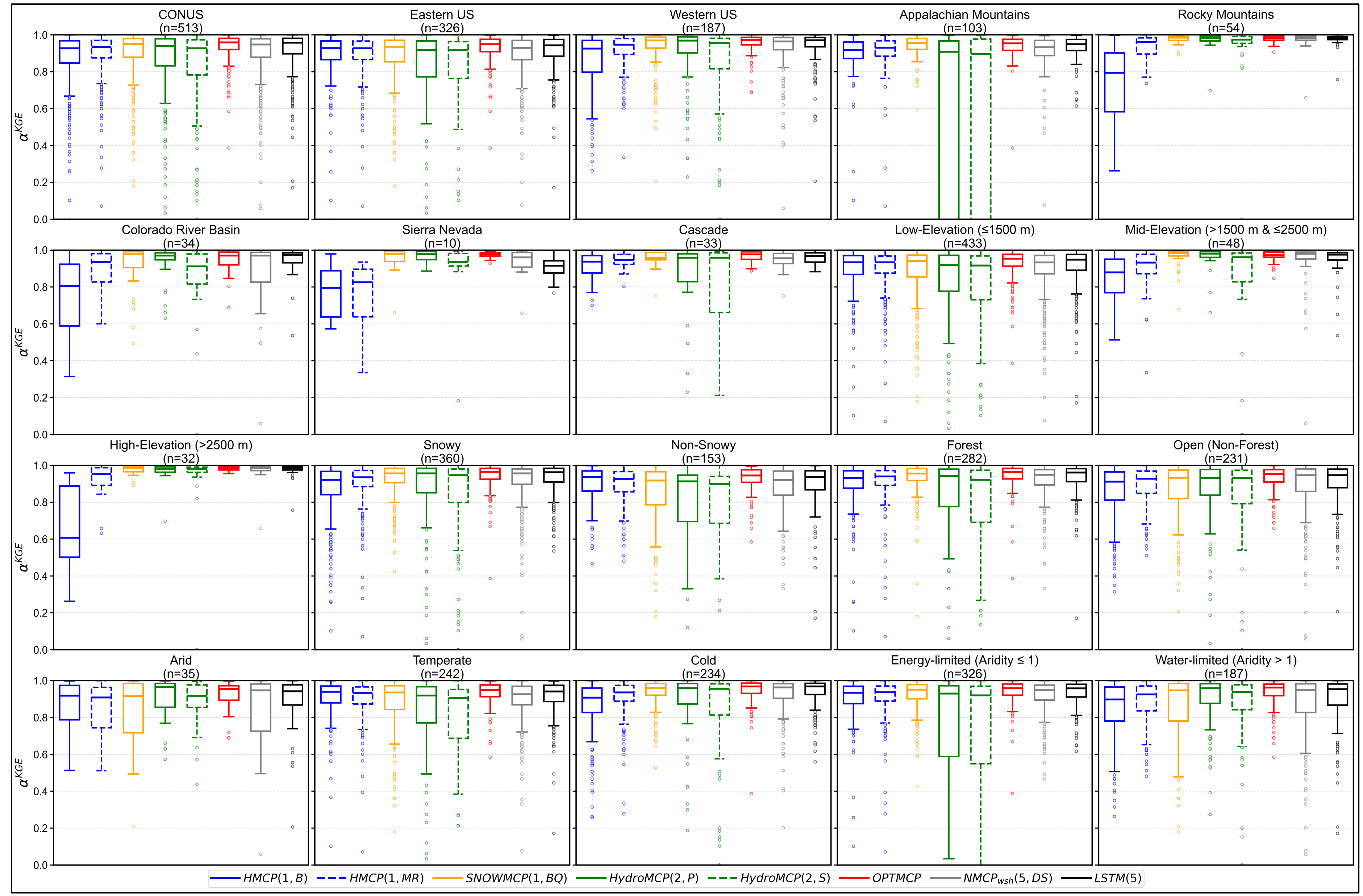

**Figure S20:** Box–whisker plots of $\alpha_*^{KGE}$ across CAMELS-US basin groups defined by geographical region, mountain region, elevation, snow condition, land cover, climate class, and aridity regime. Each panel shows the distribution of $\alpha_*^{KGE}$ for the corresponding basin group, with the number of basins indicated in parentheses. Note that $\alpha_*^{KGE} = 1 - |1 - \alpha^{KGE}|$, which is optimized at 1 and is unbounded below. Only two basins, 07083000 and 09034900, are classified as polar-climate basins; their corresponding $\alpha_*^{KGE}$ scores, ordered from left to right according to the legend, are (0.59, 0.35), (0.98, 0.98), (0.98, 0.99), (0.97, 0.70), (0.98, NA), (0.98, 1.00), (1.00, 0.98), and (0.99, 1.00), respectively.

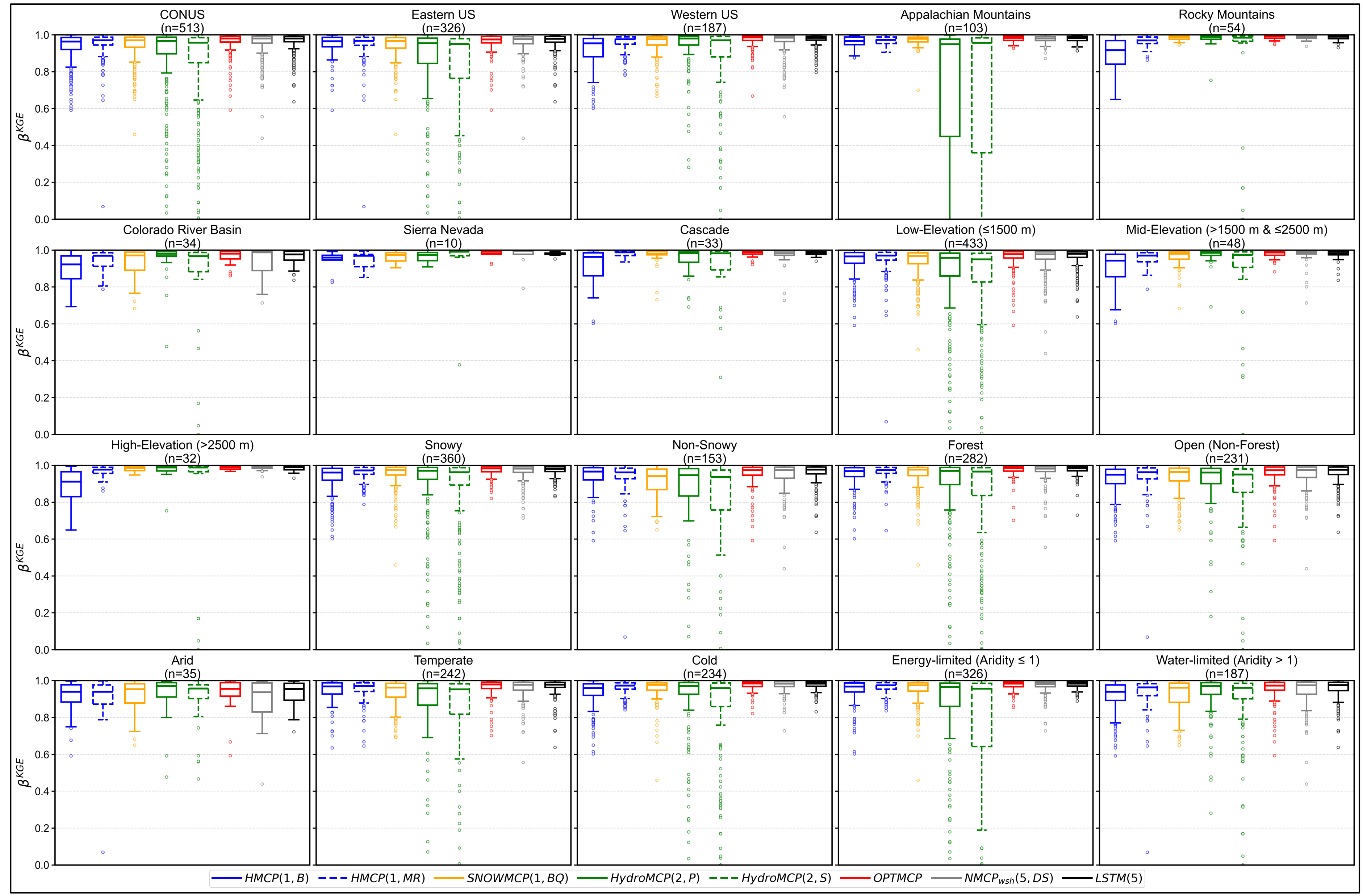

**Figure S21:** Box–whisker plots of $\beta_*^{KGE}$ across CAMELS-US basin groups defined by geographical region, mountain region, elevation, snow condition, land cover, climate class, and aridity regime. Each panel shows the distribution of $\beta_*^{KGE}$ for the corresponding basin group, with the number of basins indicated in parentheses. Note that $\beta_*^{KGE} = 1 - |1 - \beta^{KGE}|$, which is optimized at 1 and is unbounded below. Only two basins, 07083000 and 09034900, are classified as polar-climate basins; their corresponding $\beta_*^{KGE}$ scores, ordered from left to right according to the legend, are (0.82, 0.76), (0.99, 1.00), (0.99, 0.99), (0.99, 0.75), (0.98, NA), (0.98, 1.00), (0.99, 0.99), and (1.00, 1.00), respectively.

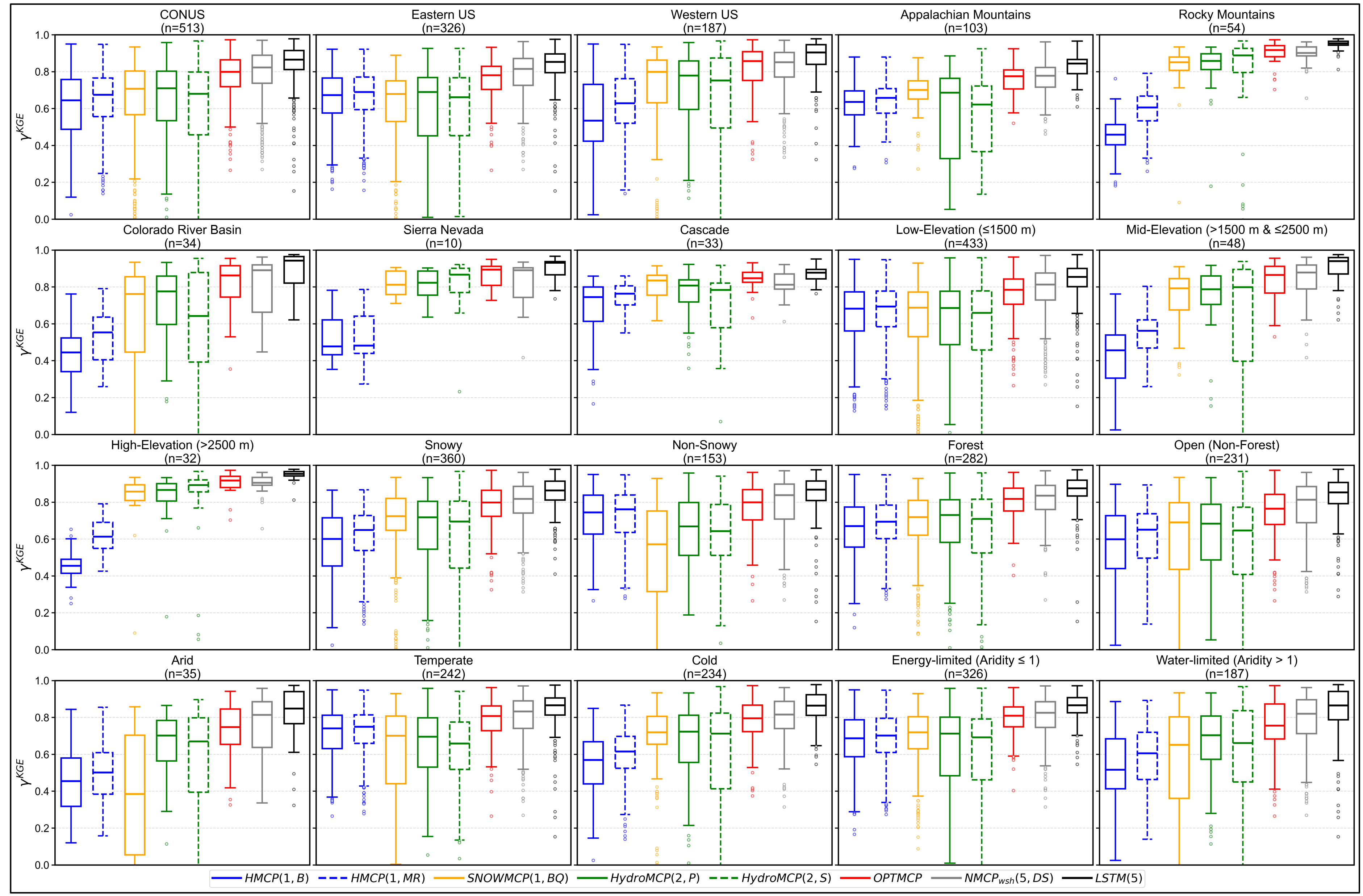

**Figure S22:** Box–whisker plots of $\gamma^{KGE}$ across CAMELS-US basin groups defined by geographical region, mountain region, elevation, snow condition, land cover, climate class, and aridity regime. Each panel shows the distribution of $\gamma^{KGE}$ for the corresponding basin group, with the number of basins indicated in parentheses. Only two basins, 07083000 and 09034900, are classified as polar-climate basins; their corresponding $\gamma^{KGE}$ scores, ordered from left to right according to the legend, are (0.46, 0.42), (0.77, 0.72), (0.92, 0.89), (0.93, 0.18), (0.93, NA), (0.94, 0.95), (0.94, 0.93), and (0.97, 0.97), respectively.